\documentclass[conference]{IEEEtran}
\IEEEoverridecommandlockouts
\usepackage{cite}
\usepackage{amsmath,amssymb,amsfonts}
\usepackage{stmaryrd}
\usepackage{algorithmic}
\usepackage{graphicx}
\usepackage{hhline}
\usepackage{tabularx}
\usepackage{multirow}
\usepackage{booktabs}
\usepackage{textcomp}
\usepackage{xcolor}
\usepackage{colortbl}
\usepackage{hyperref}
\hypersetup{
    colorlinks=true,
    linkcolor=black, 
    citecolor=blue, 
    urlcolor=blue, 
}
\makeatletter
\providecommand{\insert@pcolumn}{\insert@column}
\makeatother
\usepackage{tikz}
\usetikzlibrary{trees,calc,arrows.meta,intersections,patterns,positioning,shapes.misc,shapes.geometric,fadings,through,decorations.pathreplacing,arrows,automata,fit}
\usepackage{subcaption}
\usepackage{scalerel}
\usepackage{tikz-qtree}
\usepackage{lipsum,adjustbox}
\usepackage{url}
\usepackage{listings} 
\usepackage[activate=false,kerning]{microtype}
\SetExtraKerning[unit=character]{encoding=*}{\textemdash={100,100}}
\usepackage{bbm}
\usepackage{pifont}     
\usepackage{threeparttable}
\usepackage{adjustbox}
\newcommand{\cmark}{\textcolor{green!60!black}{\ding{51}}}

\usepackage{rotating}   
\usepackage{xspace}
\usepackage{epigraph}
\newcommand{\latex}{\LaTeX\xspace}

\lstdefinelanguage{Lean}{
  keywords={import,theorem,example,def,lemma,by,intro,intros,have,show,exact,apply,rw,simp,norm_num,induction,where,match,with,fun,forall,exists,Prop,Type},
  keywordstyle=\color{blue}\bfseries,
  ndkeywords={Nat,Int,Real,Rational,Prime,True,False},
  ndkeywordstyle=\color{purple}\bfseries,
  identifierstyle=\color{black},
  sensitive=false,
  comment=[l]{--},
  morecomment=[s]{/-}{-/},
  commentstyle=\color{green!60!black}\itshape,
  stringstyle=\color{red}\ttfamily,
  morestring=[b]',
  morestring=[b]",
  alsoletter={∀,∃,∧,∨,¬,⊆,∪,∩,∈,ℚ,ℝ},
  basicstyle=\ttfamily\small,
  breaklines=true,
  breakatwhitespace=true,
  mathescape=true,
  tabsize=2,
  columns=fullflexible,
}
\lstdefinestyle{LeanFigure}{
  language=Lean,
  basicstyle=\ttfamily\scriptsize,
  keywordstyle=\color{Violet!85!black}\bfseries,
  ndkeywordstyle=\color{Indigo!85!black}\bfseries,
  commentstyle=\color{green!45!black}\itshape,
  stringstyle=\color{Magenta!70!black},
  showstringspaces=false,
  keepspaces=true,
  breaklines=true,
  breakatwhitespace=true,
  columns=fullflexible,
  aboveskip=0pt,
  belowskip=0pt,
  literate={∀}{{$\forall$}}1 {∃}{{$\exists$}}1 {∧}{{$\wedge$}}1 {≤}{{$\leq$}}1 {ℕ}{{$\mathbb{N}$}}1,
}

\tikzset{
    boxinformal/.style={draw, rounded corners, inner sep=5pt, fill=blue!5, text width=0.9\columnwidth, align=center},
    boxformal/.style={draw, rounded corners, inner sep=5pt, fill=green!3, text width=0.9\columnwidth},
    discoveryheader/.style={text=green!50!black, font=\small},
    discoveryprogression/.style={font=\fontsize{11}{13}\selectfont\bfseries, color=green!50!black},
    hexpoint/.style={circle, fill=red, inner sep=1.2pt},
}

\definecolor{Indigo}{HTML}{211cbf}
\definecolor{Cyan}{HTML}{74d2fa}
\definecolor{Violet}{HTML}{7A4DFF}
\definecolor{Magenta}{HTML}{ec8dff}
\definecolor{LavenderOne}{HTML}{A7A8F0}
\definecolor{LavenderTwo}{HTML}{9194E4}
\definecolor{LavenderThree}{HTML}{8180DE}
\definecolor{LavenderFour}{HTML}{6E65CE}
\definecolor{LavenderFive}{HTML}{5445B5}

\definecolor{VivCyan}{HTML}{3FB8DD}
\definecolor{VivTeal}{HTML}{62E8E8}
\definecolor{VivLime}{HTML}{A4F02C}
\definecolor{VivYellow}{HTML}{F0C400}
\definecolor{VivOrange}{HTML}{FFA00B}
\definecolor{VivSalmon}{HTML}{F77B7B}
\definecolor{VivPink}{HTML}{F35DB7}
\definecolor{VivLilac}{HTML}{DC8DFF}
\definecolor{VivPurple}{HTML}{9F49FF}

\newcommand{\thead}[1]{\textbf{#1}}
\newcommand{\tgroup}[2]{\rowcolor{Violet!12}\multicolumn{#1}{l}{\textcolor{Indigo}{\bfseries \textit{#2}}}\\}

\newcommand{\timportant}{\rowcolor{Magenta!13}}
\newcommand{\tsoft}{\rowcolor{Cyan!10}}

\newcommand{\tabletop}{\arrayrulecolor{black}\toprule}
\newcommand{\tablemid}{\arrayrulecolor{black}\midrule}
\newcommand{\tablebottom}{\arrayrulecolor{black}\bottomrule\arrayrulecolor{black}}
\def\BibTeX{{\rm B\kern-.05em{\sc i\kern-.025em b}\kern-.08em
    T\kern-.1667em\lower.7ex\hbox{E}\kern-.125emX}}

\begin{document}

\title{Artificial Intelligence for Mathematical Reasoning: An Integrated Survey of Language Models, Neuro-symbolic Systems, and Verified Discovery}

\author{
\IEEEauthorblockN{Syed Rifat Raiyan$^1$,
Mohsinul Kabir$^2$\IEEEauthorrefmark{2},
Hasan Mahmud$^1$,
Md Kamrul Hasan$^1$, and Sophia Ananiadou$^2$}
\IEEEauthorblockA{$^1$Systems and Software Lab (SSL),
Department of Computer Science and Engineering,\\
Islamic University of Technology, Gazipur, Bangladesh\\
Email: \{rifatraiyan, hasan, hasank\}@iut-dhaka.edu}
\IEEEauthorblockA{$^2$Department of Computer Science,
University of Manchester, Manchester, United Kingdom\\
Email: \{mdmohsinul.kabir, sophia.ananiadou\}@manchester.ac.uk \quad\IEEEauthorrefmark{2}Corresponding author}
}

\maketitle

\begin{abstract}
Mathematical reasoning has long served as a stringent test of machine intelligence; over the past decade, it has moved from a niche problem within natural language processing to one of the most consequential artificial intelligence (AI) frontiers. This survey provides a unified account of the field's evolution, from early rule-based math word problem (MWP) solvers and template-driven geometry systems, through neural expression generation and large language model prompting, to contemporary reasoning models, multi-agent systems, neuro-symbolic theorem provers, and verified discovery workflows. We organize the landscape along four axes: (i) informal reasoning over text and diagrams, spanning MWP solving, multimodal geometry, and vision--language models; (ii) formal reasoning in proof assistants, including autoformalization, tactic prediction, compiler-guided repair, and proof search; (iii) mathematical discovery, where systems propose constructions, improve bounds, or assist attacks on open problems; and (iv) the inference- and training-time techniques, including chain-of-thought prompting, tool use, process reward models, and reinforcement learning with verifiable rewards, that increasingly connect generation with verification. We catalog major benchmarks across grade-school arithmetic, competition mathematics, geometry, formal proving, multimodal and multilingual reasoning, and expert evaluation, and we examine benchmark saturation, contamination, reporting mismatches, and the distinction between pass@$1$, majority voting, and verifier-assisted pass@$k$. We critically assess failure modes: brittleness under perturbation, reward hacking, multimodal grounding failures, fragile formalization, and the energy cost of reasoning-scale inference. Drawing on recent perspectives from working mathematicians, we identify future directions centered on verified-discovery workflows, reasoning efficiency, and infrastructure to make AI-assisted formalization broadly usable. Companion materials: \url{https://github.com/Starscream-11813/awesome-AI4Math}.
\end{abstract}


\epigraph{\itshape ``Have you reason? I have. Why don’t you use it? When it performs its proper office, what more do you require?''}{--- Marcus Aurelius, \textit{Meditations} (Book IV, \S13)}

\section{Introduction}
The automation of mathematical reasoning has been a defining ambition of artificial intelligence since its inception~\cite{feigenbaum1963computers,bobrow1964natural}. What began in the mid-1960s as brittle pattern-matching programs for templated arithmetic has, within the past four years alone, expanded into a landscape of systems that solve Olympiad-level problems, formally verify mathematical arguments in Lean~4, and contribute to the resolution of selected open problems posed by Paul Erd\H{o}s. This survey provides an integrated account of that arc, connecting the classical MWP lineage, multimodal geometry, formal theorem proving, and verified mathematical discovery through what we will call the \emph{reasoning-model era}---the period beginning with OpenAI~o1 in late~2024 and continuing through DeepSeek-R1, Kimi~k1.5, and Gemini~Deep~Think in 2025--2026, during which long-horizon chain-of-thought generation, reinforcement learning from verifiable rewards (RLVR), and test-time scaling became the dominant levers of progress on mathematical benchmarks.

\subsection{Why Mathematical Reasoning?}
Mathematics occupies a privileged place within AI research for three intertwined reasons. First, it is \emph{verifiable}: unlike open-ended dialogue or summarization, the correctness of a solution can in principle be decided mechanically, either by arithmetic or by a proof assistant. Second, it is \emph{compositional}: mastering mathematics demands not simply pattern recognition but the disciplined combination of definitions, lemmas, and logical steps, a capability that has historically eluded purely statistical approaches. Third, it is a \emph{core cognitive benchmark}: mathematical aptitude has long been used to gauge human intellectual development~\cite{jbpetersoniq,sheripsychology} and serves as a natural yardstick for machine cognition. The convergence of these three properties makes mathematical reasoning, \textit{prima facie}, both a technical challenge and a proving ground for claims about what LLMs can and cannot truly do.

\subsection{Canonical Tasks and Running Examples}
\begin{figure}[t]
        \centering
        \includegraphics[width=\linewidth]{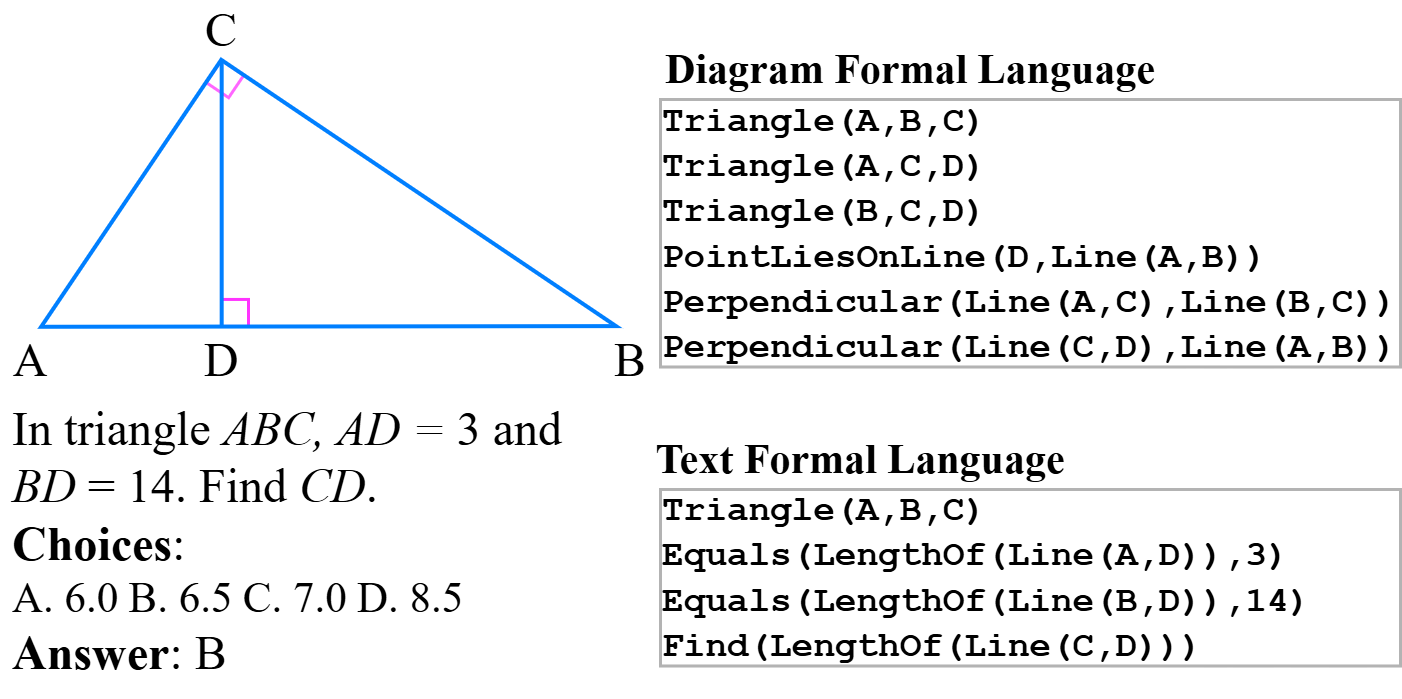}
        \caption{An example from the \textsc{Geometry3K} \cite{lu2021inter} dataset.}
        \label{fig:fig1}
    \end{figure}

Throughout this survey we distinguish four canonical families of tasks:
\begin{itemize}
    \item \textbf{Math Word Problems (MWPs).} A textual narrative describes a scenario involving one or more unknown numerical quantities and poses a question. The solver must recover a valid mathematical expression that evaluates to the correct numeric answer. Table~\ref{tab:tab1} shows a classical example.
    \item \textbf{Geometry problem solving.} A problem consists of a textual description together with one or more diagrams. The solver must perform \emph{joint reasoning} over text and image, apply axiomatic knowledge, and produce either a numeric answer or a rigorous proof. Figure~\ref{fig:fig1} illustrates an example from \textsc{Geometry3K}~\cite{lu2021inter}.
    \item \textbf{Formal theorem proving and autoformalization.} Given a mathematical statement, often originally expressed in natural language, the system must produce a proof that is mechanically verified by a proof assistant such as Lean~4~\cite{demoura2021lean4} or Isabelle. Autoformalization refers to the prerequisite task of translating informal mathematical text into formal statements.
    \item \textbf{Open-ended mathematical discovery.} The system is asked to improve known bounds, generate counterexamples, or solve problems with no published solution. This class encompasses the recent wave of AI-assisted attacks on Erd\H{o}s problems and the algorithmic discoveries of \textsc{FunSearch}~\cite{romera2024funsearch} and \textsc{AlphaEvolve}~\cite{novikov2025alphaevolve}.
\end{itemize}

    \begin{table}[t]
    \centering
    \small
    \renewcommand{\arraystretch}{1.12}
    \begin{tabular}{p{0.94\columnwidth}}
    \tabletop
    \thead{MWP Example} \\
    \tablemid
    \tsoft \textbf{Problem:} 69 handbags are sold for \$23 each. There are a total of 420 handbags in a boutique and the remaining handbags are sold for \$17 each. How much did the boutique earn after selling all the handbags? \\
    \rowcolor{Violet!8}\textbf{Expression:} $x = 69\times23+(420-69)\times17$ \\
    \timportant \textbf{Solution:} 7554 \\
    \tablebottom
    \end{tabular}
    \caption{A prototypical Math Word Problem (MWP). Shading separates the problem statement, the induced expression, and the final answer.}
    \label{tab:tab1}
    \end{table}

    \begin{figure}[h]
    \centering
    \resizebox{\columnwidth}{!}{%
    \begin{tikzpicture}[
        ibox/.style={rectangle, rounded corners=1.5mm, draw=Indigo!60,
            fill=LavenderOne!18, text width=0.44\textwidth, align=left,
            inner xsep=6pt, inner ysep=5pt, font=\small},
        fbox/.style={rectangle, rounded corners=1.5mm, draw=Violet!70,
            fill=Cyan!10, text width=0.46\textwidth, align=left,
            inner xsep=6pt, inner ysep=5pt},
        label/.style={font=\footnotesize\bfseries, text=Indigo!80},
        arr/.style={-{Latex[length=2mm]}, thick, draw=Indigo!55}
    ]
     
    \node[ibox] (informal) at (0, 0) {%
        \textbf{Theorem} (Euclid). \textit{There are infinitely many prime numbers.}\\[4pt]
        \textrm{\footnotesize\textit{Proof sketch.}} {\footnotesize Given any finite list of primes $p_1, \dots, p_k$, the number $N = p_1 \cdots p_k + 1$ has a prime factor not in the list. $\square$}
    };
     
    \node[fbox] (formal) at (0, -3.6) {%
        {\ttfamily\scriptsize
        \textcolor{green!45!black}{-- Lean 4 (\texttt{mathlib})}\\[2pt]
        \textcolor{Violet!85!black}{\textbf{import}} Mathlib.Data.Nat.Prime.Infinite\\[4pt]
        \textcolor{Violet!85!black}{\textbf{theorem}} infinitude\_of\_primes :\\
        \ \ \ \textcolor{black!75}{$\forall$} N : \textcolor{Indigo!85!black}{Nat}, \textcolor{black!75}{$\exists$} p : \textcolor{Indigo!85!black}{Nat},\\
        \ \ \ \ \ N \textcolor{black!75}{$\leq$} p \textcolor{black!75}{$\wedge$} Nat.Prime p := \textcolor{Violet!85!black}{\textbf{by}}\\
        \ \ intro N\\
        \ \ \textcolor{Violet!85!black}{\textbf{exact}} Nat.exists\_infinite\_primes N
        }
    };
     
    \node[label, anchor=south west] at ([yshift=2pt]informal.north west) {Informal (natural language)};
    \node[label, anchor=south west, text=Violet!80] at ([yshift=2pt]formal.north west) {Formal + proof (Lean 4)};
     
    \draw[arr] ([xshift=50pt]informal.south) -- node[right=3pt, font=\footnotesize, text=Indigo!70, align=left] {autoformalization\\[-1pt]\footnotesize\textit{(LLM-assisted)}} ([xshift=50pt]formal.north);
     
    \draw[arr, draw=Magenta!60] ([xshift=30pt]formal.north) -- node[left=3pt, font=\footnotesize, text=Magenta!65, align=right] {type-check\\[-1pt]\footnotesize\textit{(Lean kernel)}} ([xshift=30pt]informal.south);
     
    \end{tikzpicture}%
    }
    \caption{Autoformalization and formal theorem proving illustrated on Euclid's theorem. The downward arrow represents autoformalization: an LLM translates the informal statement into a Lean~4 statement and proof. The upward arrow represents verification: the Lean kernel mechanically checks every inference step, providing a trust guarantee that no informal argument can match. This two-way pipeline---generate informally, verify formally---is the central architecture of modern AI-assisted theorem proving~\cite{wu2022autoformalization,jiang2023draft}.}
    \label{fig:autoformalization_example}
    \end{figure}

    \begin{table}[t]
\centering
\small
\renewcommand{\arraystretch}{1.12}
\begin{tabular}{p{0.94\columnwidth}}
\tabletop
\thead{Discovery Example: Erd\H{o}s Minimum Overlap Problem} \\
\tablemid
\tsoft \textbf{Problem (Erd\H{o}s, 1955).} Let $A \cup B = [2n]$ be a partition with $|A|=|B|=n$. For each integer $-2n < k < 2n$, define $M_k$ as the number of solutions $(a,b) \in A \times B$ to $a - b = k$. Estimate
\[
M(n) \;=\; \min_{A \cup B = [2n]} \;\max_{-2n < k < 2n}\; M_k.
\]
The \emph{minimum overlap constant} $\mu = \lim_{n\to\infty} M(n)/n$ is known to satisfy $0.3564 \leq \mu \leq 0.3810$ \cite{white2022erd}. \\
\rowcolor{Violet!8}\textbf{AI contribution.} The problem appears in the \textsc{AlphaEvolve} repository of open problems~\cite{novikov2025alphaevolve}. Discovery systems attack it by evolving partition-construction programs and using $M(n)$ as an automated fitness function. The task is \emph{open-ended}: no fixed answer set exists, and any improvement to either bound constitutes a novel mathematical result. \\
\timportant \textbf{Verification.} Unlike MWPs, the ``answer'' is a construction (an explicit partition) whose quality is machine-checkable but whose optimality requires proof, increasingly, in Lean~4. \\
\tablebottom
\end{tabular}
\caption{A prototypical open-ended mathematical discovery problem. The solver must produce an explicit construction (a partition of $[2n]$) that improves a known bound, not merely match a reference answer. Verification requires evaluating the construction's fitness and, for a definitive result, a formal proof of the bound. This problem class is the target of evolutionary program-search systems such as \textsc{FunSearch}~\cite{romera2024funsearch} and \textsc{AlphaEvolve}~\cite{novikov2025alphaevolve}.}
\label{tab:discovery_example}
\end{table}
\subsection{Challenges}
While these problems are routinely solved by humans with a reasonable mathematical education, their automation exposes several deep challenges. For MWPs, a capable solver must (1)~associate quantities with the entities they modify; (2)~handle the ambiguity of natural language, including chronological and temporal relations; (3)~recognize when information is irrelevant or missing; and (4)~generalize across problem structures that differ substantially from its training distribution~\cite{patel2021nlp,mirzadeh2024gsm}. In the problem of Table~\ref{tab:tab1}, for example, a solver must link the quantity 69 to its price attribute \$23, derive the residual count $420 - 69$ before introducing the second price attribute \$17, and correctly order the operations.

For geometry, the additional challenges are (1)~diagram parsing, \textit{i.e.}\ extracting the relative configuration of points, edges, and vertices; (2)~cross-modal reference resolution, since the textual description frequently leaves implicit relationships expressed only in the diagram; and (3)~theorem retrieval and application. To solve Figure~\ref{fig:fig1}, a system must parse a right triangle from the diagram, identify $AD=3$, $BD=14$, and either apply a trigonometric identity or the altitude-on-hypotenuse theorem to recover $CD$.

For formal theorem proving and autoformalization, the difficulty is compounded by the scarcity of parallel informal/formal corpora, the compositional nature of proof tactics, and the sheer size of modern libraries such as \texttt{mathlib}~\cite{yang2023leandojo}. For open-ended discovery, finally, the challenge is conceptual: how does one design a search procedure capable of producing genuinely novel mathematical constructions rather than rehashing examples from the training set?

\subsection{Scope and Contributions}
Earlier surveys of this domain~\cite{mukherjee2008review,10.1007/978-981-10-7590-2_7,zhang2020gap,sundaram2022nlp} focus predominantly on pre-LLM work on arithmetic MWPs. Two more recent surveys~\cite{lu2023survey,ahn2024large} begin to cover LLM-based methods but stop short of the reasoning-model era inaugurated by OpenAI~o1~\cite{openai2024o1} in late~2024. Very recent ACM Computing Surveys articles provide useful LLM-centric lenses: \cite{liu2025mathematicallm} organizes mathematical language models by tasks, methods, and datasets, while \cite{wang2025survey} frames LLM mathematical reasoning as the interaction of mathematical comprehension and answer generation. The 2025 ACL Findings survey of multimodal mathematical reasoning~\cite{yan2025multimodalsurvey} offers a valuable benchmark--method--challenge taxonomy for MLLMs, but naturally concentrates on visual and multimodal settings. In parallel, LLM-based multi-agent surveys~\cite{guo2024multiagentsurvey} analyze agent profiling, communication, and capability growth across domains, leaving room for a math-specific synthesis of debate, verification, and proof-oriented collaboration. A very recent survey on formal mathematical reasoning~\cite{yang2024formal} focuses narrowly on the formal side, and a parallel survey on deep learning for theorem proving~\cite{li2024dl4tp} covers that sub-area in depth. Table~\ref{tab:survey_positioning} summarizes how these adjacent surveys motivate our scope. Our goal is to provide an integrated perspective that (i)~preserves the historical development of MWP and geometry solvers for continuity with the earlier literature, (ii)~gives a current account of LLM-based mathematical reasoning including chain-of-thought, tool use, multi-agent collaboration, and test-time scaling, (iii)~covers autoformalization and proof search in Lean~4 in parallel with the informal track, and (iv)~reflects on the emerging role of AI in actual research mathematics, informed by the public writings and interviews of mathematicians such as Terence Tao~\cite{tao2024copilot,tao2026primetime,tao2026methods}.

\begin{table*}[t]
\centering
\scriptsize
\renewcommand{\arraystretch}{1.24}
\setlength{\tabcolsep}{3pt}
\providecommand{\full}{\textcolor{Indigo}{\ding{108}}}
\providecommand{\part}{\textcolor{LavenderTwo!92!black}{\ding{108}}}
\providecommand{\none}{\textcolor{black!22}{\ding{109}}}
\providecommand{\axhd}[1]{\rotatebox[origin=lB]{52}{\fontsize{6.6}{7.4} #1}}
\begin{tabularx}{\textwidth}{@{}p{0.155\textwidth}*{8}{>{\centering\arraybackslash}p{0.018\textwidth}}p{0.22\textwidth}>{\raggedright\arraybackslash}X@{}}
\tabletop
\thead{Survey} & \axhd{MWP} & \axhd{Geo/MM} & \axhd{LLM} & \axhd{Reas.} & \axhd{MAgt} & \axhd{Formal} & \axhd{Disc.} & \axhd{Mlng.} & \thead{Distinctive strength} & \thead{What this survey adds beyond it} \\ \tablemid
\tgroup{11}{Pre-LLM MWP surveys}
Zhang et al.~\cite{zhang2020gap} & \full & \part & \none & \none & \none & \none & \none & \none & First comprehensive MWP survey; semantic-gap taxonomy             & LLM, formal, multimodal, and discovery axes; reasoning-model era \\
Sundaram et al.~\cite{sundaram2022nlp}                         & \full & \none & \none & \none & \none & \none & \none & \none & Deep-learning MWP methods: \textsc{Seq2Seq}, trees, graph-based   & Prompting, RL, formal, post-2023 LLM era, multimodal, discovery \\
\tgroup{11}{Broad reasoning surveys}
Lu et al.~\cite{lu2023survey}                                  & \full & \full & \part & \none & \none & \part & \none & \none & Task taxonomy across MWPs, geometry, proving, and math QA         & Reasoning models, RLVR, recent provers, discovery, supervision ladder \\
Ahn et al.~\cite{ahn2024large}                                 & \part & \none & \full & \none & \none & \none & \none & \none & Concise account of LLM math progress and open challenges          & Neuro-symbolic geometry, Lean-based proving, post-2024 systems \\
\tgroup{11}{LLM-centric surveys}
Liu et al.~\cite{liu2025mathematicallm}                        & \part & \part & \full & \part & \part & \part & \none & \part & Task / method / dataset map, incl.\ augmented data                 & Classical-MWP history, verification pipelines, open-problem discovery \\
Wang et al.~\cite{wang2025survey}                              & \part & \none & \full & \full & \part & \none & \none & \none & Comprehension--generation framing; long CoT, RL, test-time scaling & Formal proving and discovery as co-equal axes; CGV triad \\
\tgroup{11}{Domain-specific surveys}
Yan et al.~\cite{yan2025multimodalsurvey}                       & \none & \full & \part & \part & \none & \none & \none & \none & Post-2021 map of MLLM benchmarks, methods, failure modes          & MWP lineage; 2026 formal--informal--discovery convergence \\
Guo et al.~\cite{guo2024multiagentsurvey}                       & \none & \none & \part & \none & \full & \none & \none & \none & Agent profiles, communication, and coordination mechanisms        & Math-specific synthesis of debate, verification, proof collaboration \\
Yang et al.~\cite{yang2024formal}; Li et al.~\cite{li2024dl4tp} & \none & \none & \part & \none & \none & \full & \none & \none & Deep coverage of proof assistants, tactics, and formal benchmarks  & Integrates informal MWPs, multimodal geometry, contamination debates \\
\midrule[0.4pt]
\timportant \textbf{This survey} & \full & \full & \full & \full & \full & \full & \full & \full & \textbf{Unifies informal, multimodal, formal, and discovery-oriented systems through 2026} & \textbf{Positions mathematical reasoning as a progression from answer prediction to verified discovery} \\
\tablebottom
\end{tabularx}
\caption{Positioning of the present survey against the recent literature on AI for mathematical reasoning. The eight middle columns form a coverage matrix over the axes this survey treats as co-equal: math word problems (\textbf{MWP}), geometry and multimodal reasoning (\textbf{Geo/MM}), prompted / tool-augmented LLMs (\textbf{LLM}), reasoning-model era (\textbf{Reas.}), multi-agent systems (\textbf{MAgt}), formal theorem proving (\textbf{Formal}), mathematical discovery (\textbf{Disc.}), and multilingual evaluation (\textbf{Mlng.}). \textbf{Symbols:} \full\,=\,comprehensive coverage; \part\,=\,partial or peripheral coverage; \none\,=\,not in scope. Violet sub-headers group surveys by editorial focus. The magenta-shaded final row marks the present survey, whose distinguishing feature is being the only entry in the table to fill every axis: by design, the paper sits at the intersection rather than within any single sub-literature.}
\label{tab:survey_positioning}
\end{table*}

The contributions of this survey are as follows.
\begin{enumerate}
    \item We chronologically trace the methodology stack for mathematical reasoning, from rule-based and statistical MWP solvers of the 1980s--2010s to the large-scale reasoning and proving systems of 2024--2026 (Figure~\ref{fig:timeline}).
    \item We introduce the supervision-ladder framework, organizing the field's progression from hand-coded schemata through formal proof-assistant kernels as a sequence of increasingly informative external verifiers, and propose an updated taxonomy of mathematical reasoning systems that integrates formal, informal, multimodal, multi-agent, and discovery-oriented approaches (Figure~\ref{fig:classification}).
    \item We compile a comprehensive catalog of benchmarks, both classical (\textsc{AI2}, \textsc{MAWPS}, \textsc{Math23K}) and contemporary (\textsc{ParaMAWPS}, \textsc{GSM8K}, \textsc{MATH}, \textsc{MGSM}, \textsc{HRM8K}, \textsc{PatiGonit}, \textsc{PGPS9K}, \textsc{MathVista}, \textsc{OlympiadBench}, \textsc{MiniF2F}, \textsc{ProofNet}, \textsc{PutnamBench}, \textsc{FrontierMath}, \textsc{HLE}, \textsc{LiveBench}), with particular attention to multilinguality, saturation, and contamination dynamics.
    \item We report and synthesize performance results across these benchmarks, documenting the shift from \textless~70\% MWP accuracy in 2018 to effectively saturated grade-school arithmetic and Olympiad geometry by 2025.
    \item We provide a measured discussion of failure modes, drawing on probing studies such as \textsc{SVAMP}~\cite{patel2021nlp} and \textsc{GSM-Symbolic}~\cite{mirzadeh2024gsm}, and on recent assessments by working mathematicians.
    \item We articulate future directions, highlighting the human--AI collaborative workflows emerging around Lean~4, the division of labor between discovery and verification, and the open problems that remain stubbornly out of reach.
\end{enumerate}

\paragraph{What this survey does not cover} In return for the breadth above, we deliberately set aside several adjacent topics. \emph{First}, we exclude general code generation and programming-language benchmarks (\textsc{HumanEval}, \textsc{MBPP}, \textsc{SWE-Bench}) unless their problems are explicitly mathematical; surveys such as~\cite{liu2025mathematicallm} already treat the code-reasoning axis in depth. \emph{Second}, we do not cover end-to-end mathematics education and intelligent tutoring systems, a sizeable literature with its own pedagogical desiderata, restricting our attention instead to solver capabilities that an educational system might \emph{use}. \emph{Third}, scientific reasoning at large (chemistry, physics, biology, and engineering QA suites such as \textsc{GPQA}, \textsc{SciBench}, and \textsc{SciEval}) is referenced only where it intersects mathematical reasoning, since dedicated surveys of that frontier are already emerging. \emph{Fourth}, we touch on but do not systematically review hardware, inference-system, or serving-stack optimizations for long-horizon reasoning, which lie outside our methodological remit. \emph{Finally}, while we do report results from proprietary frontier systems (OpenAI~o-series, Gemini~Deep~Think, Claude reasoning models) for completeness, we do not attempt independent replication, and we flag throughout where reported numbers are vendor-supplied rather than independently audited.

\subsection{Survey Methodology}\label{sec:methodology}
We followed a structured literature-review process adapted from established SLR guidelines~\cite{kitchenham2007guidelines}. The goal was not a statistical meta-analysis, since evaluation protocols differ sharply across MWPs, multimodal geometry, informal LLM reasoning, and formal proof assistants. Instead, we constructed a systematic map of the field and used the map to organize the narrative, tables, and taxonomy.

\subsubsection{Research Questions}
The review is guided by five questions. RQ1 asks how informal mathematical reasoning evolved from rule-based MWP solvers to RL-trained reasoning models. RQ2 asks how multimodal systems use diagrams, figures, and visual tokens as mathematical evidence. RQ3 asks how formal theorem proving and autoformalization have progressed toward Olympiad- and research-level mathematics. RQ4 asks what counts as genuine AI-assisted mathematical discovery, as distinct from rediscovery or literature retrieval. RQ5 asks how benchmarks should be designed, maintained, and reported to resist saturation, contamination, and metric mismatch across languages and modalities.

\subsubsection{Search, Screening, and Coding}
We searched Scopus, Web of Science, ACL Anthology, IEEE Xplore, and arXiv (cs.CL, cs.AI, cs.LG, and cs.SC) between January 2023 and April~2026. Search strings combined task terms (\textit{e.g.}, ``math word problem,'' ``geometry solving,'' ``theorem proving,'' ``autoformalization,'' ``mathematical discovery''), method terms (\textit{e.g.}, ``large language model,'' ``chain-of-thought,'' ``reinforcement learning,'' ``multi-agent,'' ``neuro-symbolic''), and artifact terms (\textit{e.g.}, ``Lean,'' ``proof assistant,'' ``benchmark,'' ``process reward model''). We supplemented database search with backward and forward snowballing from recent surveys~\cite{lu2023survey,ahn2024large,liu2025mathematicallm,wang2025survey,yan2025multimodalsurvey,guo2024multiagentsurvey,yang2024formal,li2024dl4tp,sundaram2022nlp,mukherjee2008review,zhang2020gap}, targeted inclusion of seminal pre-LLM papers for historical continuity, and arXiv monitoring through April~2026 for late-breaking systems.

The initial search yielded 1{,}847 candidate records. After removing 312 duplicates, 1{,}535 records remained for title and abstract screening. We excluded 1{,}089 records outside the mathematical-reasoning scope, primarily general NLP, pure code generation, education technology, or scientific QA without an explicit mathematical-reasoning component. Full-text review of the remaining 446 records excluded a further 208 for scope mismatch, insufficient technical detail, superseded versions, or lack of quantitative evaluation. The final corpus contains 238 included records.

\begin{table}[t]
\centering
\scriptsize
\renewcommand{\arraystretch}{1.18}
\setlength{\tabcolsep}{3pt}
\begin{tabularx}{\columnwidth}{@{}>{\centering\arraybackslash}p{0.05\columnwidth}>{\centering\arraybackslash}p{0.065\columnwidth}X@{}}
\tabletop
\thead{} & \thead{ID} & \thead{Criterion} \\ \tablemid
\multirow[c]{5}{0.075\columnwidth}{\centering\rotatebox[origin=c]{90}{\textbf{Inclusion\qquad\qquad}}}
 & \textbf{I1} & Proposes, evaluates, or analyzes a method, model, dataset, benchmark, or evaluation protocol for mathematical reasoning \\
 & \textbf{I2} & Covers at least one focus area: MWPs, geometry or multimodal math, theorem proving, autoformalization, discovery, or benchmark design \\
 & \textbf{I3} & Appears in a peer-reviewed venue, workshop, technical report, or arXiv preprint with public artifacts, substantial uptake, or clear relevance to the 2024--2026 frontier \\
 & \textbf{I4} & Reports quantitative results, introduces a dataset or benchmark, or provides a survey/theoretical framework central to the taxonomy \\
 & \textbf{I5} & Is publicly accessible by April~2026; older seminal works are retained when needed for historical continuity \\ \tablemid
\multirow[c]{4}{0.075\columnwidth}{\centering\rotatebox[origin=c]{90}{\textbf{Exclusion\qquad}}}
 & \textbf{E1} & Addresses general NLP, code generation, scientific QA, or education without a specific mathematical-reasoning component \\
 & \textbf{E2} & Duplicates or is superseded by a more complete version of the same work \\
 & \textbf{E3} & Provides only abstracts, slides, editorials, or non-archival commentary, except official system reports needed to document frontier model releases \\
 & \textbf{E4} & Lacks accessible full text or sufficient methodological detail for extraction \\
\tablebottom
\end{tabularx}
\caption{Eligibility criteria used during screening and full-text review. The arXiv allowance reflects the rapid pace of AI-for-mathematics research: several influential systems first appeared as preprints or official technical reports before journal or conference publication.}
\label{tab:inclusion_exclusion}
\end{table}

\begin{figure*}[t]
\centering
\includegraphics[width=\textwidth]{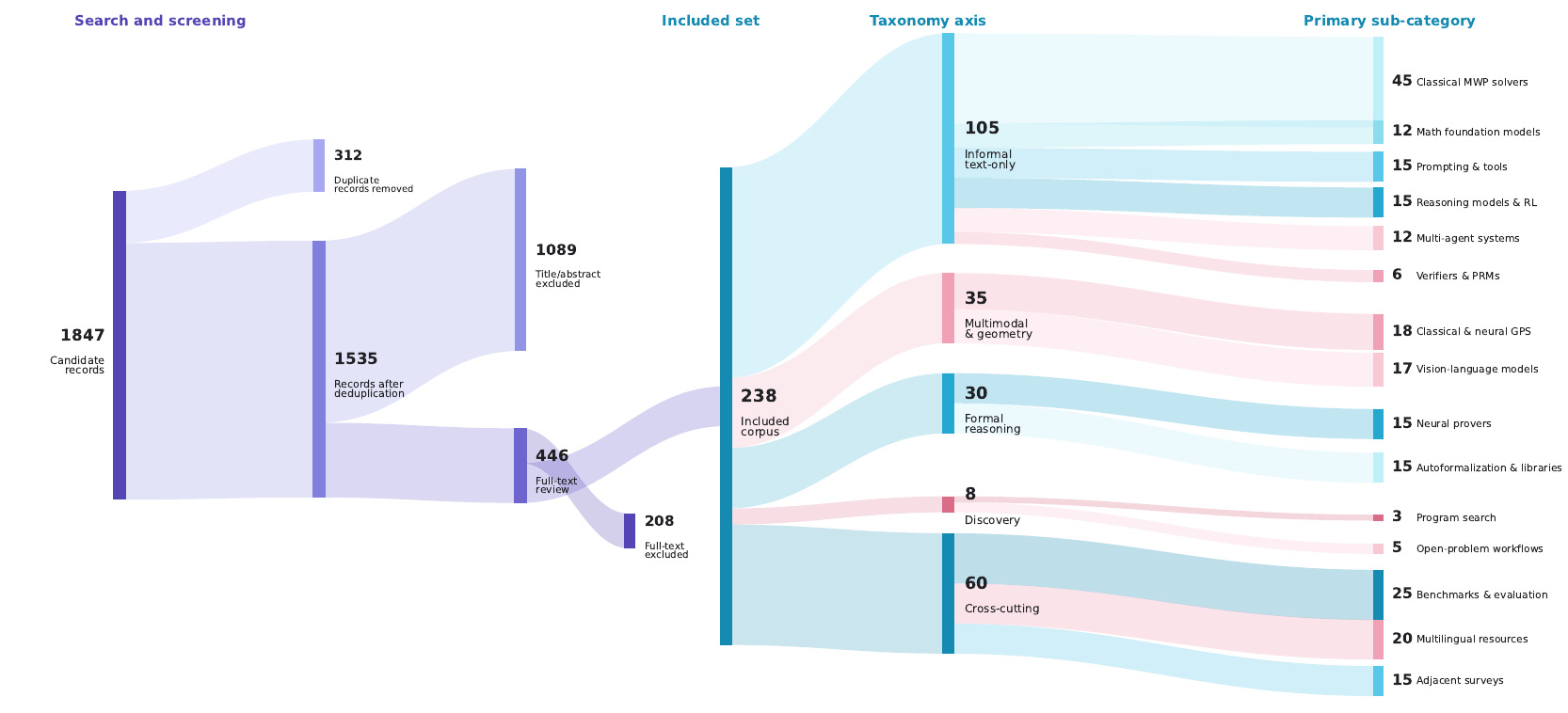}
\caption{Search, screening, and taxonomy coding pipeline for the survey corpus. The left side shows the record-selection flow from 1{,}847 candidate records to 238 included records after duplicate removal, title/abstract screening, and full-text review. The right side assigns each included record to one primary taxonomy category for counting: informal text-only reasoning dominates because of the long MWP lineage, while discovery remains small in volume but unusually consequential for the field's frontier. Cross-cutting records include benchmark, multilingual-resource, and adjacent-survey papers used to structure the dataset and evaluation discussion.}
\label{fig:sankey}
\end{figure*}

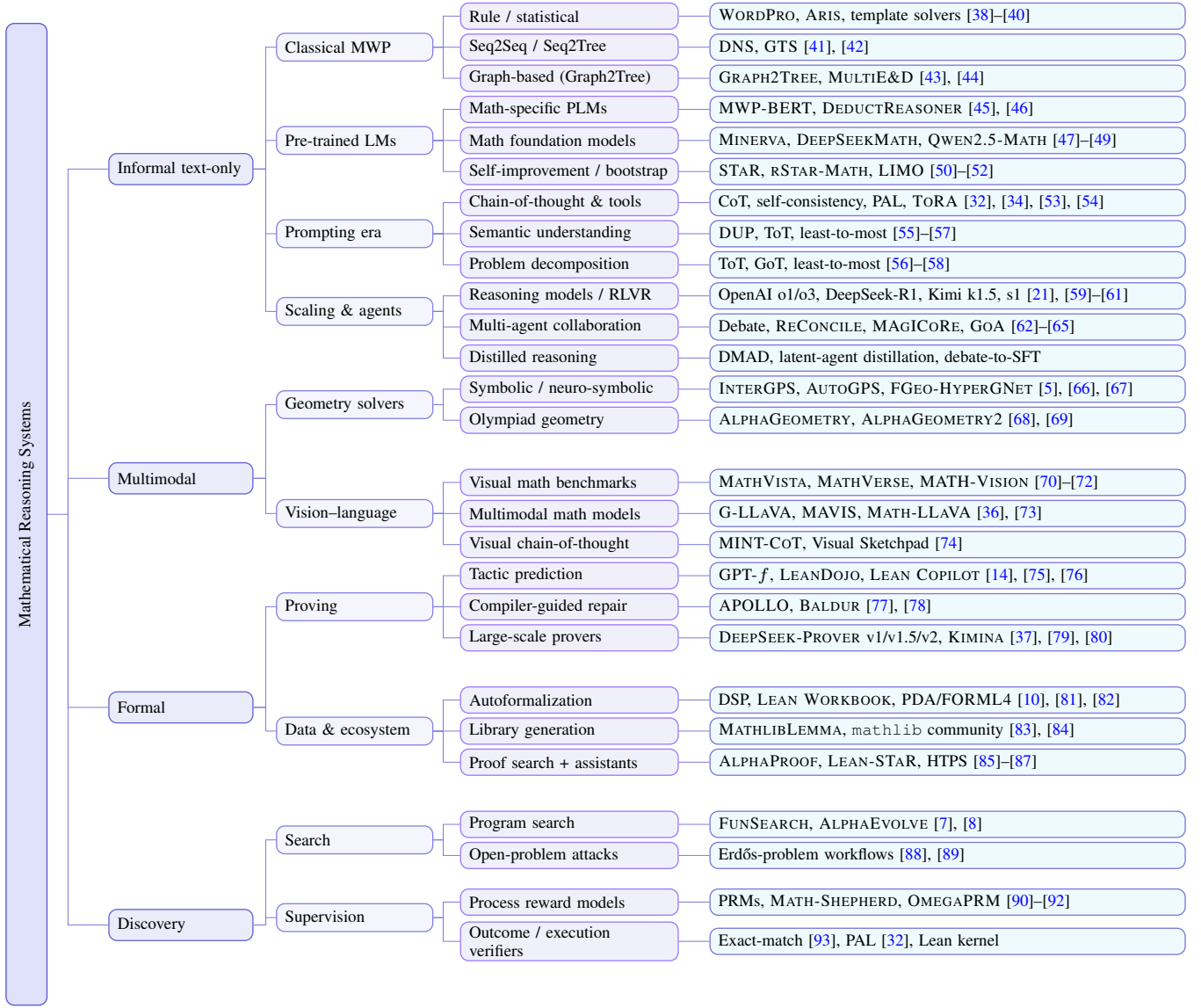
\begin{figure*}[t]
\centering
\resizebox{0.98\textwidth}{!}{%
\begin{tikzpicture}[
  taxfont/.style={font=\fontsize{5.8}{6.6}\selectfont},
  root/.style={taxfont, rectangle, draw=Indigo!75, rounded corners=1mm,
    fill=LavenderOne!35, minimum width=.50cm, minimum height=12.0cm,
    align=center, inner sep=1.5pt},
  group/.style={taxfont, rectangle, draw=Indigo!70, rounded corners=1mm,
    fill=LavenderTwo!18, text width=1.55cm, minimum height=.38cm,
    align=left, inner xsep=3pt, inner ysep=2pt},
  subgroup/.style={taxfont, rectangle, draw=Indigo!55, rounded corners=1mm,
    fill=LavenderOne!12, text width=1.70cm, minimum height=.34cm,
    align=left, inner xsep=3pt, inner ysep=1.6pt},
  method/.style={taxfont, rectangle, draw=Violet!75, rounded corners=1mm,
    fill=LavenderThree!12, text width=2.45cm, minimum height=.32cm,
    align=left, inner xsep=3pt, inner ysep=1.4pt},
  example/.style={taxfont, rectangle, draw=Indigo!60, rounded corners=1mm,
    fill=Cyan!12, text width=5.6cm, minimum height=.32cm,
    align=left, inner xsep=3pt, inner ysep=1.4pt},
  branch/.style={draw=Indigo!58, line width=.30pt}
]
\def\colRoot{0}
\def\colGroup{1.00}
\def\colSub{3.05}
\def\colMethod{5.30}
\def\colExample{8.35}

\node[root] (root) at (\colRoot,-5.90) {\rotatebox{90}{Mathematical Reasoning Systems}};

\node[group, anchor=west] (g1) at (\colGroup,-1.68)  {Informal text-only};
\node[group, anchor=west] (g2) at (\colGroup,-5.46)  {Multimodal};
\node[group, anchor=west] (g3) at (\colGroup,-8.26)  {Formal};
\node[group, anchor=west] (g4) at (\colGroup,-10.92) {Discovery};

\node[subgroup, anchor=west] (s1) at (\colSub,-0.19) {Classical MWP};
\node[subgroup, anchor=west] (s2) at (\colSub,-1.33) {Pre-trained LMs};
\node[subgroup, anchor=west] (s3) at (\colSub,-2.47) {Prompting era};
\node[subgroup, anchor=west] (s4) at (\colSub,-3.42) {Scaling \& agents};
\node[subgroup, anchor=west] (s5) at (\colSub,-4.56) {Geometry solvers};
\node[subgroup, anchor=west] (s6) at (\colSub,-5.89) {Vision--language};
\node[subgroup, anchor=west] (s7) at (\colSub,-7.03) {Proving};
\node[subgroup, anchor=west] (s8) at (\colSub,-8.55) {Data \& ecosystem};
\node[subgroup, anchor=west] (s9)  at (\colSub,-9.88) {Search};
\node[subgroup, anchor=west] (s10) at (\colSub,-10.83) {Supervision};

\node[method, anchor=west] (m1)  at (\colMethod,0.19)  {Rule / statistical};
\node[method, anchor=west] (m2)  at (\colMethod,-0.19) {Seq2Seq / Seq2Tree};
\node[method, anchor=west] (m3)  at (\colMethod,-0.57) {Graph-based (Graph2Tree)};
\node[method, anchor=west] (m4)  at (\colMethod,-0.95) {Math-specific PLMs};
\node[method, anchor=west] (m5)  at (\colMethod,-1.33) {Math foundation models};
\node[method, anchor=west] (m6)  at (\colMethod,-1.71) {Self-improvement / bootstrap};
\node[method, anchor=west] (m7)  at (\colMethod,-2.09) {Chain-of-thought \& tools};
\node[method, anchor=west] (m8)  at (\colMethod,-2.47) {Semantic understanding};
\node[method, anchor=west] (m9)  at (\colMethod,-2.85) {Problem decomposition};
\node[method, anchor=west] (m10) at (\colMethod,-3.23) {Reasoning models / RLVR};
\node[method, anchor=west] (m11) at (\colMethod,-3.61) {Multi-agent collaboration};
\node[method, anchor=west] (m12) at (\colMethod,-3.99) {Distilled reasoning};
\node[method, anchor=west] (m13) at (\colMethod,-4.37) {Symbolic / neuro-symbolic};
\node[method, anchor=west] (m14) at (\colMethod,-4.75) {Olympiad geometry};
\node[method, anchor=west] (m15) at (\colMethod,-5.51) {Visual math benchmarks};
\node[method, anchor=west] (m16) at (\colMethod,-5.89) {Multimodal math models};
\node[method, anchor=west] (m17) at (\colMethod,-6.27) {Visual chain-of-thought};
\node[method, anchor=west] (m18) at (\colMethod,-6.65) {Tactic prediction};
\node[method, anchor=west] (m19) at (\colMethod,-7.03) {Compiler-guided repair};
\node[method, anchor=west] (m20) at (\colMethod,-7.41) {Large-scale provers};
\node[method, anchor=west] (m21) at (\colMethod,-8.17) {Autoformalization};
\node[method, anchor=west] (m22) at (\colMethod,-8.55) {Library generation};
\node[method, anchor=west] (m23) at (\colMethod,-8.93) {Proof search + assistants};
\node[method, anchor=west] (m24) at (\colMethod,-9.69) {Program search};
\node[method, anchor=west] (m25) at (\colMethod,-10.07){Open-problem attacks};
\node[method, anchor=west] (m26) at (\colMethod,-10.64){Process reward models};
\node[method, anchor=west] (m27) at (\colMethod,-11.12){Outcome / execution verifiers};

\node[example, anchor=west] (e1)  at (\colExample,0.19)  {\textsc{WordPro}, \textsc{Aris}, template solvers~\cite{fletcher1985understanding,hosseini2014learning,kushman2014learning}};
\node[example, anchor=west] (e2)  at (\colExample,-0.19) {\textsc{DNS}, \textsc{GTS}~\cite{wang2017deep,xie2019goal}};
\node[example, anchor=west] (e3)  at (\colExample,-0.57) {\textsc{Graph2Tree}, \textsc{MultiE\&D}~\cite{zhang2020graph,shen2021generate}};
\node[example, anchor=west] (e4)  at (\colExample,-0.95) {\textsc{MWP-BERT}, \textsc{DeductReasoner}~\cite{liang2021mwp,jie2022deductive}};
\node[example, anchor=west] (e5)  at (\colExample,-1.33) {\textsc{Minerva}, \textsc{DeepSeekMath}, \textsc{Qwen2.5-Math}~\cite{lewkowycz2022solving,shao2024deepseekmath,yang2024qwen2math}};
\node[example, anchor=west] (e6)  at (\colExample,-1.71) {\textsc{STaR}, \textsc{rStar-Math}, \textsc{LIMO}~\cite{zelikman2022star,guan2025rstarmath,ye2025limo}};
\node[example, anchor=west] (e7)  at (\colExample,-2.09) {CoT, self-consistency, PAL, \textsc{ToRA}~\cite{wei2022chain,wang2023selfconsistency,gao2023pal,gou2024tora}};
\node[example, anchor=west] (e8)  at (\colExample,-2.47) {\textsc{DUP}, ToT, least-to-most~\cite{zhong2024dup,yao2023tree,zhou2023leasttomost}};
\node[example, anchor=west] (e9)  at (\colExample,-2.85) {ToT, GoT, least-to-most~\cite{yao2023tree,besta2024graph,zhou2023leasttomost}};
\node[example, anchor=west] (e10) at (\colExample,-3.23) {OpenAI o1/o3, DeepSeek-R1, Kimi k1.5, s1~\cite{openai2024o1,guo2025deepseekr1,team2025kimi,muennighoff2025s1}};
\node[example, anchor=west] (e11) at (\colExample,-3.61) {Debate, \textsc{ReConcile}, \textsc{MAgICoRe}, \textsc{GoA}~\cite{du2023multiagentdebate,chen2024reconcile,chen2025magicore,yun2026goa}};
\node[example, anchor=west] (e12) at (\colExample,-3.99) {DMAD, latent-agent distillation, debate-to-SFT};
\node[example, anchor=west] (e13) at (\colExample,-4.37) {\textsc{InterGPS}, \textsc{AutoGPS}, \textsc{FGeo-HyperGNet}~\cite{lu2021inter,ping2025autogps,zhang2025fgeohypergnet}};
\node[example, anchor=west] (e14) at (\colExample,-4.75) {\textsc{AlphaGeometry}, \textsc{AlphaGeometry2}~\cite{trinh2024alphageometry,chervonyi2025alphageometry2}};
\node[example, anchor=west] (e15) at (\colExample,-5.51) {\textsc{MathVista}, \textsc{MathVerse}, \textsc{MATH-Vision}~\cite{lu2024mathvista,zhang2024mathverse,wang2024mathvision}};
\node[example, anchor=west] (e16) at (\colExample,-5.89) {\textsc{G-LLaVA}, \textsc{MAVIS}, \textsc{Math-LLaVA}~\cite{gao2023gllava,zhang2024mavis}};
\node[example, anchor=west] (e17) at (\colExample,-6.27) {\textsc{MINT-CoT}, Visual Sketchpad~\cite{chen2025mintcot}};
\node[example, anchor=west] (e18) at (\colExample,-6.65) {\textsc{GPT-$f$}, \textsc{LeanDojo}, \textsc{Lean Copilot}~\cite{polu2020generative,yang2023leandojo,song2024leancopilot}};
\node[example, anchor=west] (e19) at (\colExample,-7.03) {\textsc{APOLLO}, \textsc{Baldur}~\cite{ospanov2026apollo,first2023baldur}};
\node[example, anchor=west] (e20) at (\colExample,-7.41) {\textsc{DeepSeek-Prover} v1/v1.5/v2, \textsc{Kimina}~\cite{xin2024deepseekprover,ren2025deepseekproverv2,wang2025kimina}};
\node[example, anchor=west] (e21) at (\colExample,-8.17) {DSP, \textsc{Lean Workbook}, \textsc{PDA}/\textsc{FORML4}~\cite{jiang2023draft,ying2024leanworkbook,lu2024process}};
\node[example, anchor=west] (e22) at (\colExample,-8.55) {\textsc{MathlibLemma}, \texttt{mathlib} community~\cite{liu2026mathliblemma,mathlib2020}};
\node[example, anchor=west] (e23) at (\colExample,-8.93) {\textsc{AlphaProof}, \textsc{Lean-STaR}, HTPS~\cite{hubert2025alphaproof,lin2025leanstar,lample2022hypertree}};
\node[example, anchor=west] (e24) at (\colExample,-9.69) {\textsc{FunSearch}, \textsc{AlphaEvolve}~\cite{romera2024funsearch,novikov2025alphaevolve}};
\node[example, anchor=west] (e25) at (\colExample,-10.07){Erd\H{o}s-problem workflows~\cite{bloom2024erdos,alexeev2025erdos}};
\node[example, anchor=west] (e26) at (\colExample,-10.64){PRMs, \textsc{Math-Shepherd}, \textsc{OmegaPRM}~\cite{lightman2023verify,wang2024mathshepherd,luo2024improve}};
\node[example, anchor=west] (e27) at (\colExample,-11.12){Exact-match~\cite{cobbe2021training}, PAL~\cite{gao2023pal}, Lean kernel};

\draw[branch] (root.east) -- ++(.25,0);
\foreach \g in {g1,g2,g3,g4} {
  \draw[branch] ([xshift=.25cm]root.east) |- (\g.west);
}
\foreach \s in {s1,s2,s3,s4}   {\draw[branch] (g1.east) -- ++(.15,0) |- (\s.west);}
\foreach \s in {s5,s6}         {\draw[branch] (g2.east) -- ++(.15,0) |- (\s.west);}
\foreach \s in {s7,s8}         {\draw[branch] (g3.east) -- ++(.15,0) |- (\s.west);}
\foreach \s in {s9,s10}        {\draw[branch] (g4.east) -- ++(.15,0) |- (\s.west);}
\foreach \m in {m1,m2,m3}      {\draw[branch] (s1.east) -- ++(.12,0) |- (\m.west);}
\foreach \m in {m4,m5,m6}      {\draw[branch] (s2.east) -- ++(.12,0) |- (\m.west);}
\foreach \m in {m7,m8,m9}      {\draw[branch] (s3.east) -- ++(.12,0) |- (\m.west);}
\foreach \m in {m10,m11,m12}   {\draw[branch] (s4.east) -- ++(.12,0) |- (\m.west);}
\foreach \m in {m13,m14}       {\draw[branch] (s5.east) -- ++(.12,0) |- (\m.west);}
\foreach \m in {m15,m16,m17}   {\draw[branch] (s6.east) -- ++(.12,0) |- (\m.west);}
\foreach \m in {m18,m19,m20}   {\draw[branch] (s7.east) -- ++(.12,0) |- (\m.west);}
\foreach \m in {m21,m22,m23}   {\draw[branch] (s8.east) -- ++(.12,0) |- (\m.west);}
\foreach \m in {m24,m25}       {\draw[branch] (s9.east) -- ++(.12,0) |- (\m.west);}
\foreach \m in {m26,m27}       {\draw[branch] (s10.east) -- ++(.12,0) |- (\m.west);}
\foreach \i in {1,...,27} {
  \draw[branch] (m\i.east) -- (e\i.west);
}
\end{tikzpicture}%
}
\caption{A taxonomy of mathematical reasoning systems. The four main axes (informal text-only, multimodal, formal, discovery) are subdivided into thematic clusters and specific method families, with representative systems cited for each leaf. The tree reads left-to-right: root $\to$ axis $\to$ subfield $\to$ method $\to$ examples.}
\label{fig:classification}
\end{figure*}

Each included record was coded by primary contribution, task family, method family, modality, supervision signal, evaluation benchmark, and reported performance. For visual clarity, Figure~\ref{fig:sankey} assigns each record to one primary Sankey category; papers with multiple roles were noted during extraction and cross-referenced in the relevant sections, but not double-counted in the figure. This keeps the corpus distribution interpretable while preserving the paper's integrated treatment of systems that span categories.

The remainder of the paper is organized as follows. Section~\ref{sec:problem} formalizes the canonical tasks. Section~\ref{sec:mwp} surveys MWP solvers from rule-based methods through the modern pre-training era. Section~\ref{sec:llm} examines the LLM and reasoning-model era. Section~\ref{sec:multimodal} covers multimodal and geometry systems. Section~\ref{sec:formal} treats formal theorem proving and autoformalization. Section~\ref{sec:discovery} discusses mathematical discovery and its engagement with open problems. Section~\ref{sec:datasets} catalogs benchmarks and performance. Section~\ref{sec:synthesis} synthesizes the methodological lessons shared across these tracks. Section~\ref{sec:failures} reviews failure modes and critiques. Section~\ref{sec:future} outlines future directions. Section~\ref{sec:conclusion} concludes.

\section{Problem Formulation}\label{sec:problem}
We formalize the four canonical tasks treated in this survey.

\textbf{Math Word Problem.} An MWP instance is a sequence $P$ of word tokens $V_P = \{v_1,\dots,v_m\}$ and numeric values $n_P = \{n_1,\dots,n_l\}$, the former comprising entities (names, objects, units, rates) and the latter the numerical amounts relevant to those entities. The goal of an MWP solver is to map $P$ to a valid mathematical expression $E_P$ composed from $n_P$, a set of auxiliary constants $\mathcal{C}$ (\textit{e.g.}\ $\pi$), and operators $O = \{+,-,\times,\div,\dots\}$, such that $\llbracket E_P\rrbracket$ equals the reference answer. In many modern systems, $E_P$ is replaced by a Python program~\cite{gao2023pal,chen2022program} or a sequence of tool calls~\cite{gou2024tora}; the evaluation is delegated to a deterministic interpreter.

\textbf{Geometry Problem.} A geometry instance is a tuple $\langle t, d, \mathbf{c}\rangle$, where $t$ is the problem text, $d$ is a diagram image, and $\mathbf{c} = \{c_0, \dots, c_k\}$ is either a set of multiple-choice numerical candidates or a free-form numerical answer space. The solver must predict $c_i \in \mathbf{c}$. In more recent work, the solver may also be required to return a structured proof or an interpretable program~\cite{chen2021geoqa,gao2023gllava}.

\textbf{Autoformalization and Formal Theorem Proving.} Given a natural language statement $s_{\text{NL}}$, the autoformalization task is to produce a formal statement $s_{\text{F}}$ in a target proof assistant language (Lean~4, Isabelle, Coq) that is type-checkable and faithful to $s_{\text{NL}}$~\cite{wu2022autoformalization}. The theorem-proving task, given $s_{\text{F}}$, is to produce a proof term $\pi$ such that the kernel of the proof assistant accepts $(s_{\text{F}}, \pi)$. Modern systems may interleave these two tasks~\cite{jiang2023draft,xin2024deepseekprover}.

\textbf{Mathematical Discovery.} Given an underspecified mathematical problem, for instance, to establish a new lower bound for a combinatorial quantity, or to produce an explicit counterexample, the system must output either a construction $\omega$ along with a verification certificate, or a program $p$ whose execution produces such a construction. Frameworks such as \textsc{FunSearch}~\cite{romera2024funsearch} and \textsc{AlphaEvolve}~\cite{novikov2025alphaevolve} instantiate this setting by having the LLM evolve a program that maximizes an automated fitness function.

\textbf{Boundary Tasks: Arithmetic Representation and Calculation.} Recent mathematical-LM surveys separate \emph{mathematical calculation} from higher-level reasoning~\cite{liu2025mathematicallm}. This category includes numerical representation, digit-level arithmetic, unit conversion, and expression evaluation. We treat these not as separate end goals but as enabling skills: failure at arithmetic representation can corrupt an otherwise correct MWP parse, while reliable calculation is often best delegated to tools once the model has inferred the correct expression. The literature covered in this survey has a cutoff of April 2026. Systems released or published after this date are not systematically reviewed.

\subsection{Outputs, Supervision, and Grading}
The four task families above differ not only in their input modality but also in the kind of artifact expected from the model. This distinction matters because mathematical reasoning systems improve fastest when the artifact can be automatically checked. A final numerical answer gives only weak supervision; an executable program permits deterministic evaluation; a formal proof supplies a kernel-checked certificate; and an open-ended construction requires a task-specific verifier. Table~\ref{tab:interfaces} summarizes this progression.

\begin{table*}[t]
\centering
\scriptsize
\renewcommand{\arraystretch}{1.32}
\setlength{\tabcolsep}{3.5pt}
\providecommand{\full}{\textcolor{Indigo}{\ding{108}}}
\providecommand{\none}{\textcolor{black!22}{\ding{109}}}
\providecommand{\stripe}[1]{\textcolor{#1}{\rule[-.45ex]{1.6pt}{1.7ex}}\hspace{2pt}}
\providecommand{\axhd}[1]{\rotatebox[origin=lB]{50}{\fontsize{6.6}{7.4}\selectfont #1}}
\begin{tabularx}{\textwidth}{@{}p{0.18\textwidth}>{\raggedright\arraybackslash}p{0.255\textwidth}*{5}{>{\centering\arraybackslash}p{0.022\textwidth}}>{\raggedright\arraybackslash}X@{}}
\tabletop
\thead{Task family} & \thead{Typical output} & \axhd{Ans.\ match} & \axhd{Code exec.} & \axhd{Symb.\ solver} & \axhd{Proof kernel} & \axhd{Human / expert} & \thead{Main supervision signal} \\ \tablemid
\stripe{VivCyan!75!black}MWP \& competition QA                   & Numeric answer, symbolic expression, or short rationale                         & \full & \none & \none & \none & \none & Worked solutions, CoT traces, outcome rewards \\
\stripe{VivOrange!75!black}Tool-integrated math                  & Python or calculator calls; mixed NL\,$+$\,program traces                       & \none & \full & \none & \none & \none & Program-of-thought demonstrations, execution feedback \\
\stripe{VivLilac!80!black}Geometry \& multimodal math            & Numeric answer, construction, diagram annotations, or proof sketch              & \full & \none & \full & \none & \full & Diagram labels, formal-language parses, theorem-application traces \\
\stripe{VivPink!75!black}Formal theorem proving                  & Lean / Isabelle / Coq statement $+$ proof term or tactic script                 & \none & \none & \none & \full & \none & Tactic traces, synthetic proof corpora, type-check and proof-state feedback \\
\timportant \stripe{VivPurple!85!black}Discovery \& open problems & Construction, counterexample, bound-improving program, or formalized theorem    & \none & \full & \none & \full & \full & Evolutionary fitness, verifier feedback, literature audit, formal certificates \\
\tablebottom
\end{tabularx}
\caption{Evaluation interfaces for mathematical reasoning. The five middle columns form a coverage matrix over the grader types each task family relies on: \textbf{Ans.\ match} (exact match, algebraic equivalence, or answer parser), \textbf{Code exec.} (Python interpreter, unit tests), \textbf{Symb.\ solver} (symbolic geometry or algebra engines), \textbf{Proof kernel} (Lean / Isabelle / Coq), and \textbf{Human / expert} review. \full\,=\,used; \none\,=\,not used. Colored stripes follow the paradigm palette of Figure~\ref{fig:timeline} and saturate from light to deep as the supervision constraint strengthens, mirroring the ladder argument of Section~\ref{sec:synthesis}. Geometry and the magenta-shaded Discovery row are the only task families whose verification requires three distinct grader types operating in combination, the visual signature of the multi-checker turn that the survey argues defines the 2024--2026 frontier.}
\label{tab:interfaces}
\end{table*}

This artifact-level view also clarifies why the field has repeatedly moved from unstructured text toward structured intermediates. Equation trees, operation programs, Python snippets, theorem-prover tactics, and Lean proof terms all reduce ambiguity at the cost of greater annotation or search burden. Much of the current research frontier can be interpreted as a search for the most useful intermediate representation: expressive enough to capture human mathematical intent, but formal enough for machines to check and learn from.


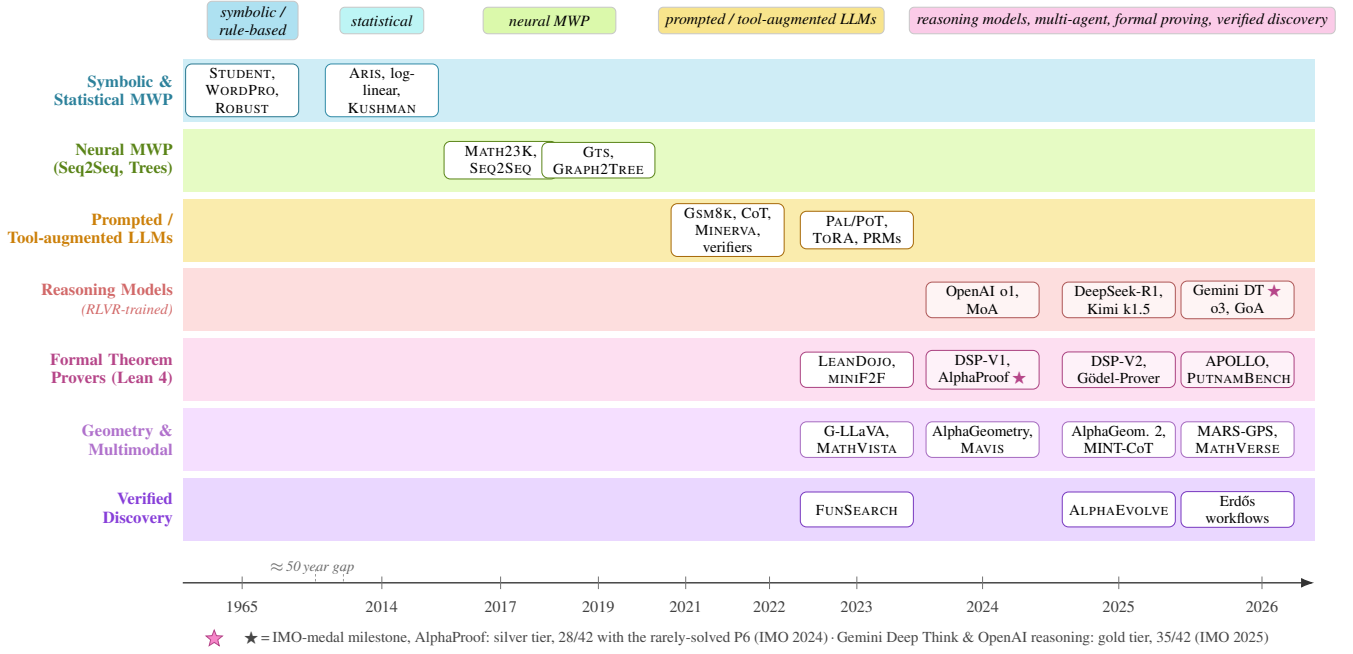
\begin{figure*}[t]
\centering
\resizebox{0.98\textwidth}{!}{%
\begin{tikzpicture}[
    laneTitle/.style={anchor=east, align=right, font=\bfseries\fontsize{6.8}{7.6}\selectfont,
                      text width=2.55cm, inner sep=1pt},
    yearlbl/.style={font=\fontsize{6.6}{7.4}\selectfont, anchor=north, text=black!75},
    era/.style={rectangle, rounded corners=.7mm, draw=black!22, inner xsep=3pt, inner ysep=2pt,
                minimum height=.38cm, align=center, font=\fontsize{6.4}{7.2}\selectfont\itshape},
    ev/.style={rectangle, rounded corners=.9mm, line width=.4pt, fill=white,
               minimum height=.50cm, align=center, inner xsep=2.4pt, inner ysep=1.6pt,
               text width=1.45cm, font=\fontsize{6.2}{7.0}\selectfont},
    axisline/.style={draw=black!80, line width=.55pt, -{Latex[length=1.9mm]}},
    starM/.style={star, star points=5, star point ratio=2.3, draw=VivPink!75!black,
                  fill=VivPink!75, line width=.3pt, minimum size=7pt, inner sep=0pt},
    legendlbl/.style={font=\fontsize{6.2}{7.0}\selectfont, anchor=west, text=black!72}
]
\def\xA{0.85}    
\def\xB{2.85}    
\def\xC{4.55}    
\def\xD{5.95}    
\def\xE{7.20}    
\def\xF{8.40}    
\def\xG{9.65}    
\def\xH{11.45}   
\def\xI{13.40}   
\def\xJ{15.45}   
\def\xMAX{16.20}

\def\yERA{9.30}
\def\yLa{8.30}   
\def\yLb{7.30}   
\def\yLc{6.30}   
\def\yLd{5.30}   
\def\yLe{4.30}   
\def\yLf{3.30}   
\def\yLg{2.30}   
\def\yAX{1.25}   
\def\yLEG{0.45}  

\fill[VivCyan!25]    (0,\yLa-.45) rectangle (\xMAX,\yLa+.45);
\fill[VivLime!28]    (0,\yLb-.45) rectangle (\xMAX,\yLb+.45);
\fill[VivYellow!32]  (0,\yLc-.45) rectangle (\xMAX,\yLc+.45);
\fill[VivSalmon!25]  (0,\yLd-.45) rectangle (\xMAX,\yLd+.45);
\fill[VivPink!20]    (0,\yLe-.45) rectangle (\xMAX,\yLe+.45);
\fill[VivLilac!28]   (0,\yLf-.45) rectangle (\xMAX,\yLf+.45);
\fill[VivPurple!22]  (0,\yLg-.45) rectangle (\xMAX,\yLg+.45);

\node[laneTitle, text=VivCyan!70!black]    at (-.10,\yLa) {Symbolic \&\\Statistical MWP};
\node[laneTitle, text=VivLime!55!black]    at (-.10,\yLb) {Neural MWP\\(Seq2Seq, Trees)};
\node[laneTitle, text=VivOrange!80!black]  at (-.10,\yLc) {Prompted /\\Tool-augmented LLMs};
\node[laneTitle, text=VivSalmon!85!black]  at (-.10,\yLd) {Reasoning Models\\\textnormal{\itshape\fontsize{6.0}{6.8}\selectfont(RLVR-trained)}};
\node[laneTitle, text=VivPink!75!black]    at (-.10,\yLe) {Formal Theorem\\Provers (Lean 4)};
\node[laneTitle, text=VivLilac!80!black]   at (-.10,\yLf) {Geometry \&\\Multimodal};
\node[laneTitle, text=VivPurple!85!black]  at (-.10,\yLg) {Verified\\Discovery};

\node[era, fill=VivCyan!42, minimum width=1.3cm, align=center]
   at (\xA+.15,\yERA) {symbolic /\\rule-based};
\node[era, fill=VivTeal!42, minimum width=1.2cm] at (\xB,\yERA) {statistical};
\node[era, fill=VivLime!40, minimum width=1.9cm] at ({(\xC+\xD)/2},\yERA) {neural MWP};
\node[era, fill=VivYellow!45, minimum width=3.0cm] at ({(\xE+\xG)/2},\yERA) {prompted / tool-augmented LLMs};
\node[era, fill=VivPink!32, minimum width=5.6cm] at ({(\xH+\xJ)/2},\yERA) {reasoning models, multi-agent, formal proving, verified discovery};

\draw[axisline] (0,\yAX) -- (\xMAX,\yAX);
\foreach \x/\yr in {\xA/1965,\xB/2014,\xC/2017,\xD/2019,\xE/2021,\xF/2022,\xG/2023,\xH/2024,\xI/2025,\xJ/2026}{
    \draw[draw=black!55, line width=.4pt] (\x,\yAX+.09) -- (\x,\yAX-.09);
    \node[yearlbl] at (\x,\yAX-.13) {\yr};
}
\node[font=\fontsize{5.9}{6.7}\selectfont\itshape, text=black!58] at ({(\xA+\xB)/2},\yAX+.22) {$\approx$\,50\,year gap};
\draw[draw=black!40, line width=.3pt, dash pattern=on 1pt off 1pt] (\xA+1.05,\yAX+.02) -- (\xA+1.05,\yAX+.20);
\draw[draw=black!40, line width=.3pt, dash pattern=on 1pt off 1pt] (\xB-.55,\yAX+.02) -- (\xB-.55,\yAX+.20);

\node[ev, draw=VivCyan!65!black] at (\xA,\yLa)
    {\textsc{Student},\\\textsc{WordPro}, \textsc{Robust}};
\node[ev, draw=VivCyan!75!black] at (\xB,\yLa)
    {\textsc{Aris}, log-linear,\\\textsc{Kushman}};

\node[ev, draw=VivLime!55!black] at (\xC,\yLb)
    {\textsc{Math23K},\\\textsc{Seq2Seq}};
\node[ev, draw=VivLime!55!black] at (\xD,\yLb)
    {\textsc{Gts},\\\textsc{Graph2Tree}};

\node[ev, draw=VivOrange!65!black] at ({(\xE+\xF)/2},\yLc)
    {\textsc{Gsm8k}, CoT,\\\textsc{Minerva}, verifiers};
\node[ev, draw=VivOrange!65!black] at (\xG,\yLc)
    {\textsc{Pal}/\textsc{PoT},\\\textsc{ToRA}, PRMs};

\node[ev, draw=VivSalmon!70!black, fill=VivSalmon!8] at (\xH,\yLd)
    {OpenAI o1,\\MoA};
\node[ev, draw=VivSalmon!70!black, fill=VivSalmon!8] at (\xI,\yLd)
    {DeepSeek-R1,\\Kimi k1.5};
\node[ev, draw=VivSalmon!70!black, fill=VivSalmon!8] at (\xJ-.35,\yLd)
    {Gemini DT\,\textcolor{VivPink!75!black}{$\bigstar$}\\o3, GoA};

\node[ev, draw=VivPink!70!black] at (\xG,\yLe)
    {\textsc{LeanDojo},\\\textsc{miniF2F}};
\node[ev, draw=VivPink!70!black, fill=VivPink!8] at (\xH,\yLe)
    {DSP-V1,\\AlphaProof\,\textcolor{VivPink!75!black}{$\bigstar$}};
\node[ev, draw=VivPink!70!black, fill=VivPink!8] at (\xI,\yLe)
    {DSP-V2,\\G\"odel-Prover};
\node[ev, draw=VivPink!70!black, fill=VivPink!8] at (\xJ-.35,\yLe)
    {APOLLO,\\\textsc{PutnamBench}};

\node[ev, draw=VivLilac!75!black] at (\xG,\yLf)
    {G-LLaVA,\\\textsc{MathVista}};
\node[ev, draw=VivLilac!75!black] at (\xH,\yLf)
    {AlphaGeometry,\\\textsc{Mavis}};
\node[ev, draw=VivLilac!75!black] at (\xI,\yLf)
    {AlphaGeom.~2,\\MINT-CoT};
\node[ev, draw=VivLilac!75!black] at (\xJ-.35,\yLf)
    {MARS-GPS,\\\textsc{MathVerse}};

\node[ev, draw=VivPurple!75!black] at (\xG,\yLg)
    {\textsc{FunSearch}};
\node[ev, draw=VivPurple!75!black] at (\xI,\yLg)
    {\textsc{AlphaEvolve}};
\node[ev, draw=VivPurple!75!black] at (\xJ-.35,\yLg)
    {Erd\H{o}s\\workflows};

\node[starM] at (.45,\yLEG) {};
\node[legendlbl] at (.75,\yLEG)
    {$\bigstar$\,=\,IMO-medal milestone, AlphaProof: silver tier, 28/42 with the rarely-solved P6 (IMO 2024)\,\textperiodcentered\,Gemini Deep Think \& OpenAI reasoning: gold tier, 35/42 (IMO 2025)};

\end{tikzpicture}%
}
\caption{Chronology of AI systems for mathematical reasoning, organized as paradigm swimlanes. Early symbolic and statistical MWP solvers (1965--2016, shown on a compressed axis) yield to neural expression generators (2017--2020), prompted and tool-augmented LLMs (2021--2023), and the 2024--2026 convergence of RLVR-trained reasoning models, Lean-based proof assistants, multimodal geometers, and verified-discovery systems. The seven lanes are color-coded to match Figure~\ref{fig:classification}'s four-axis taxonomy. Stars ($\bigstar$) mark IMO-medal milestones. Abbreviations: MWP\,=\,math word problem; CoT\,=\,chain-of-thought; \textsc{Pal}/\textsc{PoT}\,=\,program-aided / program-of-thoughts prompting; PRM\,=\,process reward model; RLVR\,=\,reinforcement learning from verifiable rewards; MoA\,=\,mixture-of-agents; GoA\,=\,graph-of-agents; DSP\,=\,DeepSeek-Prover; DT\,=\,Deep Think.}
\label{fig:timeline}
\end{figure*}

\section{Math Word Problem Solving}\label{sec:mwp}
The dawn of research on MWP solving was in the mid-1960s~\cite{feigenbaum1963computers,bobrow1964natural}. Since then, a rich succession of approaches has been developed. Following and extending earlier surveys~\cite{10.1007/978-981-10-7590-2_7,zhang2020gap,sundaram2022nlp}, we organize the literature into seven categories: rule-based, statistical, tree-based, semantic-parsing-based, similarity-based, template-based, and deep-learning-based methods. The deep-learning family is further decomposed into \textsc{Seq2Seq}, reinforcement-learning, graph-based, and complex encoder/decoder sub-families.

\subsection{Rule-based Methods}
Rule-based methods are chronologically the earliest approaches. They leverage manually hard-coded schemata about the language being analyzed to identify regularities. The author of~\cite{fletcher1985understanding} proposed \textsc{WordPro}, coded in Interlisp-D, which could solve one-step arithmetic problems with four predefined schemata. The author of~\cite{bakman2007robust} later proposed \textsc{Robust}, which conceptualized free-format multi-step MWPs with extraneous information using six predefined schemata. In~2010, \textsc{Mswpas}~\cite{yuhui2010frame} solved multi-step addition and subtraction problems by converting the problem statements into \emph{Problem Frames} containing the whole semantic information of the problem. The principal drawbacks of these methods are (i)~high dependency on manual features, and (ii)~an inability to generate novel templates for new problems. We provide a brief overview here; interested readers may consult~\cite{mukherjee2008review} for detailed descriptions.

\subsection{Statistical Methods}
Statistical methods use generic machine-learning models such as support vector machines and Bayes classifiers to extract entities, quantities, and operators from the problem statement, and to infer the numeric answer with simple logic. \cite{kushman2014learning}~proposed an algorithm that reasoned across sentence boundaries and defined a joint log-linear distribution over systems of equations, employing a two-step process that first selects an equation template and then instantiates it with quantities from the problem. This method laid the groundwork for incorporating semantic interpretation and information extraction into MWP solving, but it falters on problems with new compositional language because it lacks sufficient background knowledge.

\cite{roy2015reasoning}~proposed a Quantity Entailment scheme employing a cascade of three classifiers: a Quantity Pair Classifier outputs the pair of quantities required, an Operation Classifier selects one of $\{+,-,\times,\div\}$, and an Order Classifier (relevant for $-$ and $\div$) determines the most likely order of quantities. This scheme is, however, limited to single-operator expressions. \cite{hosseini2014learning}~proposed \textsc{Aris}, an early attempt at more advanced logic templates for multi-step problems. \textsc{Aris} represents the problem text as a world-state tuple $\langle E,C,R \rangle$ of Entities, Containers, and Relations, and introduces a seven-category verb categorization. It was accompanied by the \textsc{Ai\textsuperscript{3}} dataset of 395 problems, but it handles only $+$ and $-$.

\cite{zhou2015learn}~considered all possible equation systems in the hypothesis space and obtained a robust decision hyperplane using support vector machines trained with quadratic programming; it outperformed~\cite{kushman2014learning} but was still unable to resolve MWPs with complex noun phrases and lexical features. \cite{mitra2016learning}~proposed a system that learns to apply formulae categorized as \emph{part-whole}, \emph{change}, and \emph{comparison}, scored against the \textsc{AddSub} dataset, but still restricted to additive operators. \cite{liang2016tag,liang-etal-2016-meaning}~transformed problem scenarios and questions into tag-based logic forms for inference.

Two major drawbacks characterize this family: (i)~they require manual annotation that is costly to scale, and (ii)~they rely on pre-defined templates that are inflexible with respect to multiplication and division.

\subsection{Tree-based Methods}
Algebraic and arithmetic expressions have an inherent binary-tree structure, which motivates tree-based solvers. In these trees, leaf nodes represent constants or variables while internal nodes represent operators; operators with lower priority occupy positions higher in the tree.

\begin{figure}[h]
    \centering
    \tikzset{every tree node/.style={minimum width=1.2em,draw,circle},
             blank/.style={draw=none},
             edge from parent/.style=
             {draw,edge from parent path={(\tikzparentnode) -- (\tikzchildnode)}},
             level distance=1.3cm}
    \begin{tikzpicture}[->,auto,every state/.style={draw=black!60, fill=black!5, very thick}]
    \Tree
    [.$+$
        [.$\div$
            \edge[]; [.$n_1$ ]
            \edge[blank]; \node[blank]{};
            \edge[]; [.$n_2$ ]
        ]
        [.$\times$
        \edge[]; [.$-$
                 \edge[]; {$n_3$}
                 \edge[blank]; \node[blank]{};
                 \edge[]; {$n_4$}
             ]
        \edge[blank]; \node[blank]{};
        \edge[]; {$n_5$}
        ]
    ]
    \end{tikzpicture}
    \caption{Expression Tree representing $n_1 \div n_2 + (n_3-n_4)\times n_5$.}
    \label{fig:fig5}
\end{figure}
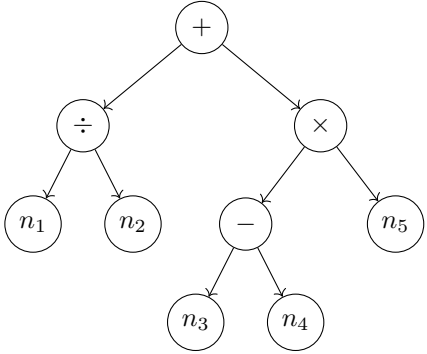

\begin{figure}[h]
    \centering
    \tikzset{every tree node/.style={minimum width=1.2em,draw,circle},
             blank/.style={draw=none},
             edge from parent/.style=
             {draw,edge from parent path={(\tikzparentnode) -- (\tikzchildnode)}},
             level distance=1.3cm}
    \begin{tikzpicture}[->,auto,every state/.style={draw=black!60, fill=black!5, very thick}]
    \Tree
    [.$=$
        [.$\div$
            \edge[]; [.$+$
                 \edge[]; {$x$}
                 \edge[blank]; \node[blank]{};
                 \edge[]; {$n_1$}
             ]
             \edge[blank]; \node[blank]{};
             \edge[]; [.$n_2$ ]
        ]
        [.$n_3$ ]
    ]
    \end{tikzpicture}
    \caption{Equation Tree representing $(x + n_1) \div n_2 = n_3$.}
    \label{fig:fig6}
\end{figure}
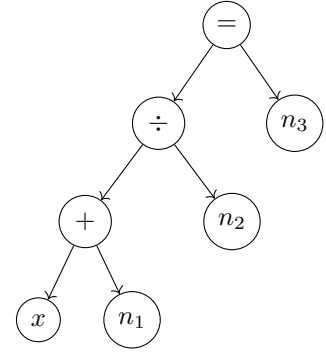

\cite{roy2016solving}~pioneered the use of expression trees for MWP solving by first applying a binary classifier to discard irrelevant quantities, then aggregating simple prediction problems to determine the Lowest Common Ancestor (LCA) of quantity pairs. The score of an expression $E$ represented by the tree $T$ is
\begin{equation}\label{eq:1}
\begin{aligned}
\mathbf{Score}(E) =\ & w_{\textsc{Irr}}\!\!\!\sum_{q \in I(E)}\!\!\!\textsc{Irr}(q) \\
&+ \!\!\!\sum_{q_i,q_j \notin I(E)}\!\!\!\textsc{Pair}(q_i,q_j,\odot_{\textsc{LCA}}(q_i,q_j,T)),
\end{aligned}
\end{equation}
where $\textsc{Irr}(q)$ scores the probability that quantity $q$ is irrelevant to the solution, $\textsc{Pair}(q_i, q_j, \odot)$ scores the likelihood that quantities $q_i, q_j$ should be combined by operator~$\odot$, and $\odot_{\textsc{LCA}}(q_i, q_j, T)$ retrieves the operator at the lowest common ancestor of $q_i$ and $q_j$ in expression tree~$T$. Inference selects the highest-scoring expression $E^* = \operatorname*{argmax}_{E \in C} \mathbf{Score}(E)$.
The concept of a \emph{quantity schema} parses out the information relevant to each quantity. The system was evaluated on \textsc{Ai2}, \textsc{Il}, and \textsc{CommonCore}, with an online version implemented in~\cite{roy2016illinois}.

\textsc{Alges}~\cite{koncel2015parsing} generates equation trees from multi-sentence MWPs using Integer Linear Programming and scores them via local and global discriminative models using a compact Quantified-Set (\emph{Qset}) representation, selecting the tree that maximizes the product of local subtree likelihoods and a global coherence factor conditioned on the problem text.

\cite{roy2016equation}~efficiently parses problem text into projective equations, assuming at most two variables and using each quantity at most once. \cite{roy2017unit}~built on~\cite{roy2016solving} by introducing \emph{Unit Dependency Graphs} (UDGs) that capture relationships between the units of quantities. Each vertex in the UDG corresponds to a quantity and is scored by its unit role (\textit{e.g.}, rate \textit{vs.}\ raw count), while edges encode pairwise compatibility; rate--count pairs, for instance, signal multiplication rather than addition. The best graph is selected by maximizing the sum of vertex and edge scores, weighted by a hyperparameter~$\lambda$.

\cite{wang2018translating,chiang2018semantically}~used implicit tree structures and \textsc{Seq2Seq} models; \textsc{MathEN}~\cite{wang2018translating} introduced an equation-normalization method combined with an ensemble of Bi-LSTM~\cite{wu2016google}, \textsc{ConvS2S}~\cite{gehring2017convolutional}, and Transformer~\cite{vaswani2017attention} components. These tree-based models advantageously do not require additional manual annotations such as templates, tags, or logic forms, and directly informed the subsequent \textsc{Gts}~\cite{xie2019goal} and \textsc{Graph2Tree}~\cite{zhang2020graph} lines.

\subsection{Semantic Parsing-based Methods}
\cite{shi2015automatically}~presented \textsc{SigmaDolphin}, using a newly designed Dolphin Language (DOL) meaning-representation. DOL trees are produced by a CFG~\cite{chomsky1956three,gildea2002automatic} parser imbued with 9{,}600 grammar rules, scored by
\begin{equation}\label{eq:6}
\mathbf{Score}(T) = \frac{\sum_{i=1}^{k}L(T_i)\cdot\mathbf{Score}(T_i)}{\sum_{i=1}^{k}L(T_i)}\cdot p(T),
\end{equation}
and passed to a reasoning module to produce the final answer. It introduced a 1{,}878-problem dataset from \texttt{algebra.com} and \texttt{answers.yahoo.com}. \textsc{Text2Math}~\cite{zou2019text2math} performs end-to-end latent-variable prediction without \textit{a priori} knowledge of operators. Such parsing-based methods tend to be limited to narrow classes of number word problems.

\subsection{Similarity-based Methods}
\textsc{Sim}~\cite{huang2016well} computes similarity between a test sample and training examples via weighted Jaccard coefficients of TF--IDF vectors, applies the equation system of the most similar training sample, and fills its slots from the test problem. It also introduced \textsc{Dolphin18K}, a large-scale dataset of 18{,}460 annotated MWPs. Similarity-based methods fail on problems whose structural templates do not appear in the training set.

\subsection{Template-based Methods}
Template-based approaches identify a candidate equation template from a pre-defined corpus and fill numeric and variable slots with quantities extracted from the problem. Several works discussed under statistical methods adopt this scheme~\cite{kushman2014learning,zhou2015learn,roy2016equation}. Because the search space is exponential in the number of slots, beam-search is typically used. \textsc{MixedSP}~\cite{upadhyay2016learning} uses both explicit (equations) and implicit (solutions) supervision via structured-output perceptrons~\cite{collins2002discriminative} and introduced \textsc{Sol2k}. \textsc{FGExpression}~\cite{huang2017learning} captures rich information from templates by parsing them into tree structures and defining \emph{template fragments}, yielding a fine-grained mapping based on Longest Common Substring.

\subsection{Deep Learning-based Methods}
Deep-learning methods learn representations directly from data, avoiding hand-designed features.

\paragraph*{Seq2Seq methods}
\cite{wang2017deep}~introduced \textsc{Dns}, an RNN-based \textsc{Seq2Seq} with a similarity-based retrieval component. The encoder uses GRUs~\cite{chung2014empirical}, the decoder uses LSTM~\cite{hochreiter1997long}, and a Significant Number Identification (SNI) model identifies relevant numbers in the problem text. It released the landmark \textsc{Math23K} corpus of~23{,}161 Chinese MWPs. \textsc{EquGener}~\cite{mishra2018equgener} employs a memory-network encoder with an LSTM decoder and supports all four fundamental operators using \textsc{GloVe}~\cite{pennington2014glove} and learned embeddings.

\paragraph*{Deep reinforcement learning methods}
\textsc{MathDQN}~\cite{wang2018mathdqn} was the first application of deep RL to MWPs. It formulates expression generation as a Markov decision process in which the agent sequentially selects quantities and operators, trained via standard DQN with an $\epsilon$-greedy exploration strategy. Its principal contribution was to show that RL could avoid the exposure-bias problem of teacher-forced \textsc{Seq2Seq} training, but its accuracy gains over supervised baselines were modest.

\paragraph*{Improved Seq2Seq methods}
\textsc{Cass}~\cite{huang2018neural} added a copy-and-align mechanism and an RL objective, finding empirically that RL is preferable to maximum-likelihood estimation for this task. \textsc{GroupAtt}~\cite{li2019modeling} proposed a group-attention mechanism that partitions the encoder's self-attention into four parallel modules, global, quantity-related, quantity-pair, and question-related, before aggregating them into a unified representation; this decomposition lets the encoder attend to different functional roles of the input simultaneously, improving quantity--operator alignment. \textsc{T-Rnn}~\cite{wang2019template} uses a recursive neural network to predict a tree-structured template, composing left- and right-child representations bottom-up and selecting operators at each internal node via softmax; the key idea is that tree structure is predicted \emph{before} numeric slot-filling, separating structural reasoning from quantity assignment. \textsc{S-Aligned}~\cite{chiang2018semantically} is a neural symbolic model whose decoder generates equations by stack operations mimicking human reasoning; it introduces operator-specific ``Semantic Transformers'' that apply distinct learned nonlinear projections per arithmetic operator, enabling the model to learn operator-dependent composition rules rather than a single generic combination function.

\paragraph*{Graph-based methods}
\textsc{Gts}~\cite{xie2019goal} imitates human problem-solving with a goal-driven recursive tree-expansion decoder. It improves upon pure \textsc{Seq2Seq} by avoiding mathematically invalid equations but cannot generate multiple valid solutions. \textsc{Ast-Dec}~\cite{liu2019tree} is a hierarchical \textsc{Seq2Tree} model with an auxiliary-stack decoder producing prefix-notation equations. \textsc{D-Decoder}~\cite{meng2019solving} pioneered Transformer decoders for math equations with two decoders operating in opposite directions, improving BERT-style training~\cite{jdevlin2018bert}.

\textsc{Graph2Tree}~\cite{zhang2020graph} fuses Graph-Transformer encoders~\cite{yun2019graph,cai2020graph} with tree decoders, using two complementary graphs: a \emph{Quantity Cell Graph}, whose edges connect quantities that co-occur in the same sentence or share a syntactic dependency, and a \emph{Quantity Comparison Graph}, whose edges encode magnitude or unit-type comparisons between quantities. The reason this graph structure helps is concrete: in a problem that mentions a rate (``\$23 each'') and a count (``69 handbags''), a sequential encoder can learn the rate-times-count pattern only from word-order proximity, which breaks under paraphrase or reordering; a graph encoder encodes this relationship as an explicit edge, making the representation invariant to surface permutation. This is precisely the ``quantity attachment'' failure mode that perturbation benchmarks later diagnosed (Section~\ref{sec:failures}).

Architecturally, each of the $K$ graph-convolution heads applies two layers of message-passing over one adjacency view $A_k$, the $K$ outputs are concatenated, and a feed-forward block with residuals and layer-normalization produces the graph embedding, which is then fed to a tree decoder resembling \textsc{Gts}. The training loss is the standard token-level cross-entropy over the decoder's prefix-order output:
\begin{equation}\label{eq:24}
L(T,P) = -\sum_{t=1}^{E}\log P(y_t|q_t,G_c,P).
\end{equation}
A parallel \textsc{Graph2Tree}~\cite{li2020graph} uses hierarchical tree decoders with parent- and sibling-feeding. \textsc{Roda}~\cite{liu2020reverse} introduces a \emph{reversion-based} data-augmentation scheme that rewrites a problem with its inferred answer substituted for a known quantity. \textsc{Smart}~\cite{hong2021smart} draws on the Situation Model~\cite{kintsch1985understanding} from cognitive psychology, using attributed grammar over a hierarchical parse graph; it was accompanied by the \textsc{Asp6.6K} dataset and an OOD evaluation protocol.

\paragraph*{Complex encoder-decoder methods}
\textsc{Ept}~\cite{kim2020point} is an Expression-Pointer Transformer built on ALBERT~\cite{lan2019albert} that addresses expression fragmentation and operand-context separation. \textsc{MultiE/D}~\cite{shen2020solving} uses both sequence- and graph-based encoders (the latter based on \textsc{GraphSAGE}~\cite{hamilton2017inductive}) and sequence- and tree-based decoders. \textsc{Ka-S2T}~\cite{wu2020knowledge} integrates external knowledge via a Graph Attention Network over an entity graph. \textsc{Tsn-Md}~\cite{zhang2020teacher} derives multiple correct expressions per problem via a teacher--student ensemble with knowledge distillation~\cite{hinton2015distilling}. \textsc{Rpkhs}~\cite{yu2021improving} combines a pre-trained knowledge encoder with a hierarchical-reasoning encoder, reaching $89.8\%$ on \textsc{Mawps}. \textsc{Lbf}~\cite{hong2021learning} introduces a \emph{Learning-by-fixing} framework with tree regularization. \textsc{Hms}~\cite{lin2021hms} uses a hierarchical word--clause--problem encoder. \textsc{Ns-Solver}~\cite{qin2021neural} combines a problem encoder, a symbolic equation-generator decoder, and a symbolic executor with four auxiliary objectives, also releasing the \textsc{Cm17K} benchmark. \textsc{Real}~\cite{huang2021recall} emulates human analogical learning via memory-augmented retrieval. Diverging from the generative sequence-to-tree paradigm, \textsc{DeductReasoner}~\cite{jie2022deductive} frames MWP solving as a complex relation extraction task. By iteratively predicting primitive operations over pairs of quantities, it produces explainable deductive reasoning steps while achieving strong performance across classical benchmarks.

Before the LLM era, the strongest systems were \textsc{MWP-Bert}~\cite{liang2021mwp}, which uses a BERT-based encoder further pre-trained on \textsc{Ape210K} with a tree decoder and achieved $84.4\%$ on \textsc{Math23K} and $84.3\%$ on \textsc{Ape210K}; \textsc{Generate\&Rank}~\cite{shen2021generate}, a BART-based~\cite{lewis2019bart} multi-task framework reaching $85.4\%$ on \textsc{Math23K}; and OpenAI's verifier-incorporated GPT-3~\cite{cobbe2021training,brown2020language}, which introduced the \textsc{Gsm8K} benchmark and demonstrated that sampling many high-temperature solutions and scoring them with a trained verifier could yield performance gains equivalent to a 30\(\times\) increase in model size. This last result is in some sense the seed from which the entire reasoning-model research program of 2024--2026 would grow.

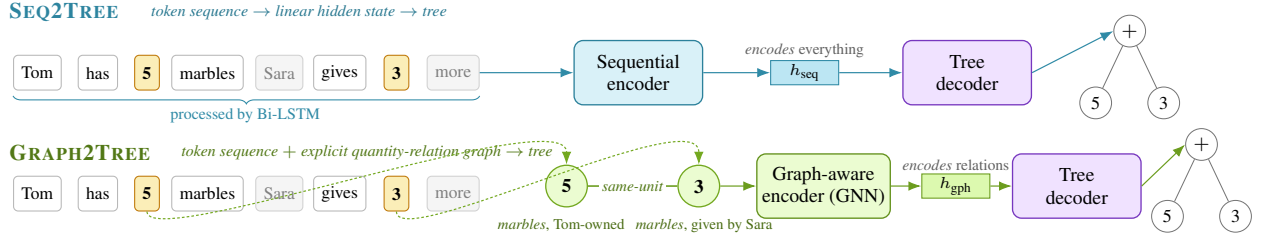
\begin{figure*}[t]
\centering
\begin{tikzpicture}[
    tok/.style={rectangle, draw=black!28, rounded corners=.6mm, fill=white,
                font=\fontsize{6.6}{7.4}\selectfont, minimum height=.46cm,
                inner xsep=3pt, inner ysep=2pt},
    quant/.style={tok, fill=VivYellow!28, draw=VivOrange!75!black, font=\fontsize{6.6}{7.4}\bfseries\selectfont},
    elide/.style={tok, fill=black!4, text=black!50, draw=black!18},
    blk/.style={rectangle, rounded corners=1.2mm, line width=.5pt,
                align=center, font=\fontsize{7.4}{8.4}\selectfont,
                minimum width=1.7cm, minimum height=.85cm, inner sep=2pt},
    blkS/.style={blk, draw=VivCyan!70!black, fill=VivCyan!20},
    blkG/.style={blk, draw=VivLime!60!black, fill=VivLime!22},
    blkD/.style={blk, draw=VivPurple!75!black, fill=VivPurple!16},
    hidS/.style={rectangle, draw=VivCyan!75!black, fill=VivCyan!35,
                 minimum width=0.9cm, minimum height=.32cm, line width=.4pt},
    hidG/.style={rectangle, draw=VivLime!65!black, fill=VivLime!38,
                 minimum width=0.9cm, minimum height=.32cm, line width=.4pt},
    qnode/.style={circle, draw=VivLime!70!black, fill=VivLime!22, line width=.5pt,
                  inner sep=0pt, font=\fontsize{7.2}{8.0}\bfseries\selectfont,
                  minimum size=.55cm},
    treenode/.style={circle, draw=black!55, fill=white, line width=.35pt,
                     inner sep=0pt, font=\fontsize{6.8}{7.6}\selectfont,
                     minimum size=.42cm},
    arr/.style={-{Latex[length=1.7mm]}, draw=black!60, line width=.45pt},
    arrC/.style={arr, draw=VivCyan!75!black},
    arrG/.style={arr, draw=VivLime!65!black},
    elab/.style={font=\fontsize{5.8}{6.6}\selectfont\itshape, text=black!62, inner sep=1pt},
    plab/.style={font=\fontsize{8.4}{9.4}\bfseries\selectfont},
    note/.style={font=\fontsize{6.0}{6.8}\selectfont, text=black!65, align=left, inner sep=0pt}
]
\node[rectangle, draw=black!22, rounded corners=.8mm, fill=black!3,
      text width=13.5cm, align=center, inner xsep=4pt, inner ysep=3pt,
      font=\fontsize{7.6}{8.6}\selectfont\itshape] (prob) at (8,7.0)
   {Problem: ``Tom has \textbf{5} marbles. Sara gives him \textbf{3} more. How many marbles does Tom have?''};

\node[plab, anchor=west, text=VivCyan!72!black] at (0.05,5.85) {\textsc{Seq2Tree}};
\node[note, anchor=west, text=VivCyan!65!black] at (2.05,5.85) {\emph{token sequence $\to$ linear hidden state $\to$ tree}};

\node[tok]   (s1) at ( 0.55, 5.05) {Tom};
\node[tok]   (s2) at ( 1.35, 5.05) {has};
\node[quant] (s3) at ( 2.00, 5.05) {5};
\node[tok]   (s4) at ( 2.80, 5.05) {marbles};
\node[elide] (s5) at ( 3.75, 5.05) {Sara};
\node[tok]   (s6) at ( 4.55, 5.05) {gives};
\node[quant] (s7) at ( 5.30, 5.05) {3};
\node[elide] (s8) at ( 6.05, 5.05) {more};

\draw[decorate, decoration={brace, mirror, raise=2pt, amplitude=3pt},
      draw=VivCyan!75!black, line width=.4pt]
   (s1.south west) -- (s8.south east);
\node[note, text=VivCyan!75!black, anchor=north] at ($(s1.south)!0.5!(s8.south) + (0,-.22)$)
   {processed by Bi-LSTM};

\node[blkS] (encS) at ( 8.5, 5.05) {Sequential\\encoder};
\draw[arrC] (s8.east) -- (encS.west);

\node[hidS] (hS) at (10.7, 5.05) {};
\node[font=\fontsize{6.0}{6.8}\selectfont, anchor=center] at (hS) {$h_{\text{seq}}$};
\node[elab, anchor=south] at (hS.north) {encodes \emph{everything}};
\draw[arrC] (encS.east) -- (hS.west);

\node[blkD] (decS) at (12.85, 5.05) {Tree\\decoder};
\draw[arrC] (hS.east) -- (decS.west);

\node[treenode] (sT1) at (15.0, 5.60) {$+$};
\node[treenode] (sT2) at (14.55, 4.65) {5};
\node[treenode] (sT3) at (15.45, 4.65) {3};
\draw[draw=black!55, line width=.35pt] (sT1) -- (sT2);
\draw[draw=black!55, line width=.35pt] (sT1) -- (sT3);
\draw[arrC] (decS.east) -- (sT1.west);

\node[plab, anchor=west, text=VivLime!50!black] at (0.05,4) {\textsc{Graph2Tree}};
\node[note, anchor=west, text=VivLime!50!black] at (2.45,4) {\emph{token sequence $+$ explicit quantity-relation graph $\to$ tree}};

\node[tok]   (g1) at ( 0.55, 3.45) {Tom};
\node[tok]   (g2) at ( 1.35, 3.45) {has};
\node[quant] (g3) at ( 2.00, 3.45) {5};
\node[tok]   (g4) at ( 2.80, 3.45) {marbles};
\node[elide] (g5) at ( 3.75, 3.45) {Sara};
\node[tok]   (g6) at ( 4.55, 3.45) {gives};
\node[quant] (g7) at ( 5.30, 3.45) {3};
\node[elide] (g8) at ( 6.05, 3.45) {more};

\node[qnode] (q5) at ( 7.55, 3.55) {5};
\node[qnode] (q3) at ( 9.30, 3.55) {3};
\draw[draw=VivLime!65!black, line width=.6pt]
   (q5) -- node[elab, fill=white, text=VivLime!50!black] {same-unit} (q3);

\node[note, anchor=north, text=VivLime!45!black, font=\fontsize{5.6}{6.4}\selectfont]
   at ([xshift=-2pt,yshift=-4pt]q5.south) {{\itshape marbles}, Tom-owned};
\node[note, anchor=north, text=VivLime!45!black, font=\fontsize{5.6}{6.4}\selectfont]
   at ([xshift=2pt,yshift=-4pt]q3.south) {{\itshape marbles}, given by Sara};

\draw[arrG, draw=VivLime!70!black, dash pattern=on 1pt off .8pt]
   (g3.south) .. controls +(0,-.55) and +(0,.85) .. (q5.north);
\draw[arrG, draw=VivLime!70!black, dash pattern=on 1pt off .8pt]
   (g7.south) .. controls +(0,-.55) and +(0,.85) .. (q3.north);

\node[blkG] (encG) at (10.95, 3.55) {Graph-aware\\encoder (GNN)};
\draw[arrG] (q3.east) -- (encG.west);

\node[hidG] (hG) at (12.70, 3.55) {};
\node[font=\fontsize{6.0}{6.8}\selectfont, anchor=center] at (hG) {$h_{\text{gph}}$};
\node[elab, anchor=south] at (hG.north) {encodes \emph{relations}};
\draw[arrG] (encG.east) -- (hG.west);

\node[blkD] (decG) at (14.30, 3.55) {Tree\\decoder};
\draw[arrG] (hG.east) -- (decG.west);

\node[treenode] (gT1) at (15.95, 4.10) {$+$};
\node[treenode] (gT2) at (15.50, 3.15) {5};
\node[treenode] (gT3) at (16.40, 3.15) {3};
\draw[draw=black!55, line width=.35pt] (gT1) -- (gT2);
\draw[draw=black!55, line width=.35pt] (gT1) -- (gT3);
\draw[arrG] (decG.east) -- (gT1.west);

\end{tikzpicture}
\caption{A worked-example comparison of \textsc{Seq2Tree} (top) and \textsc{Graph2Tree} (bottom) on the same problem; yellow-highlighted tokens are extracted quantities. \textsc{Seq2Tree} must \emph{infer} from the LSTM's hidden state that the two numbers share the unit ``marbles''; \textsc{Graph2Tree} receives this fact explicitly through a quantity-relation graph encoding ``5 \emph{same-unit} 3'', so the GNN's hidden state separates \emph{which} quantities can combine arithmetically from \emph{how} they appear in surface text, a structural prior that helps most on multi-sentence problems with distractor quantities. Both pipelines share a tree decoder and produce the same expression tree.}
\label{fig:seq_graph_tree}
\end{figure*}

\begin{table}[ht]
\centering
\resizebox{\columnwidth}{!}{%
\begin{tabular}{lcccc}
    \tabletop
    \thead{Model}       & \thead{Seq-Enc.}   & \thead{Graph-Enc.} & \thead{Seq-Dec.}   & \thead{Tree-Dec.}  \\ \tablemid
    DNS~\cite{wang2017deep}         & \cmark    &               & \cmark    &               \\
    Math-EN~\cite{wang2018translating}     & \cmark    &               & \cmark    &               \\
    T-RNN~\cite{wang2019template}     & \cmark    &               & \cmark    &               \\
    S-Aligned~\cite{chiang2018semantically}   & \cmark    &               &        \cmark       &               \\
    Group-ATT~\cite{li2019modeling}   & \cmark    &               &       \cmark        &               \\
    D-Decoder~\cite{meng2019solving}  & \cmark    &               &       \cmark        &               \\
    AST-Dec~\cite{liu2019tree}     & \cmark    &               &               & \cmark    \\
    GTS~\cite{xie2019goal}         & \cmark    &               &               & \cmark    \\
    \tsoft Graph2Tree~\cite{zhang2020graph}  & \cmark    & \cmark    &               & \cmark    \\
    \tsoft Graph2Tree~\cite{li2020graph}  & \cmark    & \cmark    &               & \cmark    \\
    \timportant Multi-E/D~\cite{shen2020solving}  & \cmark    & \cmark    & \cmark    & \cmark    \\
    \tsoft MWP-BERT~\cite{liang2021mwp}  & \cmark (BERT)  &               &               & \cmark    \\
    \tsoft Generate\&Rank~\cite{shen2021generate}  & \cmark (BART)  &               & \cmark    &               \\
    \tablebottom
\end{tabular}%
}
\caption{Architectural breakdown of representative pre-LLM neural MWP solvers. Cyan shading marks systems that add graph, pretraining, or reranking structure beyond the earlier sequence-only pattern; magenta marks the fully hybrid encoder--decoder design.}
\label{tab:tab2}
\end{table}

\subsection{Feature Engineering in the Pre-LLM Era}
A distinguishing characteristic of pre-deep-learning work was the explicit design of hand-crafted features. Quantity-related features determine whether a number is a rate versus a raw count~\cite{roy2016solving,roy2017unit,wang2018mathdqn,zhou2015learn,upadhyay2016learning}; context-related features leverage POS tags and dependency types within a text window~\cite{roy2016solving,roy2017unit,wang2018mathdqn}; quantity-pair features distinguish same-unit quantities (typically combined by $+/-$) from rate--unit pairs (typically combined by $\times/\div$)~\cite{koncel2015parsing,roy2016solving,roy2017unit,wang2018mathdqn}. Question-related features identify the unit or noun phrase referenced by the question~\cite{kushman2014learning,roy2015reasoning,koncel2015parsing,roy2016solving,roy2017unit,wang2018mathdqn}. Verb-related features include the dependent verb of a quantity~\cite{koncel2015parsing,roy2016solving,roy2017unit,wang2018mathdqn,hosseini2014learning,liang2016tag}. Global features capture document-level properties, including $n$-gram statistics~\cite{kushman2014learning,roy2015reasoning,roy2016equation,roy2017unit,wang2018mathdqn}. The transition to neural models obviated most of these features, although several reappear implicitly in the attention patterns of modern architectures.

\begin{table}[t]
\centering
\scriptsize
\renewcommand{\arraystretch}{1.32}
\setlength{\tabcolsep}{3.5pt}
\providecommand{\stripe}[1]{\textcolor{#1}{\rule[-.40ex]{1.6pt}{1.7ex}}\hspace{2pt}}
\begin{tabularx}{\columnwidth}{@{}p{0.235\columnwidth}>{\raggedright\arraybackslash}X p{0.295\columnwidth}@{}}
\tabletop
\thead{Era} & \thead{Innovation \emph{(why it helped)}} & \thead{Bottleneck that triggered the next era} \\ \tablemid
\stripe{VivCyan!65!black}Rule-based                       & Verb categories \& problem frames \emph{(make implicit semantics explicit)}                                       & Fixed schema vocabulary \\
\stripe{VivTeal!75!black}Statistical / parsing            & Equation templates, log-linear scoring, logical forms \emph{(cross-sentence information; interpretable slots)}    & New compositional language \\
\stripe{VivLime!55!black}Tree-based                       & Expression trees as the output space \emph{(grammar-constrained valid equations)}                                  & Multi-solution problems \\
\stripe{VivYellow!65!black}Seq2Seq                        & End-to-end neural mapping \emph{(removes many hand-crafted features)}                                              & Fragile under perturbation; invalid equations \\
\stripe{VivOrange!75!black}Graph-based                    & Quantity-cell graphs $+$ tree decoders \emph{(captures inter-quantity relational structure)}                       & Multi-step commonsense reasoning \\
\stripe{VivPink!75!black}PLM-based                        & BERT/BART pretraining $+$ decoders / rerankers \emph{(transfer learning, stronger representations)}                & Dataset-specific shortcuts; transition to LLMs \\
\timportant \stripe{VivPurple!85!black}Verifier-aug.\ LLMs & Sample many traces, then rank or check candidates \emph{(generation becomes search-and-selection)}                & Reliable graders required; semantic errors can survive execution \\
\tablebottom
\end{tabularx}
\caption{Methodological progression of MWP solving, read as a ladder. Each row's right column lists the failure mode that the \emph{next} row's innovation was specifically designed to dissolve, so the chain runs as bottleneck\,$\to$\,resolution\,$\to$\,new bottleneck through seven generations. Colored stripes follow the paradigm palette of Figure~\ref{fig:timeline} and saturate from light to deep as the supervision constraint strengthens, an in-table miniature of the supervision-ladder argument developed in Section~\ref{sec:synthesis}. The magenta-shaded final row marks the current frontier, which is also the only generation whose bottleneck has not yet been resolved by a successor era.}
\label{tab:mwp_progression}
\end{table}

\subsection{Legacy of Classical MWP Work}
Although the benchmark leaderboards have moved far beyond hand-engineered MWP systems, the classical literature remains important for three reasons. First, it made explicit the linguistic phenomena that still cause failures: quantity attachment, rate interpretation, unit conversion, comparison phrases, temporal order, and irrelevant clauses. Modern LLMs often solve such cases without exposing a symbolic representation, but perturbation benchmarks such as \textsc{SVAMP}~\cite{patel2021nlp} and \textsc{GSM-Symbolic}~\cite{mirzadeh2024gsm} show that these same phenomena remain diagnostic.

Second, the pre-LLM systems anticipated the current emphasis on intermediate structure. A \textsc{Seq2Seq} model that directly maps a paragraph to a prefix expression differs in scale from an o-series reasoning model, but both are asked to emit a latent computation trace. The difference is that older systems constrained the trace by grammar and templates, while current systems learn longer free-form traces and rely on verifiers, rerankers, or tools to select among them.

Third, classical MWP work provides a useful warning about overfitting to benchmark form. Many systems achieved strong accuracy on \textsc{MAWPS} or \textsc{Math23K} but failed under small paraphrases or quantity perturbations. This pattern recurs at every scale: whenever the dataset distribution is narrow, models learn shortcuts; whenever the evaluation is refreshed, adversarial, or mechanically generated, the same models expose gaps between answer accuracy and robust mathematical understanding.

\subsection{Why Structured Representations Helped: A Retrospective}
 
Viewed from the vantage point of 2026, the pre-LLM MWP
literature reveals a clear pattern:\ \emph{each successful
method gained its edge by making an implicit linguistic
structure explicit}.  Rule-based systems made verb
categories and problem frames explicit; statistical methods
made equation templates explicit; and the decisive leap came
when \textsc{Graph2Tree}~\cite{zhang2020graph} made
\emph{inter-quantity relational structure} explicit through
graph encoders.  The reason graph encoders improved over
sequential encoders was not simply architectural fashion:
quantity-cell graphs captured cross-quantity dependencies
(\textit{e.g.}, that a rate and a count should be multiplied, not
added) that a left-to-right encoder could only learn
indirectly from word order.  This is precisely the
``quantity attachment'' failure mode identified in
Section~\ref{sec:failures} and the perturbation studies of
\textsc{SVAMP}~\cite{patel2021nlp}: when surface order is
changed but relational structure is preserved, graph-aware
models degrade less.
 
The pattern also explains what broke each era.  Templates
broke when problem language exceeded the template
vocabulary.  Supervised expression generation broke when
evaluation moved beyond in-distribution test sets.  And even
\textsc{Graph2Tree} broke when the problems required
multi-step common-sense reasoning that no graph over
explicit quantities could capture, a gap that only the
implicit world knowledge of large pretrained models would
begin to close.  The transition to Section~\ref{sec:llm} is
therefore not merely chronological; it reflects a shift from
\emph{designing the right intermediate structure by hand} to
\emph{learning it from data and then verifying it externally},
a shift whose consequences pervade the remainder of this
survey.

\section{The LLM and Reasoning-Model Era}\label{sec:llm}
Systems discussed in this section correspond to the ``LLMs \& Agents'' and ``Reasoning \& Verification'' bands in Figure~\ref{fig:timeline}. The publication of \textsc{GSM8K}~\cite{cobbe2021training} in late 2021 marked a pivot point: grade-school math word problems ceased to be a task researchers tried to solve \textit{ad hoc} and became a standard benchmark against which the reasoning capability of general-purpose large language models would be measured. The subsequent five years have transformed the landscape more radically than the preceding five decades. To navigate this dense enumeration of systems, we organize the section around what we call the \emph{comprehension--generation--verification (CGV) triad}: the three irreducible competences a mathematical reasoner must exhibit, parsing the problem (comprehension), producing a candidate solution path (generation), and certifying that the path is in fact valid (verification). The CGV triad refines the two-stage comprehension--generation framing of recent LLM-centric surveys~\cite{wang2025survey,liu2025mathematicallm} by promoting verification from an optional post-hoc filter to a co-equal component that participates both at inference time and during training. As detailed later in Section~\ref{sec:synthesis}, this triad maps directly onto the supervision ladder that drove the era's progression: from mimicking generation, to learning verification, to verifiable search.

\subsection{Comprehension, Generation, and Verification}
Recent LLM-centric surveys usefully distinguish between \emph{mathematical comprehension}, the ability to parse notation, quantities, diagrams, definitions, and problem intent, and \emph{answer generation}, the ability to synthesize a valid solution path and final answer~\cite{wang2025survey}. We extend this framing with a third component: \emph{verification}. For mathematical reasoning, comprehension and generation are necessary but insufficient, because a fluent derivation is not \textit{ipso facto} a valid one. The decisive shift of 2024--2026 is that verification increasingly enters both inference and training, through executable programs, PRMs, symbolic geometry solvers, and proof-assistant kernels.

\begin{table}[t]
\centering
\scriptsize
\renewcommand{\arraystretch}{1.2}
\setlength{\tabcolsep}{4pt}
\begin{tabularx}{\columnwidth}{@{}p{0.18\columnwidth}p{0.16\columnwidth}XX@{}}
\tabletop
\thead{Stage} & \thead{Category} & \thead{Typical mechanisms} & \thead{Common failure modes} \\ \tablemid
\multirow{2}{0.18\columnwidth}{Comprehension} 
  & \textit{Linguistic} & Math-heavy pretraining, numeric representation & Misreading quantities or notation \\
  & \textit{Contextual} & Multimodal encoders, retrieval, diagram parsing & Missing hidden assumptions or geometric constraints \\
\cmidrule(lr){1-4}
\multirow{2}{0.18\columnwidth}{Generation}
  & \textit{Single-path}  & CoT, PoT/PAL, least-to-most, long CoT & Logically invalid or needlessly verbose derivations \\
  & \textit{Multi-path}   & Multi-agent debate, self-consistency & Consensus on incorrect paths, mode collapse \\
\cmidrule(lr){1-4}
\multirow{3}{0.18\columnwidth}{Verification}
  & \textit{Lightweight} & Exact-match graders, Python execution & Checker mismatch, brittle parsing \\
  & \textit{Learned}     & Process reward models (PRMs), agent reviewers & Reward hacking, false positive signals \\
  & \textit{Rigorous} & Theorem provers, expert human audit & High formalization and annotation costs \\ \tablebottom
\end{tabularx}
\caption{A three-component view of LLM mathematical reasoning. The table separates comprehension, generation, and verification mechanisms and lists common failure modes for each component.}
\label{tab:cgverify}
\end{table}

This triad also clarifies the role of long chain-of-thought. Compared with short CoT, long CoT gives the model room for planning, exploration, backtracking, and reflection~\cite{wang2025survey}. Its benefit, however, depends on whether search is coupled to a reliable selector. Long reasoning traces can improve Pass@k and self-correction, but they can also amplify verbosity, hide local mistakes, and spend unnecessary compute. Consequently, the strongest systems increasingly pair longer generation with stronger verification, rather than treating length itself as a proxy for reasoning quality.

\begin{figure}[t]
\centering
\resizebox{\columnwidth}{!}{%
\begin{tikzpicture}[
vtx/.style={
    rectangle,
    rounded corners=1.2mm,
    draw=Indigo!75,
    minimum width=2.15cm,
    minimum height=0.72cm,
    align=center,
    inner sep=2pt,
    font=\footnotesize\bfseries
},
mechs/.style={
    font=\fontsize{5.2}{6.0}\selectfont,
    text=black!65,
    align=left,
    text width=1.95cm
},
el/.style={
    font=\fontsize{5.1}{5.9}\selectfont,
    text=Indigo!85,
    align=center,
    inner sep=1pt
},
pel/.style={
    font=\fontsize{5.1}{5.9}\selectfont,
    text=Magenta!70,
    align=center,
    inner sep=1pt
},
fe/.style={
    -{Latex[length=1.5mm]},
    thick,
    draw=Indigo!65
},
fb/.style={
    -{Latex[length=1.4mm]},
    thick,
    draw=Magenta!60,
    dashed
}
]

\node[vtx,fill=LavenderOne!30] (C) at (-2.1,1.45) {Comprehension};
\node[vtx,fill=LavenderOne!30] (G) at ( 2.1,1.45) {Generation};
\node[vtx,fill=Cyan!17,draw=Violet!75] (V) at (0,-0.35) {Verification};

\node[mechs,anchor=north west] 
    at ([xshift=-0.5cm,yshift=0.0cm]C.south west) {%
    \textbullet\ Math pretraining\\
    \textbullet\ Multimodal enc.\\
    \textbullet\ Retrieval
};

\node[mechs,anchor=north east] 
    at ([xshift=1.2cm,yshift=0cm]G.south east) {%
    \textbullet\ CoT / PoT\\
    \textbullet\ Multi-agent\\
    \textbullet\ Self-improve
};

\node[mechs,anchor=north,text width=2.45cm] 
    at ([yshift=0cm]V.south) {%
    \textbullet\ Exact match / exec.\\
    \textbullet\ PRM / ORM\\
    \textbullet\ Lean kernel
};

\node[
    font=\fontsize{6.0}{6.8}\selectfont,
    text=Indigo!75,
    align=center
] at (0,0.72) {\emph{Math reasoning}\\[-1pt]\emph{quality}};

\draw[fe]
    (C.east) --
    node[el,midway,above=1pt] {\scriptsize parsed intent}
    (G.west);

\draw[fe]
    (G.south west) to[out=-105,in=35]
    node[el,pos=.47,right=1pt] {\scriptsize candidate\\[-1pt]\scriptsize soln.}
    (V.north east);

\draw[fb]
    (V.west) to[out=165,in=-70]
    node[pel,pos=.50,left=3pt] {\scriptsize error\\[-1pt]\scriptsize signal}
    (C.south);

\draw[fb]
    (V.east) to[out=15,in=-110]
    node[pel,pos=.50,right=4pt] {\scriptsize reward}
    (G.south);

\end{tikzpicture}%
}
\caption{The comprehension--generation--verification triad. Solid arrows show forward reasoning; dashed arrows show verification feedback. This extends the two-stage view of recent surveys~\cite{wang2025survey} with an explicit verification loop.}
\label{fig:cgv_triangle}
\end{figure}
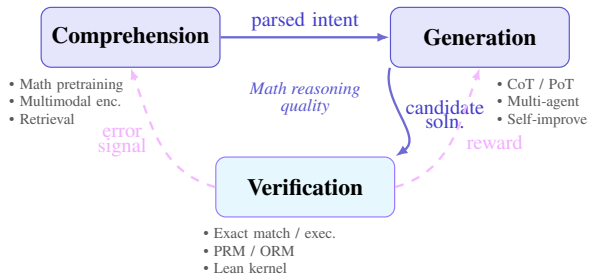
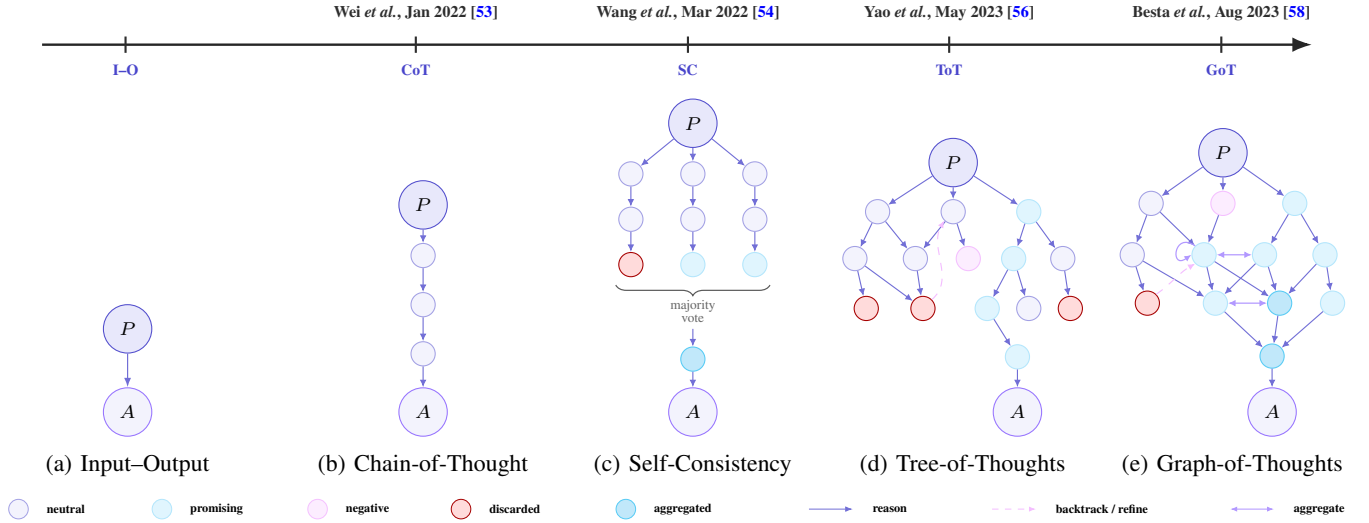
\begin{figure*}[t]
\centering
\captionsetup[subfigure]{justification=centering}

\tikzset{
panel/.style={
    draw=Indigo!68,
    dashed,
    rounded corners=1.5mm,
    line width=0.7pt
},
prob/.style={
    circle,
    draw=Indigo!78,
    fill=LavenderOne!26,
    minimum size=0.64cm,
    inner sep=0pt,
    font=\fontsize{7}{8}\selectfont\bfseries
},
ans/.style={
    circle,
    draw=Violet!78,
    fill=LavenderOne!14,
    minimum size=0.64cm,
    inner sep=0pt,
    font=\fontsize{7}{8}\selectfont\bfseries
},
neutral/.style={
    circle,
    draw=Indigo!42,
    fill=LavenderOne!14,
    minimum size=0.32cm,
    inner sep=0pt
},
good/.style={
    circle,
    draw=Cyan!55,
    fill=Cyan!22,
    minimum size=0.32cm,
    inner sep=0pt
},
agg/.style={
    circle,
    draw=cyan!55,
    fill=cyan!22,
    minimum size=0.32cm,
    inner sep=0pt
},
neg/.style={
    circle,
    draw=Magenta!55,
    fill=Magenta!16,
    minimum size=0.32cm,
    inner sep=0pt
},
disc/.style={
    circle,
    draw=red!65!black,
    fill=red!14,
    minimum size=0.32cm,
    inner sep=0pt
},
mainedge/.style={
    -{Latex[length=1.25mm]},
    draw=Indigo!64,
    line width=0.45pt
},
backedge/.style={
    -{Latex[length=1.15mm]},
    draw=Magenta!58,
    dashed,
    line width=0.45pt
},
aggedge/.style={
    {Latex[length=1.15mm]}-{Latex[length=1.15mm]},
    draw=Violet!58,
    line width=0.48pt
},
plabel/.style={
    font=\fontsize{5.0}{5.8}\selectfont,
    text=black!60,
    align=center
},
timetxt/.style={
    font=\fontsize{6.1}{6.9}\selectfont\bfseries,
    text=black!82,
    align=center
},
timeaux/.style={
    font=\fontsize{4.9}{5.6}\selectfont,
    text=black!60,
    align=center
}
}

\begin{tikzpicture}[x=0.48cm,y=0.6cm]
\draw[-{Latex[length=2.6mm]}, line width=0.9pt, draw=black!85] (0,0) -- (35,0);

\foreach \x in {2.3,10.3,17.8,25.0,32.5}{
    \draw[line width=0.8pt, draw=black!82] (\x,-0.18) -- (\x,0.18);
}

\node[timetxt] at (10.3,0.7) {Wei \textit{et al.}, Jan 2022 \cite{wei2022chain}};
\node[timetxt] at (17.8,0.7) {Wang \textit{et al.}, Mar 2022 \cite{wang2023selfconsistency}};
\node[timetxt] at (25.0,0.7) {Yao \textit{et al.}, May 2023 \cite{yao2023tree}};
\node[timetxt] at (32.5,0.7) {Besta \textit{et al.}, Aug 2023 \cite{besta2024graph}};

\node[font=\fontsize{5.8}{6.6}\selectfont\bfseries,text=Indigo!82] at (2.3,-0.55)  {I--O};
\node[font=\fontsize{5.8}{6.6}\selectfont\bfseries,text=Indigo!82] at (10.3,-0.55) {CoT};
\node[font=\fontsize{5.8}{6.6}\selectfont\bfseries,text=Indigo!82] at (17.8,-0.55) {SC};
\node[font=\fontsize{5.8}{6.6}\selectfont\bfseries,text=Indigo!82] at (25.0,-0.55) {ToT};
\node[font=\fontsize{5.8}{6.6}\selectfont\bfseries,text=Indigo!82] at (32.5,-0.55) {GoT};
\end{tikzpicture}

\vspace{0.55em}

\noindent
\begin{subfigure}[t]{0.19\textwidth}
\centering
\begin{tikzpicture}[x=1cm,y=1cm]
\node[prob] (p) at (0,1.20) {$P$};
\node[ans]  (a) at (0,0.10) {$A$};
\draw[mainedge] (p) -- (a);
\end{tikzpicture}
\caption{Input--Output}
\label{fig:topo_io}
\end{subfigure}
\hspace{0.012\textwidth}
\begin{subfigure}[t]{0.19\textwidth}
\centering
\begin{tikzpicture}[
    x=1cm,y=1cm,
    cotnode/.style={
        circle,
        draw=Indigo!42,
        fill=LavenderOne!14,
        minimum size=0.32cm,
        inner sep=0pt
    },
    cotedge/.style={
        -{Latex[length=1.0mm,width=0.9mm]},
        draw=Indigo!64,
        line width=0.42pt
    }
]

\node[prob]    (p)  at (0,1.72) {$P$};
\node[cotnode] (t1) at (0,1.05) {};
\node[cotnode] (t2) at (0,0.40) {};
\node[cotnode] (t3) at (0,-0.25) {};
\node[ans]     (a)  at (0,-1.02) {$A$};

\draw[cotedge] (p)  -- (t1);
\draw[cotedge] (t1) -- (t2);
\draw[cotedge] (t2) -- (t3);
\draw[cotedge] (t3) -- (a);


\end{tikzpicture}
\caption{Chain-of-Thought}
\label{fig:topo_cot}
\end{subfigure}
\begin{subfigure}[t]{0.19\textwidth}
\centering
\begin{tikzpicture}[
    x=1cm,y=1cm,
    scedge/.style={
        -{Latex[length=1.0mm,width=0.9mm]},
        draw=Indigo!64,
        line width=0.42pt
    },
    votelabel/.style={
        font=\fontsize{5.0}{5.8}\selectfont,
        text=black!60,
        align=center
    }
]

\node[prob] (p) at (0,1.72) {$P$};

\node[neutral] (l1) at (-0.82,1.05) {};
\node[neutral] (l2) at (-0.82,0.45) {};
\node[disc]    (l3) at (-0.82,-0.15) {};   

\node[neutral] (m1) at (0,1.05) {};
\node[neutral] (m2) at (0,0.45) {};
\node[good]    (m3) at (0,-0.15) {};       

\node[neutral] (r1) at (0.82,1.05) {};
\node[neutral] (r2) at (0.82,0.45) {};
\node[good]    (r3) at (0.82,-0.15) {};    

\node[agg] (sel) at (0,-1.4) {};

\node[ans] (a) at (0,-2.1) {$A$};

\draw[scedge] (p) -- (l1);
\draw[scedge] (p) -- (m1);
\draw[scedge] (p) -- (r1);

\draw[scedge] (l1) -- (l2);
\draw[scedge] (l2) -- (l3);

\draw[scedge] (m1) -- (m2);
\draw[scedge] (m2) -- (m3);

\draw[scedge] (r1) -- (r2);
\draw[scedge] (r2) -- (r3);

\draw[
    decorate,
    decoration={brace,mirror,amplitude=4pt},
    draw=black!65,
    line width=0.45pt
] (-1.02,-0.42) -- (1.02,-0.42);

\node[votelabel] (vote) at (0,-0.76) {majority\\[-1pt]vote};

\draw[scedge] (0,-1) -- (sel);
\draw[scedge] (sel) -- (a);


\end{tikzpicture}
\caption{Self-Consistency}
\label{fig:topo_sc}
\end{subfigure}
\begin{subfigure}[t]{0.19\textwidth}
\centering
\begin{tikzpicture}[
    x=1cm,y=1cm,
    totedge/.style={
        -{Latex[length=1.0mm,width=0.9mm]},
        draw=Indigo!64,
        line width=0.42pt
    },
    totdash/.style={
        -{Latex[length=0.95mm,width=0.85mm]},
        draw=Magenta!58,
        dashed,
        line width=0.42pt
    }
]

\node[prob] (p) at (0,1.95) {$P$};

\node[neutral] (t1) at (-1.00,1.30) {};
\node[neutral] (t2) at ( 0.00,1.30) {};
\node[good]    (t3) at ( 1.00,1.30) {};

\node[neutral] (u1) at (-1.30,0.68) {};
\node[neutral] (u2) at (-0.50,0.68) {};
\node[neg]     (u3) at ( 0.20,0.68) {};
\node[good]    (u4) at ( 0.80,0.68) {};
\node[neutral] (u5) at ( 1.45,0.68) {};

\node[disc]    (v1) at (-1.15,0.02) {};
\node[disc]    (v2) at (-0.40,0.02) {};
\node[good]    (v3) at ( 0.45,0.02) {};
\node[neutral] (v4) at ( 1.00,0.02) {};
\node[disc]    (v5) at ( 1.55,0.02) {};

\node[good]    (w1) at (0.85,-0.62) {};

\node[ans]     (a)  at (0.85,-1.35) {$A$};

\draw[totedge] (p) -- (t1);
\draw[totedge] (p) -- (t2);
\draw[totedge] (p) -- (t3);

\draw[totedge] (t1) -- (u1);
\draw[totedge] (t1) -- (u2);

\draw[totedge] (t2) -- (u2);
\draw[totedge] (t2) -- (u3);

\draw[totedge] (t3) -- (u4);
\draw[totedge] (t3) -- (u5);

\draw[totedge] (u1) -- (v1);
\draw[totedge] (u1) -- (v2);

\draw[totedge] (u2) -- (v2);

\draw[totedge] (u4) -- (v3);
\draw[totedge] (u4) -- (v4);

\draw[totedge] (u5) -- (v5);

\draw[totedge] (v3) -- (w1);

\draw[totedge] (w1) -- (a);

\draw[totdash] (v2.north east) to[out=40,in=-120] (t2.south west);

\end{tikzpicture}
\caption{Tree-of-Thoughts}
\label{fig:topo_tot}
\end{subfigure}
\begin{subfigure}[t]{0.19\textwidth}
\centering
\begin{tikzpicture}[
    x=1cm,y=1cm,
    gotedge/.style={
        -{Latex[length=1.0mm,width=0.9mm]},
        draw=Indigo!64,
        line width=0.42pt
    },
    gotback/.style={
        -{Latex[length=0.95mm,width=0.85mm]},
        draw=Magenta!58,
        dashed,
        line width=0.42pt
    },
    gotagg/.style={
        {Latex[length=1.0mm,width=0.9mm]}-{Latex[length=1.0mm,width=0.9mm]},
        draw=Violet!58,
        line width=0.46pt
    },
    gotloop/.style={
        -{Latex[length=0.95mm,width=0.85mm]},
        draw=Violet!58,
        line width=0.46pt
    }
]

\node[prob] (p) at (0,1.95) {$P$};

\node[neutral] (t1) at (-0.95,1.28) {};
\node[neg]     (t2) at ( 0.00,1.28) {};
\node[good]    (t3) at ( 0.95,1.28) {};

\node[neutral] (u1) at (-1.20,0.60) {};
\node[good]    (u2) at (-0.25,0.60) {};
\node[good]    (u3) at ( 0.55,0.60) {};
\node[good]    (u4) at ( 1.35,0.60) {};

\node[disc]    (v1) at (-1.00,-0.04) {};
\node[good]    (v2) at (-0.10,-0.04) {};
\node[agg]     (v3) at ( 0.75,-0.04) {};
\node[good]    (v4) at ( 1.45,-0.04) {};

\node[agg]     (w1) at (0.65,-0.74) {};

\node[ans]     (a)  at (0.65,-1.48) {$A$};

\draw[gotedge] (p) -- (t1);
\draw[gotedge] (p) -- (t2);
\draw[gotedge] (p) -- (t3);

\draw[gotedge] (t1) -- (u1);
\draw[gotedge] (t1) -- (u2);

\draw[gotedge] (t2) -- (u2);

\draw[gotedge] (t3) -- (u3);
\draw[gotedge] (t3) -- (u4);

\draw[gotedge] (u1) -- (v1);
\draw[gotedge] (u1) -- (v2);

\draw[gotedge] (u2) -- (v2);
\draw[gotedge] (u2) -- (v3);

\draw[gotedge] (u3) -- (v2);
\draw[gotedge] (u3) -- (v3);

\draw[gotedge] (u4) -- (v3);
\draw[gotedge] (u4) -- (v4);

\draw[gotedge] (v2) -- (w1);
\draw[gotedge] (v3) -- (w1);
\draw[gotedge] (v4) -- (w1);

\draw[gotedge] (w1) -- (a);

\draw[gotloop] (u2.north west) to[out=145,in=215,looseness=7] (u2.west);

\draw[gotback] (v1.north east) to[out=35,in=-150] (u2.south west);

\draw[gotagg] (u2) -- (u3);
\draw[gotagg] (v2) -- (v3);

\end{tikzpicture}
\caption{Graph-of-Thoughts}
\label{fig:topo_got}
\end{subfigure}

\vspace{0.65em}

\resizebox{0.98\textwidth}{!}{%
\begin{tikzpicture}[x=1cm,y=1cm]
\node[neutral] at (0,0) {};
\node[font=\fontsize{5.7}{6.4}\selectfont\bfseries,anchor=west] at (0.34,0) {neutral};

\node[good] at (2.35,0) {};
\node[font=\fontsize{5.7}{6.4}\selectfont\bfseries,anchor=west] at (2.69,0) {promising};

\node[neg] at (4.90,0) {};
\node[font=\fontsize{5.7}{6.4}\selectfont\bfseries,anchor=west] at (5.24,0) {negative};

\node[disc] at (7.25,0) {};
\node[font=\fontsize{5.7}{6.4}\selectfont\bfseries,anchor=west] at (7.59,0) {discarded};

\node[agg] at (9.95,0) {};
\node[font=\fontsize{5.7}{6.4}\selectfont\bfseries,anchor=west] at (10.29,0) {aggregated};

\draw[mainedge] (12.95,0) -- +(0.72,0);
\node[font=\fontsize{5.7}{6.4}\selectfont\bfseries,anchor=west] at (13.87,0) {reason};

\draw[backedge] (15.95,0) -- +(0.72,0);
\node[font=\fontsize{5.7}{6.4}\selectfont\bfseries,anchor=west] at (16.87,0) {backtrack / refine};

\draw[aggedge] (19.85,0) -- +(0.72,0);
\node[font=\fontsize{5.7}{6.4}\selectfont\bfseries,anchor=west] at (20.77,0) {aggregate};
\end{tikzpicture}%
}

\caption{Evolution of prompting topologies for mathematical reasoning. The upper timeline sketches the shift from direct input--output prompting to chain-based, self-consistent, tree-structured, and graph-structured reasoning. The subfigures below use a unified visual language: the problem is represented by the circular node $P$, the final answer by $A$, and intermediate reasoning states by smaller circular nodes. Cyan nodes indicate promising or aggregated states; magenta/red nodes indicate negatively scored or discarded states.}
\label{fig:prompting_topology_evolution}
\end{figure*}
\subsection{Prompting-era Innovations}
\textit{This subsection concerns the generation and comprehension components of the triad.}

\paragraph*{Chain-of-Thought prompting} \cite{wei2022chain}~demonstrated that, when prompted with a handful of exemplars containing intermediate reasoning steps (``chains of thought''), sufficiently large language models exhibit emergent multi-step reasoning capability. On GSM8K, prompting PaLM-540B with eight chain-of-thought (CoT) exemplars yielded state-of-the-art performance, outperforming task-specific fine-tuned systems. \cite{kojima2022large}~further showed that a simple zero-shot trigger phrase, ``Let's think step by step'', suffices to elicit similar behavior in sufficiently capable models.

\paragraph*{Self-consistency} \cite{wang2023selfconsistency}~introduced a decoding strategy that samples a diverse set of reasoning paths from a CoT-prompted model and selects the most consistent final answer by majority vote. Self-consistency yielded double-digit percentage-point gains on GSM8K ($+17.9\%$), SVAMP ($+11.0\%$), and AQuA ($+12.2\%$), establishing a simple but foundational result: reasoning quality scales with the budget allocated to inference-time sampling.

\paragraph*{Problem decomposition} \cite{zhou2023leasttomost}~proposed \emph{least-to-most prompting}, in which an LLM first decomposes a complex problem into a sequence of simpler sub-problems and then solves them in order. \emph{Tree of Thoughts}~\cite{yao2023tree} generalizes CoT to a search over a tree of intermediate states, using the LM itself as both thought generator and state evaluator, and combining with breadth-first or depth-first search. More elaborate scaffolds couple LLMs with Monte Carlo Tree Search over reasoning steps, drawing an explicit analogy with AlphaGo-style game-tree search.

\paragraph*{Tool-integrated reasoning} A parallel line of work addresses a fundamental limitation of pure LLM reasoning: LLMs are unreliable numerical calculators. Program-Aided Language Models (PAL)~\cite{gao2023pal} and Program of Thoughts (PoT)~\cite{chen2022program} have the LLM translate a math problem into executable Python code, delegating the actual computation to a deterministic interpreter. On GSM8K, PAL with Codex achieved $72.0\%$ top-1 accuracy, surpassing PaLM-540B with CoT by 15 absolute points. A recent survey of code-enhanced reasoning usefully separates this family into single-execution code generation, dynamic code--language interleaving, non-executable program representations, and training-time code supervision~\cite{yang2025codethink}. This taxonomy matters for mathematics because code is not merely a calculator: its structured syntax, modular decomposition, executable semantics, and error feedback make it an intermediate representation that is both generative and partially verifiable. \textsc{ToRA}~\cite{gou2024tora} interleaves natural-language reasoning with tool invocation in a single loop, supervised by trajectories distilled from GPT-4 and refined via output-space shaping. Tool-integrated reasoning has become the \textit{de facto} choice for any LLM-based math system when calculator-like precision is required.

\paragraph*{Semantic understanding and error reduction} While CoT and tool-integrated methods address calculation errors, a complementary failure mode, \emph{semantic misunderstanding} of the problem statement, persists even when arithmetic is correct. \textsc{DUP} (Deeply Understanding the Problems)~\cite{zhong2024dup} directly targets this class of errors through a three-stage prompting protocol: (i)~extract the core question from the problem text, filtering irrelevant background; (ii)~identify only the information relevant to that core question; and (iii)~generate the solution conditioned on both. On GSM8K, \textsc{DUP} achieves $97.1\%$ zero-shot accuracy, substantially outperforming standard CoT, and ablations confirm that the gains arise specifically from reduced semantic misinterpretation rather than improved calculation. This result is important for the survey's verification emphasis (see~\S\ref{sec:verification}): semantic errors are precisely the class that execution-based verifiers \emph{cannot} catch, because a program that faithfully encodes a misunderstood problem will execute correctly but produce the wrong answer. Methods like \textsc{DUP} thus complement tool-integrated reasoning by operating upstream of the computation.

\begin{table*}[t]
\centering
\scriptsize
\setlength{\tabcolsep}{2.5pt}
\renewcommand{\arraystretch}{1.08}
\resizebox{\textwidth}{!}{%
\begin{tabular}{p{0.13\textwidth}p{0.10\textwidth}p{0.20\textwidth}rrrrrrrrr}
\tabletop
\thead{Method} & \thead{Model} & \thead{Settings} & \thead{GSM8K} & \thead{GSM-H} & \thead{SVAMP} & \thead{ASDiv} & \thead{SingleEq} & \thead{AddSub} & \thead{MultiArith} & \thead{MATH} & \thead{AQuA} \\ \tablemid
\tgroup{12}{\textit{Direct / Chain-of-Thought prompting}}
Direct         & Codex              & Few-shot direct prompting       & 19.7  & 5.0   & 69.9  & 74.0  & 86.8  & 90.9  & 44.0  & --    & 29.5 \\
               & PaLM-540B          & Few-shot CoT                    & 56.9  & --    & 79.0  & 73.9  & 92.3  & 91.9  & 94.7  & 8.8   & 35.8 \\
CoT            & GPT-4              & Few-shot CoT                    & 92.0  & --    & 97.0  & --    & --    & --    & --    & 42.5  & 72.4 \\
               & GPT-4o-mini        & 0-shot CoT                      & --    & --    & --    & --    & --    & --    & --    & 50.6  & --   \\
               & GPT-3.5            & 0-shot CoT                      & 81.6  & --    & 78.2  & --    & 93.1  & 86.1  & 96.7  & --    & --   \\
\textsc{DUP}~\cite{zhong2024dup} & GPT-4 & 0-shot semantic decomposition   & 97.1  & --    & 94.2  & --    & 96.0  & 95.1  & 98.1  & --    & 77.1 \\
\tgroup{12}{\textit{Program-aided / tool-integrated reasoning}}
PAL            & Codex              & Few-shot program-aided LM       & 72.0  & 61.2  & 79.4  & 79.6  & 96.1  & 92.5  & 99.2  & --    & --   \\
               & GPT-3.5            & Few-shot program-aided LM       & 80.6  & --    & 79.5  & --    & 97.6  & 89.1  & 97.0  & --    & --   \\
               & GPT-4o-mini        & 0-shot program-aided LM         & --    & --    & --    & --    & --    & --    & --    & 36.6  & --   \\
PoT            & Codex              & Few-shot program of thought     & 71.6  & 61.8  & 85.2  & 85.2  & --    & 92.2  & 99.5  & --    & 54.1 \\
               & Codex              & Program of thought + SC         & 80.0  & --    & 89.1  & --    & --    & --    & --    & --    & 58.6 \\
               & GPT-4              & Few-shot program of thought     & 97.2  & --    & 97.4  & --    & --    & --    & --    & --    & 84.4 \\
\textsc{ToRA}  & Llama2-70B         & 0-shot tool-integrated          & 84.3  & 67.2  & 82.7  & 86.8  & --    & --    & --    & 49.7  & --   \\
\tgroup{12}{\textit{Code interleaving / program representations}}
MathCoder      & Llama2-70B         & 0-shot code interleaving        & 83.9  & --    & 84.9  & --    & --    & --    & --    & 45.1  & --   \\
MathCoder2     & CodeLlama-34B      & 0-shot code interleaving        & 81.7  & --    & 82.5  & --    & --    & --    & --    & 45.2  & --   \\
CodePlan       & Mistral-7B         & Few-shot code-form planning     & 59.5  & --    & 61.4  & --    & --    & --    & --    & 34.3  & --   \\
INC-Math       & GPT-4o-mini        & 0-shot code prompting           & --    & --    & --    & --    & --    & --    & --    & 51.4  & --   \\
CoC            & text-davinci-003   & Chain of Code                   & 71.0  & --    & --    & --    & --    & --    & --    & --    & --   \\
CodePrompt     & GPT-3.5            & Few-shot code prompting         & 80.6  & --    & 79.6  & --    & --    & --    & --    & --    & --   \\
\tablebottom
\end{tabular}%
}
\caption{Performance of representative prompting and code-aided reasoning methods on classical MWP and competition benchmarks. Scores are reported accuracies (\%) from the corresponding source papers. Dashes indicate that a result was not reported under the matching model and prompting/tool-use setting. GSM-H abbreviates GSM-HARD.}
\label{tab:code_aided}
\end{table*}

\paragraph*{The code--mathematics overlap}
PAL, PoT, and \textsc{ToRA} are presented in this survey as
tool-use strategies, but they also represent the convergence
of mathematical reasoning with code generation.  Models
trained on code (\textit{e.g.}, Codex, Code Llama, \textsc{Llemma})
consistently outperform text-only models of similar size on
math benchmarks, even without explicit math fine-tuning.
This is not coincidental:\ programming and mathematics share
a requirement for precise symbolic manipulation, state
tracking, compositional reasoning, and executable checking.
The same properties also explain the failure mode of code-aided
reasoning: execution can certify that a program ran correctly,
but not that it formalized the intended problem. The overlap implies
that code-generation benchmarks (\textsc{HumanEval},
\textsc{MBPP}, \textsc{LiveCodeBench}) and mathematical
benchmarks are not independent evaluations but partially
redundant measures of a shared underlying capability.
Future work should clarify whether strong performance on one
reliably predicts strong performance on the other, and
whether joint training on code and mathematical proofs yields
synergies beyond what either domain provides alone.

\paragraph*{Self-improvement and bootstrapping} A complementary line of work asks whether models can improve their own reasoning traces rather than merely consume human-written rationales. \textsc{STaR} bootstraps reasoning by generating rationales from a few examples, retaining or rationalizing those that lead to correct answers, and then fine-tuning on the resulting traces~\cite{zelikman2022star}. \textsc{Quiet-STaR} generalizes this idea beyond question answering by training models to infer latent rationales inside arbitrary text before predicting difficult tokens~\cite{zelikman2024quietstar}. \textsc{V-STaR} observes that failed self-generated solutions are also informative: it uses both correct and incorrect candidates to train a verifier, which then selects among many proposed solutions at inference time~\cite{hosseini2024vstar}. \textsc{ReFT} combines SFT warm-up with online reinforcement learning over automatically sampled reasoning paths, using ground-truth answers as task rewards for mathematical reasoning~\cite{luong2024reft}. \textsc{LIMO} pushes the data-efficiency side of the same agenda, arguing that carefully selected ``cognitive templates'' can elicit strong mathematical reasoning from a sufficiently knowledgeable foundation model with surprisingly little SFT data~\cite{ye2025limo}. \textsc{SCoRe} shifts from rationale generation to self-correction, training a model through multi-turn reinforcement learning to improve its own second attempt after an initial answer~\cite{kumar2024score}. A particularly striking result in this vein is \textsc{rStar-Math}~\cite{guan2025rstarmath}, which demonstrates that small language models (1.5B--7B parameters) can rival frontier models on competition-level mathematics by iteratively self-evolving through Monte Carlo Tree Search: the model generates solution candidates, a co-trained process reward model scores each reasoning step, and the verified rollouts are used to improve both the policy and the reward model in successive rounds, all without distilling traces from a larger teacher. Together, these methods bridge prompting, SFT, verification, and RL: they treat the model's own attempts as a renewable training source, but their success still depends on reliable filters such as exact answers, verifiers, or external rewards.

\subsection{Multi-Agent and Agentic Mathematical Reasoning}
\textit{This subsection concerns the generation and verification components of the triad.}
A distinct inference-time strand treats mathematical reasoning not as a single trace but as a collaborative process among multiple LLM agents. Early multi-agent debate work~\cite{du2023multiagentdebate} asked several model instances to propose answers, exchange arguments over multiple rounds, and converge on a common final answer; it reported gains on mathematical and strategic reasoning while reducing hallucinated factual claims. The \textsc{MAD} framework~\cite{liang2024mad} sharpened this idea into adversarial ``tit-for-tat'' debate with a judge, motivated by the observation that self-reflection can degenerate when a model becomes overconfident in an initially wrong solution.

For mathematics, the central benefit of multi-agent systems is diversity: agents can explore different solution paths, check each other's arithmetic, and expose hidden assumptions. \textsc{ReConcile}~\cite{chen2024reconcile} made this explicit through a round-table protocol in which diverse LLM agents discuss grouped answers, confidence scores, and answer-rectifying explanations before a confidence-weighted vote; the paper reports an 8\% gain on \textsc{MATH} and finds model diversity to be a key driver of improvement. \textsc{DyLAN}~\cite{liu2024dylan} introduced dynamic team optimization, selecting agents by an unsupervised Agent Importance Score and then allowing the selected team to communicate through a task-specific dynamic network, improving arithmetic reasoning with moderate compute.

The next wave moves from discussion to structured orchestration. \textsc{Mixture-of-Agents}~\cite{wang2024moa} stacks layers of LLM agents, where each layer conditions on the outputs of the previous one, demonstrating that heterogeneous model collaboration can outperform a single strong model but at substantial token and latency cost. \textsc{Graph-of-Agents}~\cite{yun2026goa} addresses this cost by selecting only the most relevant agents from model-card metadata, constructing directed edges from peer relevance scores, passing messages from high-relevance to lower-relevance agents and back, and then pooling the refined responses. On the \textsc{MATH} benchmark, \textsc{GoA-Mean} reaches $73.12\%$ with three selected agents, compared with $71.60\%$ for six-agent Refine and $65.80\%$ for six-agent \textsc{MoA}; on MMLU-Pro it also reduces calls and tokens relative to \textsc{MoA}. A complementary \emph{hierarchical-orchestrator} topology, a project-coordinator agent supervising workstream-coordinator agents that in turn dispatch specialised sub-agents and reviewer agents, is used by the AI co-mathematician of Zheng et al.~\cite{zheng2026comathematician} to scale to multi-day open-ended research workflows rather than single-problem solves. The lesson across these designs is not simply ``more agents,'' but better routing, relevance scoring, and aggregation.

Math-specific multi-agent systems increasingly combine collaboration with verifiers. \textsc{MAgICoRe}~\cite{chen2025magicore} uses Solver, Reviewer, and Refiner agents, with step-wise reward-model scores guiding targeted feedback; it beats self-consistency by 3.4\%, Best-of-$k$ by 3.2\%, and Self-Refine by 4.0\% across five math datasets while using fewer samples. \textsc{Mars-PO}~\cite{lou2024marspo} turns multi-agent outputs into preference-optimization data by building shared positive samples and agent-specific negative samples, raising Llama3.1-8B-Instruct on \textsc{MATH} from $50.38\%$ to $57.82\%$. \textsc{MALT}~\cite{motwani2025malt} divides reasoning into heterogeneous generation, verification, and refinement roles, then propagates rewards through a multi-agent search tree; it reports relative improvements of $15.66\%$ on \textsc{MATH} and $7.42\%$ on \textsc{GSM8K}. Finally, \textsc{MATTRL}~\cite{hu2026mattrl} injects test-time experiences into multi-expert deliberation and consensus, reporting gains over both single-agent and multi-agent baselines across medicine, math, and education.

\begin{table}[t]
\centering
\scriptsize
\renewcommand{\arraystretch}{1.32}
\setlength{\tabcolsep}{3.5pt}
\providecommand{\stripe}[1]{\textcolor{#1}{\rule[-.45ex]{1.6pt}{1.7ex}}\hspace{2pt}}
\begin{tabularx}{\columnwidth}{@{}p{0.20\columnwidth}>{\raggedright\arraybackslash}p{0.215\columnwidth}>{\columncolor{Cyan!9}\raggedright\arraybackslash}X>{\columncolor{Magenta!9}\raggedright\arraybackslash}X@{}}
\tabletop
\thead{Protocol} & \thead{Representative systems} & \thead{Mathematical role } & \thead{Main risk } \\ \tablemid
\stripe{VivLime!60!black}Debate / consensus           & Debate, \textsc{MAD}, \textsc{ReConcile}             & Diverse attempts with confidence-weighted voting                    & Judge bias or consensus on a shared error \\
\stripe{VivOrange!70!black}Dynamic routing             & \textsc{DyLAN}, \textsc{GoA}                          & Select task-relevant specialists and route messages                  & Bad agent selection or noisy relevance scores \\
\stripe{VivPink!75!black}Role pipelines                & \textsc{MAgICoRe}, \textsc{MALT}                      & Solver--reviewer--refiner or generator--verifier--refiner workflows & Reviewers may miss subtle mathematical errors \\
\stripe{VivLilac!80!black}Training from teams          & \textsc{Mars-PO}, \textsc{MALT}                       & Convert multi-agent traces into preference or SFT data               & Preference pairs may encode correlated flaws \\
\stripe{VivPurple!85!black}Test-time RL                & \textsc{MATTRL}                                        & Reuse turn-level experience during deliberation                     & Credit assignment and cost remain open problems \\
\tablebottom
\end{tabularx}
\caption{Multi-agent protocols for mathematical reasoning, with each row paired against its dominant failure mode. The cyan-tinted column collects the architectural \emph{benefit} of each protocol; the magenta-tinted column collects the corresponding \emph{risk}, so any row reads as a single trade-off. Colored stripes on the Protocol column follow the paradigm palette of Figure~\ref{fig:timeline}. The final row marks the emerging test-time-learning frontier, whose benefit and risk are still the least well understood empirically.}
\label{tab:multi_agent}
\end{table}

Reading across Table~\ref{tab:multi_agent}, a clear pattern emerges: protocols that rely solely on dialogue, debate, consensus voting, are vulnerable to correlated errors whenever agents share a training distribution, because majority voting cannot correct a mistake that all voters reproduce. Protocols that introduce external verification into the loop (role pipelines with PRM scoring, solver--reviewer--refiner workflows) trade communication cost for reliability, but their gains plateau when the reviewer itself lacks the domain-specific knowledge to catch subtle mathematical errors. The strongest emerging results come from systems that combine \emph{both} agent diversity \emph{and} a verifiable checkpoint, such as \textsc{MAgICoRe}'s reward-model-guided refinement or \textsc{MALT}'s multi-agent search tree with propagated rewards. This analysis suggests a practical guideline: multi-agent collaboration is cost-effective when (i)~agents bring genuinely different capabilities or training data rather than being copies of the same model, (ii)~the task decomposes into independently verifiable subtasks, and (iii)~routing or relevance scoring controls communication cost. When these conditions are not met, single-agent sampling with a strong verifier typically achieves the same diversity benefits at lower token cost (see~\S\ref{sec:verification}).

A natural next question is whether the benefits of multi-agent debate can be \emph{internalized} into a single model, avoiding the token and latency cost of explicit inter-agent communication at inference time. Recent work on \emph{debate distillation} pursues exactly this idea: complete multi-agent debate transcripts are used as supervised training data, or diverse multi-agent trajectories are converted into preference pairs for DPO/GRPO fine-tuning, so that the student model learns to reproduce the self-correcting, multi-perspective reasoning pattern of the teacher ensemble in a single forward pass. Frameworks such as \textsc{DMAD} (Diverse Multi-Agent Debate) break the ``mental set'' of a single model by forcing agents to adopt distinct reasoning strategies during the distillation process, and interaction-graph-based methods compress the agent communication structure into compact representations for student training. Complementary mechanistic work on ``latent agents'' investigates whether reasoning models trained via RLVR already develop internal agent-like subspaces, functionally distinct reasoning modes that activate on different problem types, suggesting that long chain-of-thought models may be performing implicit multi-agent debate without explicit orchestration. These findings have implications for both efficiency (debate-distilled single models can match multi-agent accuracy at a fraction of the token cost) and safety (internalized debate is harder to audit than explicit inter-agent transcripts, raising questions about oversight and interpretability in high-stakes mathematical verification).

\begin{figure*}[t]
\centering
\captionsetup[subfigure]{justification=centering}

\tikzset{
ag/.style={
    circle,
    draw=Indigo!70,
    fill=LavenderOne!35,
    minimum size=0.55cm,
    inner sep=0pt,
    font=\fontsize{6.5}{7.5}\selectfont\bfseries
},
agoff/.style={
    circle,
    draw=black!22,
    fill=black!5,
    text=black!30,
    minimum size=0.55cm,
    inner sep=0pt,
    font=\fontsize{6.5}{7.5}\selectfont\bfseries
},
jg/.style={
    circle,
    draw=Violet!75,
    fill=Magenta!15,
    minimum size=0.55cm,
    inner sep=0pt,
    font=\fontsize{6.5}{7.5}\selectfont\bfseries
},
rt/.style={
    diamond,
    draw=cyan!75,
    fill=cyan!20,
    minimum size=0.45cm,
    inner sep=0pt,
    aspect=1.4,
    font=\fontsize{6}{7}\selectfont\bfseries
},
msg/.style={
    -{Latex[length=1.3mm]},
    draw=Indigo!55,
    line width=.35pt
},
mb/.style={
    {Latex[length=1.1mm]}-{Latex[length=1.1mm]},
    draw=Indigo!55,
    line width=.35pt
},
mr/.style={
    -{Latex[length=1.3mm]},
    draw=Violet!60,
    line width=.4pt
},
faint/.style={
    -{Latex[length=1.3mm]},
    draw=black!32,
    dashed,
    line width=.35pt
},
sd/.style={
    font=\fontsize{5.5}{6.3}\selectfont,
    text=black!55,
    align=center,
    text width=2.8cm
},
sl/.style={
    font=\fontsize{7.5}{9}\selectfont\bfseries,
    text=Indigo!80
},
lb/.style={
    rectangle,
    rounded corners=.6mm,
    draw=black!18,
    fill=Indigo!4,
    minimum width=2.4cm,
    minimum height=.6cm
},
edgetxt/.style={
    font=\fontsize{4.8}{5.5}\selectfont,
    text=black!50
}
}

\begin{subfigure}[t]{0.24\textwidth}
\centering
\begin{tikzpicture}
\node[ag] (a1) at (0,1.3) {$A_1$};
\node[ag] (a2) at (2.4,1.3) {$A_2$};
\node[ag] (a3) at (1.2,0) {$A_3$};
\node[jg] (j)  at (1.2,-1.0) {$J$};

\draw[mb] (a1) -- (a2);
\draw[mb] (a2) -- (a3);
\draw[mb] (a1) -- (a3);

\draw[msg] (a1.south) to[out=-70,in=155] (j.north west);
\draw[msg] (a2.south) to[out=-110,in=25] (j.north east);
\draw[msg] (a3) -- (j);

\node[sd] at (1.2,-1.75) {All-to-all exchange;\\judge selects};
\end{tikzpicture}
\caption{Debate}
\label{fig:multiagent_topology_debate}
\end{subfigure}
\hfill
\begin{subfigure}[t]{0.24\textwidth}
\centering
\begin{tikzpicture}
\node[ag] (s) at (0,1.3) {$S$};
\node[ag] (r) at (1.2,.4) {$R$};
\node[ag] (f) at (2.4,-.5) {$F$};

\draw[msg,thick] (s) -- node[right=1pt,edgetxt] {draft} (r);
\draw[msg,thick] (r) -- node[right=1pt,edgetxt] {critique} (f);
\draw[msg,dashed,draw=Magenta!50]
    (f.west) .. controls +(-1.2,.2) and +(-.8,-.4) ..
    node[left=1pt,font=\fontsize{4.8}{5.5}\selectfont,text=Magenta!60] {refine}
    (s.south);

\node[sd] at (1.2,-1.75) {Solver$\to$Reviewer$\to$ Refiner + feedback};
\end{tikzpicture}
\caption{Pipeline}
\label{fig:multiagent_topology_pipeline}
\end{subfigure}
\hfill
\begin{subfigure}[t]{0.24\textwidth}
\centering
\begin{tikzpicture}
\node[lb] at (1.2,1.5) {};
\node[ag,minimum size=.4cm] (l1a) at (.35,1.5) {$A_1$};
\node[ag,minimum size=.4cm] (l1b) at (1.2,1.5) {$A_2$};
\node[ag,minimum size=.4cm] (l1c) at (2.05,1.5) {$A_3$};

\node[lb] at (1.2,.4) {};
\node[ag,minimum size=.4cm] (l2a) at (.35,.4) {$B_1$};
\node[ag,minimum size=.4cm] (l2b) at (1.2,.4) {$B_2$};
\node[ag,minimum size=.4cm] (l2c) at (2.05,.4) {$B_3$};

\node[jg,minimum size=.45cm] (agg) at (1.2,-.6) {\tiny Ag};

\foreach \s in {l1a,l1b,l1c}{
    \foreach \d in {l2a,l2b,l2c}{
        \draw[msg,draw=Indigo!25,line width=.2pt] (\s) -- (\d);
    }
}
\foreach \s in {l2a,l2b,l2c}{
    \draw[msg,draw=Indigo!35] (\s) -- (agg);
}

\node[sd] at (1.2,-1.75) {Layered DAG;\\each layer reads all prior};
\end{tikzpicture}
\caption{MoA}
\label{fig:multiagent_topology_moa}
\end{subfigure}
\hfill
\begin{subfigure}[t]{0.24\textwidth}
\centering
\begin{tikzpicture}
\node[rt] (rtr) at (1.2,1.6) {\tiny Rt};

\node[ag]    (g1) at (0,.6) {$A_1$};
\node[ag]    (g2) at (1.2,.15) {$A_2$};
\node[agoff] (g3) at (2.4,.6) {$A_3$};
\node[agoff] (g4) at (-0.15,-0.62) {$A_4$};  

\draw[mr,thick] (rtr) -- (g1);
\draw[mr,thick] (rtr) -- (g2);
\draw[faint] (rtr) -- (g3);
\draw[faint] (rtr) -- (g4);

\draw[mb,thick,draw=Violet!50] (g1) -- (g2);

\node[jg,minimum size=.45cm] (pl) at (1.2,-1.0) {\tiny Pl};

\draw[msg,draw=Violet!50] (g1.south east) to[out=-28,in=155] (pl.north west);
\draw[msg,draw=Violet!50] (g2) -- (pl);

\node[sd] at (1.2,-1.75) {Router selects relevant\\subset; sparse msgs};
\end{tikzpicture}
\caption{Graph-of-Agents}
\label{fig:multiagent_topology_goa}
\end{subfigure}

\caption{Four multi-agent topologies for mathematical reasoning. (a)~Debate: fully connected rounds with a judge~\cite{du2023multiagentdebate,chen2024reconcile}. (b)~Role pipeline: Solver--Reviewer--Refiner chain with feedback~\cite{chen2025magicore,motwani2025malt}. (c)~Mixture-of-Agents: layered DAG~\cite{wang2024moa}. (d)~Graph-of-Agents: router selects relevant agents; grayed nodes are unselected~\cite{yun2026goa}. The progression reflects the shift from ``more agents'' toward routing and verification-aware orchestration.}
\label{fig:multiagent_topology}
\end{figure*}
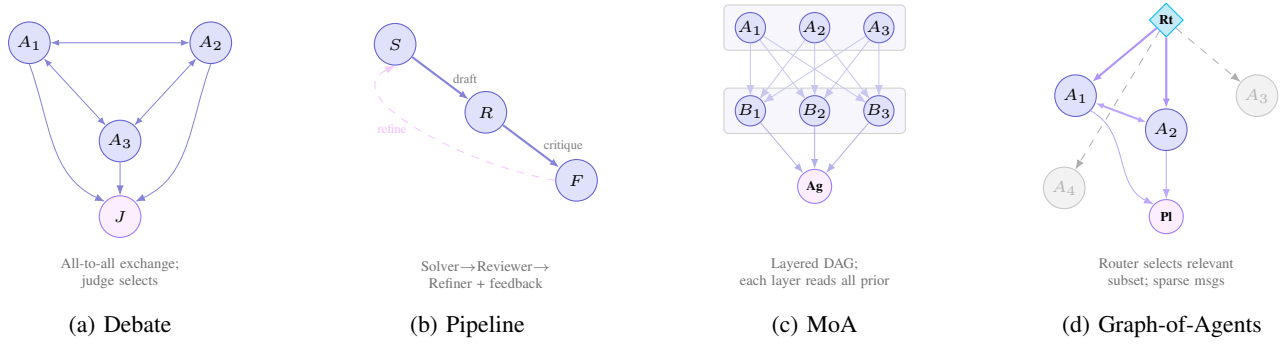
\subsection{Math-specialized Foundation Models}
\textit{This subsection primarily concerns the generation component of the triad.}
Beyond prompting, a second strand of work adapts the LLM pretraining pipeline itself to mathematics.

\textsc{Minerva}~\cite{lewkowycz2022solving} continued pretraining PaLM (8B/62B/540B) on a carefully curated mixture of arXiv papers, web mathematics, and math-related code, demonstrating that a moderate amount of domain-specific pretraining dramatically improves performance on MATH and MMLU-STEM without requiring external tools.

\textsc{Llemma}~\cite{azerbayev2024llemma} extended this recipe in open-source form. By continuing Code Llama on \textsc{Proof-Pile-2}, a 55B-token mixture including \textsc{AlgebraicStack}, \textsc{OpenWebMath}, and the arXiv subset of RedPajama, the 7B and 34B models match or exceed the unreleased Minerva models on an equi-parameter basis. Crucially, Llemma displays emergent tool-use and few-shot theorem-proving abilities in Lean 4 without any further fine-tuning, foreshadowing the unification of informal and formal tracks.

\textsc{DeepSeekMath}~\cite{shao2024deepseekmath} introduced a 120B-token high-quality math pretraining corpus mined from Common Crawl via an iteratively-retrained fastText classifier. \textsc{DeepSeekMath-Base 7B} reaches $64.2\%$ on GSM8K and $36.2\%$ on MATH, surpassing the closed-source Minerva 540B with roughly $1/77$ the parameter count. The paper also introduced \emph{Group Relative Policy Optimization} (GRPO), a critic-free RL algorithm that became the standard fine-tuning recipe for reasoning training in 2024--2025 (though subsequent systems including Kimi k1.5 and o3 moved toward DAPO and PPO variants).

\textsc{Qwen2-Math} and \textsc{Qwen2.5-Math}~\cite{yang2024qwen2math} refined the self-improvement pipeline further, using iteratively bootstrapped problem-solution pairs to train process reward models that supervise subsequent policy rounds. \textsc{InternLM-Math}~\cite{ying2024internlm} and \textsc{MetaMath}~\cite{yu2024metamath} pushed synthetic data generation, with MetaMath in particular showing that simple backward-question rewriting augmentation substantially boosts competition math performance. \textsc{MAmmoTH}~\cite{yue2024mammoth} unifies chain-of-thought and program-of-thought training data, arguing that exposure to both formats is necessary for robust math generalization. \textsc{WizardMath}~\cite{luo2023wizardmath} applies Reinforced Evol-Instruct to amplify problem difficulty iteratively.

\subsection{Verifiers and Process Reward Models}
\textit{This subsection concerns the verification component of the triad.}
The verifier idea introduced by~\cite{cobbe2021training} was dramatically extended by ``Let's Verify Step by Step''~\cite{lightman2023verify}, which showed that process reward models (PRMs), trained to score each intermediate step in a chain of thought, significantly outperform outcome reward models (ORMs) that score only the final answer, reaching $78\%$ on a representative MATH subset. OpenAI released the \textsc{PRM800K} dataset of 800{,}000 step-level correctness labels. \textsc{Math-Shepherd}~\cite{wang2024mathshepherd} removed the dependence on costly human annotation by estimating step correctness via Monte Carlo rollouts: a step is judged correct if the expected accuracy of completions starting from it is high. \textsc{OmegaPRM}~\cite{luo2024improve} accelerated this process via a divide-and-conquer MCTS algorithm, producing over 1.5M process annotations without human effort. These PRMs serve both as rerankers at inference time and as reward models for subsequent reinforcement-learning fine-tuning.

\paragraph*{Process \textit{vs.}\ outcome supervision: a direct comparison} The PRM--ORM distinction deserves explicit attention because it maps directly onto the survey's central thesis that stronger verification yields better systems. Uesato et al.~\cite{uesato2022solving} provided the earliest controlled comparison, showing that process-based feedback produces more reliable solvers than outcome-based feedback on GSM8K, particularly when the generator is strong enough that most errors are subtle single-step mistakes rather than wholesale misunderstanding. Lightman et al.~\cite{lightman2023verify} scaled this finding, demonstrating that a PRM selecting among 1{,}860 sampled solutions outperforms a best-of-$N$ ORM by a significant margin on MATH. The intuition is straightforward: an ORM can distinguish correct from incorrect final answers but cannot localize \emph{where} the reasoning went wrong; a PRM can, and this localization enables both better reranking and more targeted RL credit assignment. However, PRMs incur higher annotation cost (human or rollout-based), and recent work on \textsc{OmegaPRM} and Math-Shepherd suggests that the gap narrows when rollout-based PRM labels are abundant and cheap. The practical implication is a cost--accuracy trade-off: for budget-constrained deployments, ORM reranking with many samples may suffice; for high-stakes applications requiring traceable reasoning, PRM scoring or execution-based checking is essential.

\paragraph*{Lightweight inference-time verification} A growing family of practical systems combines CoT and PoT reasoning with lightweight verifiers that require neither human annotation nor formal proof assistants. The simplest recipe pairs a CoT generator with a Python executor that checks numerical consistency: the model generates both a natural-language derivation and executable code, and the system accepts the answer only if both agree. More sophisticated variants use a small PRM or a fine-tuned verifier to score intermediate steps, reject solutions with low-confidence transitions, and trigger re-generation. These lightweight verification pipelines occupy a middle ground between pure sampling (cheap but unreliable) and full formal verification (reliable but expensive), and they are increasingly the default configuration for deployed mathematical reasoning systems.

\subsection{The Reasoning-Model Era (2024--2026)}
\textit{This subsection concerns the synthesis of generation and verification through RL.}
In September~2024, OpenAI released \textsc{o1}~\cite{openai2024o1}, a model explicitly trained to produce long, reflective chains of thought before answering. OpenAI reported two scaling laws that would define the subsequent eighteen months of research: performance improves with \emph{train-time compute} dedicated to RL on verifiable rewards, and it improves further with \emph{test-time compute} spent generating and evaluating intermediate reasoning. On AIME~2024, GPT-4o averaged $12\%$, while \textsc{o1} averaged $74\%$ with single-sample greedy decoding, $83\%$ with 64-sample majority voting, and $93\%$ when reranking 1{,}000 samples with a learned scoring function, placing it above the USAMO cutoff.

In January~2025, DeepSeek released \textsc{R1}~\cite{guo2025deepseekr1}, which demonstrated that comparable reasoning behavior could be elicited in an open-weights base model via pure reinforcement learning on verifiable rewards, with no supervised fine-tuning on human reasoning traces. Notably, \textsc{R1}-Zero exhibited emergent ``aha moments'' and self-correction patterns purely from reward signal. Kimi k1.5~\cite{team2025kimi} achieved similar results with outcome-based and generative reward models, and \textsc{s1}~\cite{muennighoff2025s1} showed that test-time scaling could be elicited with as little as 1{,}000 curated reasoning traces. Snell et al.~\cite{snell2024scaling} formalized the theoretical underpinning, showing that optimal test-time compute allocation can be more effective than scaling model parameters.

By mid-2026 the frontier had moved into the next generation of closed-weights reasoning systems, \textsc{GPT-5.5 High} and \textsc{Gemini 3.1 Pro Thinking}, and a matching wave of open-weights efforts. \textsc{SU-01}~\cite{li2026su01} is a representative open release: a 30B activated-3B MoE backbone trained with SFT on $\sim$340K sub-8K-token trajectories followed by 200 RL steps, but capable of stable reasoning over trajectories exceeding 100K tokens at inference. With test-time scaling, \textsc{SU-01} reaches gold-medal-level scores on IMO~2025 (35/42) and USAMO~2026 (35/42), and clears the gold line on IPhO~2024 and IPhO~2025, while \textsc{GPT-5.5 High} reports $92.5\%$/$92.9\%$ on AIME~25/26 and $80.7\%$ overall on \textsc{IMO-ProofBench}~\cite{luong2025imoproofbench}, a recent benchmark that grades proof writing rather than final answers, on which \textsc{Gemini 3.1 Pro Thinking} scores $72.6\%$ and \textsc{DeepSeek-V3.2-Speciale} $45.7\%$. The persistent ranking on \textsc{IMO-ProofBench}'s advanced split ($64.8\%$, $50.0\%$, $28.6\%$ respectively) confirms that even at the 2026 frontier, proof-grade reasoning remains a distinguishing axis, not yet saturated by either training scale or test-time compute.

The RL algorithms underlying these reasoning models have evolved rapidly, and recent ACM survey work frames the progression as a move from critic-based RLHF toward increasingly lightweight critic-free online optimization~\cite{wang2025survey}. PPO, the workhorse of ChatGPT-era RLHF, uses a learned value or critic model to estimate advantages, which improves stability but substantially increases memory, compute, and tuning cost~\cite{schulman2017ppo}. ReMax removes this learned critic by using a greedy response as a REINFORCE-style baseline, exploiting the trajectory-level reward structure of LLM alignment~\cite{li2024remax}. RLOO further improves baseline estimation by sampling multiple responses to the same prompt and using a leave-one-out reward mean for each candidate~\cite{ahmadian2024back}. GRPO, introduced with \textsc{DeepSeekMath}, samples a group of completions and normalizes each reward relative to the group's mean and standard deviation, which made critic-free RLVR especially attractive for mathematical reasoning~\cite{shao2024deepseekmath}. REINFORCE++ revisits the same critic-free family with global advantage normalization, aiming to reduce the bias and instability introduced by prompt-local normalization~\cite{hu2025reinforcepp}. DAPO then specializes the recipe for long-CoT reasoning at scale by combining decoupled clipping, dynamic sampling, token-level policy-gradient loss, and overlong-response shaping; its Clip-Higher component relaxes the upper clipping bound to encourage exploration and delay entropy collapse~\cite{yu2025dapo}. Offline methods such as DPO and Step-DPO provide simpler preference-optimization alternatives, including step-wise variants for long-chain reasoning, but they do not collect new rollouts during optimization and therefore give up part of the exploration that makes online RLVR useful under distribution shift~\cite{rafailov2023dpo,lai2024stepdpo}. The practical consensus by mid-2025 was consequently not that one optimizer had solved reasoning training, but that online RL with verifiable rewards had become the dominant path for pushing mathematical reasoning beyond SFT and static preference data.
\begin{figure*}[t]
\centering
\resizebox{0.9\textwidth}{!}{%
\begin{tikzpicture}[
alg/.style={
    rectangle,
    rounded corners=1.0mm,
    draw=Indigo!70,
    fill=LavenderOne!24,
    minimum width=1.42cm,
    minimum height=0.78cm,
    align=center,
    inner sep=1.6pt,
    font=\fontsize{6.2}{7.0}\selectfont
},
hot/.style={
    rectangle,
    rounded corners=1.0mm,
    draw=Indigo!70,
    fill=Cyan!14,
    minimum width=1.42cm,
    minimum height=0.78cm,
    align=center,
    inner sep=1.6pt,
    font=\fontsize{6.2}{7.0}\selectfont
},
off/.style={
    rectangle,
    rounded corners=1.0mm,
    draw=Magenta!60,
    fill=Magenta!7,
    minimum width=1.42cm,
    minimum height=0.78cm,
    align=center,
    inner sep=1.6pt,
    font=\fontsize{6.2}{7.0}\selectfont
},
d/.style={
    font=\fontsize{4.9}{5.7}\selectfont,
    text=black!62,
    align=center
},
lane/.style={
    font=\fontsize{5.8}{6.8}\selectfont\bfseries,
    align=center
},
mf/.style={
    -{Latex[length=1.5mm]},
    thick,
    draw=Indigo!60
},
of/.style={
    -{Latex[length=1.35mm]},
    thick,
    draw=Magenta!50
},
bf/.style={
    -{Latex[length=1.3mm]},
    draw=Magenta!45,
    dashed,
    line width=0.8pt
},
note/.style={
    rectangle,
    rounded corners=.8mm,
    draw=black!22,
    fill=white,
    text=black!65,
    text width=3.0cm,
    align=center,
    inner sep=3pt,
    font=\fontsize{5.1}{6.0}\selectfont
}
]

\node[alg] (ppo) at (0.0,0.0) {\textbf{PPO}\\[1pt]{\tiny Schulman '17}\\[1pt]{\tiny\cite{schulman2017ppo}}};
\node[alg] (rmx) at (2.05,0.0) {\textbf{ReMax}\\[1pt]{\tiny Li \textit{et al.}\ '24}\\[1pt]{\tiny\cite{li2024remax}}};
\node[alg] (rlo) at (4.10,0.0) {\textbf{RLOO}\\[1pt]{\tiny Ahmadian '24}\\[1pt]{\tiny\cite{ahmadian2024back}}};
\node[hot] (grp) at (6.15,0.0) {\textbf{GRPO}\\[1pt]{\tiny DeepSeekMath}\\[1pt]{\tiny\cite{shao2024deepseekmath}}};
\node[alg,font=\fontsize{4.6}{5.5}\selectfont] (rpp) at (8.20,0.0) {\textbf{REINFORCE}\\[0.5pt]\textbf{++}\\[1pt]{\tiny Hu '25}\\[1pt]{\tiny\cite{hu2025reinforcepp}}};
\node[hot] (dap) at (10.25,0.0) {\textbf{DAPO}\\[1pt]{\tiny Yu \textit{et al.}\ '25}\\[1pt]{\tiny\cite{yu2025dapo}}};

\node[d] at (0.0,-0.78) {Critic-based};
\node[d] at (2.05,-0.78) {Greedy baseline};
\node[d] at (4.10,-0.78) {Leave-one-out MC};
\node[d] at (6.15,-0.78) {Group-normalized};
\node[d] at (8.20,-0.78) {Global advantage};
\node[d] at (10.25,-0.78) {Clip-Higher};

\draw[mf] (ppo) -- (rmx);
\draw[mf] (rmx) -- (rlo);
\draw[mf] (rlo) -- (grp);
\draw[mf] (grp) -- (rpp);
\draw[mf] (rpp) -- (dap);

\node[
    lane,
    text=Indigo!75,
    anchor=east
] at (-0.72,0.0) {Online RL\\(RLVR)};

\node[off] (dpo) at (2.35,-1.95) {\textbf{DPO}\\[1pt]{\tiny Rafailov '23}\\[1pt]{\tiny\cite{rafailov2023dpo}}};
\node[off] (sdp) at (4.40,-1.95) {\textbf{Step-DPO}\\[1pt]{\tiny Lai \textit{et al.}\ '24}\\[1pt]{\tiny\cite{lai2024stepdpo}}};

\node[d] at (2.35,-2.73) {Implicit reward};
\node[d] at (4.40,-2.73) {Step-wise pref.};

\draw[bf]
    ([xshift=1pt]ppo.south)
    .. controls +(0,-1.10) and +(-0.90,0.80) ..
    (dpo.north west);

\draw[of] (dpo) -- (sdp);

\node[
    lane,
    text=Magenta!65,
    anchor=east
] at (1.18,-1.95) {Offline\\(pref.)};

\node[note] (note) at (7.45,-1.95) {Offline methods sacrifice exploration.\\
Online RLVR dominates for math\\
reasoning by mid-2025~\cite{wang2025survey}.};

\end{tikzpicture}%
}
\caption{Evolution of RL algorithms for reasoning-model training. The main track (top) shows the progression from critic-based PPO to critic-free online methods. The offline DPO branch (bottom, dashed) provides simpler preference-based alternatives. Cyan boxes mark the most widely adopted algorithms for reasoning training.}
\label{fig:rl_evolution}
\end{figure*}
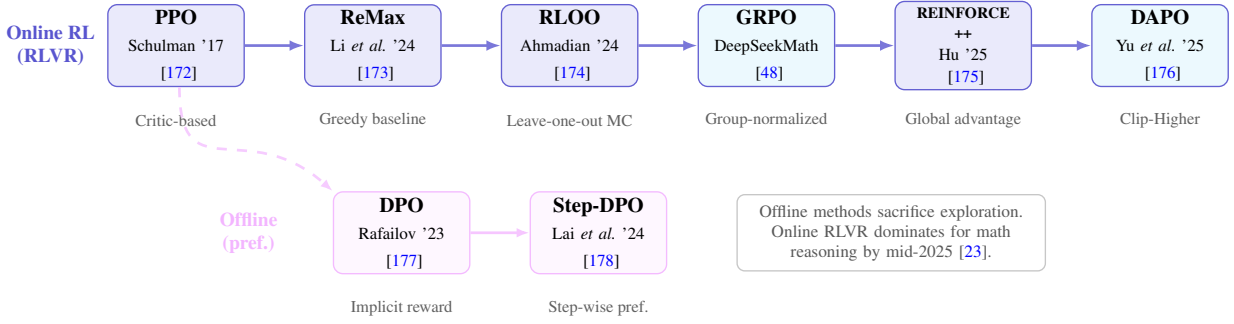

In April~2025, OpenAI released \textsc{o3} and \textsc{o4-mini}, with \textsc{o3} reaching $91.6\%$ on AIME~2024, $88.9\%$ on AIME~2025, $86.8\%$ on MathVista, and approximately $25.2\%$ on FrontierMath (versus $<2\%$ for previous frontier models). At IMO~2025, an advanced version of Gemini with Deep Think~\cite{deepmind2025geminideepthink} operated end-to-end in natural language on the official problem statements within the 4.5-hour competition time limit and solved five out of six problems for $35/42$ points, achieving gold-medal performance in the first such officially graded evaluation of an AI system. OpenAI's experimental system achieved a comparable result. These milestones closed a gap that, only 18~months earlier, required problem-by-problem formalization and multi-day compute to breach.

Table~\ref{tab:reasoning} summarizes key results across representative reasoning models.

\begin{table*}[t]
\centering
\scriptsize
\setlength{\tabcolsep}{2.5pt}
\renewcommand{\arraystretch}{1.10}
\resizebox{\textwidth}{!}{%
\begin{tabular}{p{0.19\textwidth}lrrrrrrr}
\tabletop
\thead{Model} & \thead{Params} & \thead{MATH-500} & \thead{GSM8K} & \thead{AIME'24} & \thead{AIME'25} & \thead{FMath} & \thead{GPQA-D} \\ \tablemid
\tgroup{8}{\textit{General-purpose frontier LLMs}}
GPT-4o                                    & --       & 76.6  & 94.6  & 12.0  & --     & --       & 53.6  \\
Claude 3.5 Sonnet                         & --       & 78.3  & 96.4  & --    & --     & --       & 59.4  \\
Gemini 1.5 Pro                            & --       & 67.7  & --    & --    & --     & --       & 59.1  \\
\tgroup{8}{\textit{Math-specialized open-weight LLMs}}
DeepSeekMath-RL 7B                        & 7B       & 51.7  & 88.2  & --    & --     & --       & --    \\
Qwen2.5-Math 72B-Inst                     & 72B      & 85.1  & 95.9  & 30.0  & --     & --       & --    \\
InternLM2-Math-Plus 7B                    & 7B       & 53.0  & 87.2  & --    & --     & --       & --    \\
\tgroup{8}{\textit{Reasoning models (long CoT + RLVR)}}
OpenAI o1                                & --       & 94.8  & 97.1  & 83.0  & 79.2   & $<$2    & 78.0  \\
DeepSeek-R1                              & 671B     & 97.3  & --    & 79.8  & 70.0   & --       & 71.5  \\
Kimi k1.5                                & --       & 96.2  & --    & 77.5  & --     & --       & --    \\
s1 (32B, 1K traces)                      & 32B      & 93.0  & --    & --    & --     & --       & 59.6  \\
QwQ-32B-Preview                          & 32B      & 90.6  & --    & 50.0  & --     & --       & 65.2  \\
Grok 3 Beta                              & --       & --    & --    & 83.9  & 77.3   & --       & 80.2  \\
OpenAI o3                                & --       & --    & --    & 91.6  & 88.9   & 25.2    & 83.3  \\
OpenAI o4-mini (high)                    & --       & --    & --    & 91.7$^\dagger$ & 92.7 & --  & 81.4  \\
Gemini 2.5 Pro                           & --       & 98.8  & --    & 87.5$^\dagger$ & 86.7 & --  & 84.0  \\
Qwen3 (235B-A22B, thinking)              & 235B     & 98.0  & --    & 85.7  & 81.5   & --       & 71.1  \\
\tgroup{8}{\textit{Live leaderboard snapshots (accessed May~2026)}}
Gemini 3 Pro\textsuperscript{\S}          & --       & --    & --    & --    & 100.0  & --       & 91.9  \\
GPT-5.2 Pro\textsuperscript{\S}            & --       & --    & --    & --    & 100.0  & --       & 93.2  \\
Claude Mythos Preview\textsuperscript{\S}  & --       & --    & --    & --    & --     & --       & 94.6  \\
LongCat-Flash-Thinking\textsuperscript{\S} & 560B     & 99.2  & --    & --    & 90.6   & --       & --    \\
Kimi K2 Instruct\textsuperscript{\S}       & 1.0T     & 97.4  & 97.3  & --    & --     & --       & --    \\
\tgroup{8}{\textit{Distilled reasoning models}}
DeepSeek-R1-Distill-Qwen-32B             & 32B      & 94.3  & --    & 72.6  & --     & --       & 62.1  \\
DeepSeek-R1-Distill-Qwen-7B              & 7B       & 92.8  & --    & 55.5  & --     & --       & 49.1  \\
DeepScaleR 1.5B                           & 1.5B     & --    & --    & 43.1  & --     & --       & --    \\
LIMO 32B                                  & 32B      & 95.6  & --    & 63.3  & --     & --       & 70.7  \\
\tablebottom
\end{tabular}%
}
\caption{Comparison of frontier, math-specialized, reasoning, and distilled models. GPQA-D = GPQA-Diamond; FMath = FrontierMath; MATH-500 is the 500-problem subset of \textsc{MATH} commonly used in recent leaderboards. Dashes indicate unreported numbers; self-reported single-shot accuracy except where noted. $^\dagger$\textsc{MathArena} pass@1 scores~\cite{balunovic2026matharena}. \textsuperscript{\S}Rows marked this way are live leaderboard snapshots from LLM Stats, accessed May~10,~2026, and should be read as provisional self-reported comparisons rather than controlled paper evaluations~\cite{llmstats2026math}.}
\label{tab:reasoning}
\end{table*}
\subsection{Inference-Time Scaling as Search}
\textit{This subsection concerns test-time scaling for generation and verification.}
The central conceptual change of the reasoning-model era is that inference is no longer a single forward pass. Instead, a difficult problem is treated as a search problem over possible derivations. Chain-of-thought prompting expands the state space; self-consistency samples multiple trajectories; multi-agent systems distribute exploration across solvers, reviewers, verifiers, and routers; process reward models score partial states; tool-integrated systems externalize arithmetic or symbolic manipulation; and formal provers use the proof assistant itself to prune invalid branches. This does not make neural systems equivalent to classical search engines, but it moves them closer to the long-standing AI view of problem solving as guided exploration.

This perspective explains why test-time compute has become a first-class variable. If a model can generate many partially independent solution attempts, then selection quality becomes almost as important as generation quality. Majority voting helps when errors are uncorrelated; multi-agent deliberation helps when agents bring genuinely different priors or roles; PRM reranking helps when local step quality predicts final correctness; execution helps when the desired output is a program; and Lean checking helps when the desired output is a proof. The open question is how to allocate the reasoning budget adaptively: trivial arithmetic should not consume thousands of tokens, while a hard olympiad inequality may require many failed approaches before a useful invariant appears.

Snell et al.~\cite{snell2024scaling} showed that optimal
test-time compute allocation can be more effective than
scaling model parameters, but this result is stated as a
general principle rather than a quantitative law.  The
open question is whether mathematical reasoning admits
scaling exponents analogous to the Chinchilla laws for
pretraining: does doubling the number of sampled reasoning
traces yield a predictable reduction in error rate, and does
this exponent depend on problem difficulty, domain, or
verification quality?  Preliminary evidence from
self-consistency experiments (where accuracy scales roughly
as $1 - e^{-\alpha k}$ with sample count $k$ under mild
error-independence assumptions) and from the PRM reranking
literature (where gains plateau once the best-of-$k$ ceiling
is reached) suggests that such laws exist but are
domain-dependent and verifier-dependent. A first analytical step toward such laws is taken by Levi~\cite{levi2025inference}, who shows that when per-problem failure probabilities follow a heavy-tailed (Beta) difficulty distribution, the residual pass@$k$ failure rate decays as a power law in $k$ whose exponent is governed by the concentration of hard problems---a functional form that closely matches the empirical coverage curves of Brown et al.~\cite{brown2024monkeys} on \textsc{MATH}. On this account, inference-scaling exponents exist but are properties of the model--benchmark pair, \emph{pace} any hope for a universal constant. Formalizing these
relationships would enable principled compute budgeting:
allocating more search to harder problems and less to easier
ones, rather than applying a uniform inference budget.

\paragraph*{Reporting standards: pass@$k$ \textit{vs.}\ greedy \textit{vs.}\ majority vote}
A recurring source of confusion in the literature is that
reported accuracies conflate fundamentally different
inference regimes.  Three distinct metrics should be
distinguished: (i)~\emph{greedy / pass@1}, the accuracy of a
single deterministic forward pass, which measures the
policy's modal behavior; (ii)~\emph{majority vote @$k$}
(self-consistency), which samples $k$ independent traces and
returns the plurality answer, measuring whether correct
solutions are more probable than any single incorrect one;
and (iii)~\emph{pass@$k$}, the probability that at least one
of $k$ samples is correct, which measures the generator's
\emph{coverage} and is relevant when a reliable verifier is
available.  The distinction matters enormously:
\textsc{o1}'s AIME~2024 accuracy rises from $74\%$
(greedy) to $83\%$ (majority@64) to $93\%$ (best-of-1000 +
learned scorer), and \textsc{DeepSeek-Prover-V2}'s
MiniF2F score rises from $\sim$60\% (pass@1) to $88.9\%$
(pass@8192).  These numbers reflect qualitatively different
capabilities, policy quality, output diversity, and
verifier strength, respectively, and comparing them across
systems without controlling for the sampling budget and
selection mechanism is misleading.  We recommend that future
benchmarking reports include, at minimum, pass@1 (greedy),
majority@$k$ for a standardized $k$ (\textit{e.g.}, 64), and the
total token budget per problem.

\subsection{Why Verification Entered the Mainstream}\label{sec:verification}
 
The progression from prompting to tool use to multi-agent
orchestration to formal verification is not a sequence of
independent inventions; it is driven by a single recurring
failure.  Each generation of informal reasoning systems
reached a performance ceiling at which \emph{generation
quality alone could no longer be distinguished from
generation fluency}.  Chain-of-thought prompting produced
plausible but arithmetically wrong derivations; tool
integration fixed arithmetic but introduced program
faithfulness errors; multi-agent debate filtered some of
these errors but remained vulnerable to correlated
mistakes among agents sharing the same training
distribution.
 
The decisive response was to bring in an external judge
whose correctness criterion is independent of the
generator's training objective.  For arithmetic, this judge
is a Python interpreter.  For symbolic geometry, it is a
deductive database like DDAR.  For formal mathematics, it is
the Lean kernel.  In each case, the judge converts
mathematical reasoning from a language-modeling problem into
a search-and-verify problem, fundamentally changing the
scaling dynamics:\ more inference compute yields more
candidates, and a reliable judge ensures that at least one
correct candidate is selected.
 
This analysis also clarifies the conditions under which
multi-agent systems are preferable to single-agent scaling.
Table~\ref{tab:multi_agent} lists risks for each protocol
but does not synthesize them.  The emerging consensus is
that multi-agent collaboration helps when (i)~agents bring
genuinely different priors or capabilities (\textit{e.g.}, a symbolic
solver and a neural proposer), (ii)~the task decomposes into
subtasks with independently checkable outputs, and
(iii)~communication cost is controlled by routing or
relevance scoring rather than all-to-all
exchange~\cite{yun2026goa,chen2025magicore}.  When these
conditions are not met, for instance, when all agents are
instances of the same model, single-agent sampling with
a strong verifier (PRM reranking or execution) is typically
more cost-effective than
debate~\cite{snell2024scaling,wang2024moa}.

\begin{table*}[t]
\centering
\scriptsize
\renewcommand{\arraystretch}{1.18}
\setlength{\tabcolsep}{5pt}
\begin{tabularx}{0.96\textwidth}{@{}p{0.16\textwidth}p{0.18\textwidth}X X@{}}
\tabletop
\thead{Selection signal} & \thead{Artifact scored} & \thead{Main advantage} & \thead{Typical failure mode} \\ \tablemid
\textbf{Majority vote} & Final answer & Simple, model-agnostic selector; effective when correct traces form the plurality & Correlated wrong traces can dominate the vote \\
\textbf{Agent consensus} & Multiple agent responses & Exploits model, prompt, or role diversity across independent attempts & Irrelevant or redundant agents add noise and cost \\
\textbf{ORM} & Completed solution & Cheap final-answer scoring; easy to pair with best-of-$k$ sampling & Misses locally invalid steps that happen to end in a plausible answer \\
\textbf{PRM} & Intermediate steps & Localizes errors and supports targeted reranking or reinforcement learning & Requires step-level human labels or reliable rollout-based annotations \\
\textbf{Execution} & Program trace & Provides deterministic arithmetic, symbolic, or unit-test checking & The program may faithfully solve a misunderstood problem \\
\textbf{Lean kernel} & Formal proof & Supplies kernel-level proof verification and reusable certificates & Formalization, premise retrieval, and proof search remain costly \\ \tablebottom
\end{tabularx}
\caption{Common inference-time selection mechanisms for mathematical reasoning systems. The rows progress from weak final-answer selectors to process-, execution-, and kernel-level checkers. Stronger checkers make generated traces more useful for training and auditing, but they also require more structured interfaces and higher verification cost.}
\label{tab:selection}
\end{table*}

\section{Multimodal and Geometry Problem Solving}\label{sec:multimodal}
Systems discussed in this section correspond to the ``Geometry \& Multimodal'' band in Figure~\ref{fig:timeline}.
Geometry problems have, from the beginning of the field, occupied a special position as the paradigmatic multimodal mathematical task. We cover both the classical symbolic tradition and the new wave of vision--language systems.

\subsection{Classical Geometry Problem Solving}
Early work on geometry understanding dates back to the 1980s with efficient pattern-detection methods on geometric diagrams~\cite{lin1985efficient} and engines such as \textsc{GeoRep}~\cite{ferguson2000georep} for spatial descriptions of line drawings. \textsc{Beatrix}~\cite{bulko1988understanding,novak1990understanding} was an early multimodal system able to parse both English text and diagrams. The authors of~\cite{seo2014diagram} pioneered the combination of text and diagrams for geometry problem solving by maximizing the agreement between the two modalities, a design choice that anticipated subsequent multimodal fusion schemes. \cite{hopkins2019semeval} organized a SemEval shared task on math question answering; \cite{sachan2020discourse} examined the multimedia nature of geometry textbooks.

\textsc{Geos}~\cite{seo2015solving} was the first system to introduce a formal-language description for geometry questions. It over-generates a set of relations, scores them, and chooses a subset maximizing the combined text-and-diagram score, solving
\begin{equation}\label{eq:25}
L^* = \operatorname{argmax}\big(\lambda\cdot A(L',t,d) + H(L',t,d)\big),
\end{equation}
where $A$ measures affinity between the question text and literal set $L$, and $H$ measures coherence. \textsc{Geos} could solve SAT plane-geometry problems but was limited by its 186-problem dataset. \textsc{Geos++}~\cite{sachan2017textbooks,sachan2020discourse} extended the formalism by incorporating 293 axiomatic theorems as horn-clause rules. \textsc{GeoShader}~\cite{alvin2017synthesis} and \textsc{Geos-OS}~\cite{sachan-xing-2017-learning} contributed further datasets and models.

\textsc{InterGPS}~\cite{lu2021inter} introduced the large \textsc{Geometry3K} dataset (3{,}002 problems with dense formal-language annotations) and combines rule-based text parsing with neural object detection, a Transformer-based Theorem Predictor, and a Symbolic Geometry Problem Solver applying theorems to compute the final answer. It set the state of the art on \textsc{Geos}, \textsc{Geos++}, and \textsc{Geometry3K}. \textsc{Ngs}~\cite{chen2021geoqa} introduced the \textsc{GeoQA} dataset of 5{,}010 problems with interpretable programs and multiple auxiliary tasks to enhance cross-modal representation. \textsc{GeometryQA}~\cite{tsai2021geometryqa} re-annotated a geometry subset of \textsc{Math23K}~\cite{wang2018translating} with associated formulas.

The 2023--2026 plane-geometry literature sharpened the lesson that diagram grounding, not just theorem search, is the bottleneck. \textsc{PGPSNet}~\cite{zhang2023pgpsnet} introduced \textsc{PGPS9K}, a 9{,}022-problem dataset with fine-grained diagram annotations and interpretable solution programs, and converted diagrams into structural and semantic textual clauses for multimodal fusion. \textsc{GeoDRL}~\cite{peng2023geodrl} framed theorem application as a Markov decision process over a geometry logic graph, using reinforcement learning while a symbolic system maintained deductive correctness. \textsc{LANS}~\cite{li2024lans} added layout-aware pretraining and point-guided fusion, showing that geometric layout carries information missed by image-level features alone. \textsc{Pi-GPS}~\cite{zhao2025pigps} pushed this further by using an MLLM rectifier plus a geometric verifier to resolve underspecified points, shapes, and shaded areas in the text before theorem prediction, achieving a nearly 10\% improvement over prior neural-symbolic approaches on \textsc{Geometry3K}. In parallel, \textsc{FGeo-HyperGNet}~\cite{zhang2025fgeohypergnet} integrates the \textsc{FormalGeo} symbolic system with a hypergraph neural theorem predictor, making the predict--apply loop readable and traceable.

\textsc{AutoGPS}~\cite{ping2025autogps} synthesizes several of these threads into a unified neuro-symbolic framework that produces \emph{verifiable stepwise proofs}. Its Multimodal Problem Formalizer (MPF) uses neural cross-modal comprehension to translate diagram--text pairs into structured formal representations, and a Deductive Symbolic Reasoner (DSR) expands a hypergraph of derivations to produce minimal, human-readable solution steps. A feedback loop between the MPF and DSR corrects formalization errors during derivation, and the symbolic engine guarantees that each step is logically sound. On \textsc{PGPS9K} completion tasks, \textsc{AutoGPS} improves over prior SOTA by $9.2\%$, and human evaluators judge $99\%$ of its derivation steps logically correct, compared to $71\%$ for the best-performing MLLM baseline. The result is important because it demonstrates that geometry, unlike free-form MWP solving, already admits end-to-end verifiable reasoning \emph{without} a full proof assistant: the symbolic engine serves as a lightweight kernel, and the neural component handles the perceptual and search-planning tasks that symbolic systems do poorly alone.

\subsection{Neural Olympiad-Level Geometry}
The most striking geometry result of the decade was \textsc{AlphaGeometry}~\cite{trinh2024alphageometry}, published in \emph{Nature} in January~2024. It is a neuro-symbolic system in which a symbolic deduction engine (Deductive Database Arithmetic Reasoning, DDAR) is guided by a neural language model trained from scratch on 100 million synthesized theorems and proofs. On \textsc{IMO-AG-30}, a benchmark of thirty IMO geometry problems from 2000--2022, \textsc{AlphaGeometry} solved 25, approaching the 25.9 average of human gold medalists, and substantially exceeding Wu's method (10 problems). \textsc{AlphaGeometry2}~\cite{chervonyi2025alphageometry2}, published in early~2025, extended the domain language to handle movements of objects, locus-type theorems, and non-constructive problems, lifting coverage of IMO~2000--2024 geometry problems from 66\% to 88\%, and solving 42 of 50 (exceeding the average gold medalist). \textsc{AlphaGeometry2} was one component of the combined system that achieved silver-medal standard at IMO~2024 alongside \textsc{AlphaProof}~\cite{hubert2025alphaproof}; the entire pipeline relied on human-assisted translation of problems into a domain-specific language.

\subsection{Vision--Language Models for Math}
A parallel trajectory emerged from the vision--language community. \textsc{MathVista}~\cite{lu2024mathvista} introduced a 6{,}141-example benchmark drawn from 31 source datasets, spanning figure QA, geometry problem solving, MWPs with visuals, textbook QA, and visual QA, and covering algebraic, arithmetic, geometric, logical, numeric, scientific, and statistical reasoning. On \textsc{MathVista} testmini, GPT-4V reached $49.9\%$, significantly above Bard (\mbox{$34.8\%$}) but $10.4$ points below human average.

\textsc{MathVerse}~\cite{zhang2024mathverse} specialized further into plane geometry, solid geometry, and functions (2{,}612 problems across six variant versions totalling 15K test samples) and introduced a step-wise CoT evaluation protocol. A key finding was that several MLLMs achieved \emph{higher} accuracy when visual input was removed, indicating that they relied primarily on textual features. \textsc{MATH-Vision}~\cite{wang2024mathvision} collected 3{,}040 problems from real math competitions across 16 disciplines and five difficulty levels. \textsc{MV-MATH} extends these benchmarks to multi-visual contexts, with leading models (Claude 3.5 Sonnet) reaching only $33.9\%$ compared to $76.5\%$ for humans.

On the model side, \textsc{G-LLaVA}~\cite{gao2023gllava} bootstrapped a geometry corpus (Geo170K) for vision-language fine-tuning, achieving strong performance on GPS tasks. \textsc{Math-LLaVA} introduces the MathV360K dataset; \textsc{MAVIS}~\cite{zhang2024mavis} further optimizes math-specific visual encoding and provides auto-generated CoT rationales. Reasoning-enabled multimodal models such as \textsc{o3} and Gemini~Deep~Think further integrate visual reasoning with image manipulation tools, retaining the raw image throughout the reasoning trace and zooming or rotating as needed.

The related-work map in \textsc{MINT-CoT}~\cite{chen2025mintcot} clarifies a newer split inside multimodal mathematical reasoning. One branch adapts text-only visual reasoning methods, such as R1-V-style long rationales, to images; another explicitly inserts visual material into the rationale through crops, highlighted regions, or external sketching tools, as in Visual CoT, Chain-of-Spot, Chain-of-Image, Visual SKETCHPAD, and ICoT. \textsc{MINT-CoT} argues that these mechanisms are often too coarse for diagrams because mathematical evidence may be a line segment, an angle, a label, or a non-rectangular configuration. Its central device is an \emph{Interleave Token}: before each reasoning step, the model selects fine-grained visual tokens from the original image and reasons over those tokens together with text. The accompanying 54K-example dataset is built from Mulberry-260K by pairing reasoning steps with grid-indexed visual regions using OCR and GPT-4o-assisted keyword alignment, and the resulting 7B model is trained through text-only CoT SFT, interleaved-CoT SFT, and interleaved-CoT RL. On the reported math-focused evaluations, it reaches $73.70$ on \textsc{MathVista-Math}, $64.72$ on \textsc{GeoQA}, and $69.60$ on \textsc{MMStar-Math}; the ablation shows that text-only CoT SFT supplies much of the initial lift, while token-level visual interleaving and RL add further gains. The larger lesson is conceptual: visual math systems are beginning to expose not only whether a model uses a diagram, but which parts of the diagram each reasoning step depends on.

An especially recent trend imports test-time scaling into geometry. \textsc{MARS-GPS}~\cite{siddique2026marsgps}, a 2026 preprint, generates multiple parallel reasoning rollouts, checks numerical subclaims with Python, ranks candidates using token-level entropy, and aggregates them through voting and self-verification. It reports $88.8\%$ accuracy on \textsc{Geometry3K} with eight rollouts and $77.48\%$ on \textsc{PGPS9K}. The result should be read cautiously until peer review, but it is conceptually important: geometric problem solving is beginning to resemble the broader reasoning-model paradigm in which a strong generator is valuable mainly when coupled to systematic search and verification.

\section{Formal Mathematical Reasoning}\label{sec:formal}
Systems discussed in this section correspond to the ``Formal Provers'' band in Figure~\ref{fig:timeline}.
A fundamental limitation of the informal reasoning systems discussed so far is that their outputs, however fluent, carry no guarantee of correctness. Sections~\ref{sec:llm} and~\ref{sec:multimodal} documented a recurring pattern: each advance in generation quality, longer chains of thought, tool-augmented programs, multi-agent debate, vision-language interleaving, solved progressively harder problems while simultaneously revealing new failure modes that no amount of sampling or voting could reliably eliminate. Arithmetic hallucinations survive self-consistency; logically invalid steps survive PRM scoring when the PRM itself lacks domain coverage; and multimodal models sometimes achieve higher accuracy when the diagram is \emph{removed}, indicating that the visual modality introduces as many failure modes as it resolves. These observations converge on a single insight: for mathematical claims that must be trusted, theorem statements, verified conjectures, competition proofs, an external \emph{mechanical} guarantee is needed, one that does not depend on the same learned distribution that generated the claim. This has motivated a parallel research agenda centered on formal proof assistants, most prominently Lean~4~\cite{demoura2021lean4,moura2015lean}, that mechanically verify each inference step. Working mathematicians increasingly regard this formal track as the essential complement to LLM-based informal reasoning~\cite{yang2024formal,li2024dl4tp,tao2024copilot,tao2026methods}.

\subsection{From Computer-Assisted to Machine-Assisted Proof}
Tao's 2024 account of machine-assisted proof usefully situates modern LLM provers within a much older tradition of mathematical computation~\cite{tao2024machineassisted}. Long before proof assistants, mathematicians used computation to generate data, test conjectures, carry out symbolic manipulation, and certify large finite searches. Contemporary examples include SAT/SMT solvers that emit proof certificates, rigorous numerical methods such as interval arithmetic, and computer algebra systems that perform symbolic reductions under human supervision. The point is not merely historical: machine-assisted proof is best understood as a spectrum from heuristic exploration to kernel-checked certification, with different trust assumptions at each level.

This genealogy also clarifies why formal proof assistants have become central. Earlier computer-assisted proofs, such as the four-color theorem and the Kepler conjecture, depended on substantial custom computation; later formalization projects reduced the trusted surface by rebuilding the argument inside a proof assistant. Tao highlights several scale markers: the four-color theorem was formalized in Coq; the Flyspeck verification of Kepler's conjecture required an eleven-year collaboration; Scholze's liquid tensor experiment took about eighteen months in Lean; and Tao's own polynomial Freiman--Ruzsa formalization converted a 33-page human proof into Lean in roughly three weeks with about twenty collaborators. These cases show that formalization is not just \textit{post hoc} checking. It uncovers hidden assumptions, creates reusable library material, and makes unusually large mathematical collaborations possible because subtasks can be specified and independently verified.

\subsection{Tactic Prediction and Neural Theorem Proving}
Early neural theorem provers treated proving as a tactic-prediction problem: given the current proof state, predict the next tactic to apply. \textsc{GPT-$f$}~\cite{polu2020generative} demonstrated that a GPT-style model trained on Metamath proofs could recover non-trivial proofs via expert iteration. \textsc{LeanDojo}~\cite{yang2023leandojo} released a large-scale Lean environment with premise retrieval via \textsc{ReProver}, enabling retrieval-augmented tactic prediction over \texttt{mathlib}. \textsc{Lean Copilot}~\cite{song2024leancopilot} integrates LLM-based tactic suggestion directly into the Lean~4 editor, operating as an inline co-pilot for human mathematicians.

The literature reviewed around \textsc{AlphaProof} suggests a useful subdivision of modern provers. Search-first systems such as \textsc{GPT-$f$} and \textsc{HyperTree Proof Search}~\cite{polu2020generative,lample2022hypertree} pair a proof-state model with tree or hyper-tree search, often learning from failed proof attempts through expert iteration or online training. Retrieval-centered systems such as \textsc{LeanDojo}/\textsc{ReProver} make premise selection an explicit part of tactic prediction~\cite{yang2023leandojo}. Whole-proof systems such as \textsc{Baldur}~\cite{first2023baldur} instead ask a model to emit a full Isabelle/HOL proof script and then use proof-assistant errors for repair. The newest high-performing systems blur these categories: \textsc{DSP} uses informal sketches to guide formal proof search~\cite{jiang2023draft}, \textsc{Lean-STaR} and \textsc{Kimina-Prover} interleave natural-language reasoning with Lean proof generation~\cite{lin2025leanstar,wang2025kimina}, and \textsc{DeepSeek-Prover-V2} and \textsc{AlphaProof} combine decomposition, verification feedback, and reinforcement learning at scale~\cite{ren2025deepseekproverv2,hubert2025alphaproof}.

A key efficiency innovation is \textsc{APOLLO} (Automated Proof repair via LLM and Lean cOllaboration)~\cite{ospanov2026apollo}, which treats the Lean~4 compiler not as a binary pass/fail oracle but as a \emph{diagnostic} tool. When an LLM-generated proof attempt fails, \textsc{APOLLO}'s agentic pipeline analyzes the compiler's error messages to fix syntax issues, isolates failing sub-lemmas, and recursively invokes the LLM on each remaining subgoal with a minimal sampling budget. The repaired sub-proofs are recombined and re-verified, and the process repeats for multiple attempts. This compiler-guided repair loop reduces the sampling budget by orders of magnitude compared to brute-force generation: for Goedel-Prover-SFT, \textsc{APOLLO} raises MiniF2F accuracy to $65.6\%$ while reducing the required samples from ${\sim}$25{,}600 to a few hundred, and it enables general-purpose models such as o3-mini to jump from single-digit accuracy ($3$--$7\%$) to over $40\%$ on formal proofs. Among models with 8B parameters or fewer, it establishes a state-of-the-art MiniF2F accuracy of $84.9\%$. The broader lesson is that structured repair, using the proof assistant's feedback to guide targeted fixes rather than regenerating entire proofs, is a scalable paradigm that complements both expert iteration and whole-proof generation.

A complementary line of work asks whether
LLMs can prove theorems \emph{without} formal proof assistants,
operating entirely in natural-language \LaTeX.
\textsc{DeepTheorem}~\cite{zhang2025deeptheorem} introduces a
121K-example dataset of IMO-level informal theorems with
step-by-step proofs generated by o3-mini, annotated for
correctness, difficulty (levels~5--10), and topic domain, and
rigorously decontaminated against 18 standard benchmarks.
Its key methodological contribution is an \emph{RL-Zero}
training protocol for informal proving: each theorem is
expanded into logically entailing and contradictory variants,
enabling a binary reward signal (proved \textit{vs.}\ disproved) that
is verifiable without a proof assistant.  A~7B model trained
with GRPO on these variants outperforms substantially larger
systems, including DeepSeek-R1-Distill-70B, QwQ-32B, and
Qwen2.5-Math-72B-Instruct, on the FIMO, HMMT, and
PutnamBench informal-proving benchmarks, reaching an average
outcome accuracy of $47.2\%$ versus $21.5\%$ for
R1-Distill-70B.  The accompanying process evaluation
framework scores generated proofs on logical validity
($40\%$), completeness ($30\%$), correctness ($20\%$), and
clarity ($10\%$), providing a richer signal than binary
outcome accuracy alone.
 
The DeepTheorem result is important for three reasons.
First, it demonstrates that RL-Zero can be extended from
closed-form QA to open-ended proof generation by
constructing verifiable theorem variants, addressing the
reward-design bottleneck that has limited RLVR to
answer-graded tasks.  Second, its process evaluation
framework offers a concrete operationalization of the
``reasoning quality \textit{vs.}\ answer accuracy'' distinction that
recent surveys have called for but not
resolved~\cite{wang2025survey,liu2025mathematicallm}.
Third, by showing that a 7B model can match or exceed
70B-class reasoning models on informal proving, it
reinforces the finding from the broader reasoning-model
literature that data quality and training protocol often
matter more than raw parameter count.

\subsection{Autoformalization}
Autoformalization, translating natural-language mathematics into formal statements, was shown to be feasible via few-shot prompting of Codex by~\cite{wu2022autoformalization}. \textsc{DSP} (Draft, Sketch, Prove)~\cite{jiang2023draft} interleaves informal and formal reasoning, having the model first produce an informal proof sketch, autoformalize it into Isabelle, and discharge the remaining subgoals via a hammer. \textsc{ProofNet}~\cite{azerbayev2023proofnet} provided 371 paired informal/formal statements from undergraduate textbooks. The \textsc{Lean Workbook}~\cite{ying2024leanworkbook} formalized hundreds of thousands of natural-language math competition problems into Lean 4 statements. Type-checking-based filtering~\cite{poiroux2024typechecking} has become standard: autoformalized statements that fail to type-check are rejected, and iterative refinement substantially improves yield.

A more recent line of work uses the Lean~4 compiler not just as a binary type-checker but as a source of \emph{process-level supervision} for autoformalization training. The \textsc{FORML4} dataset and \textsc{PDA} (Process-Driven Autoformalization) framework~\cite{lu2024process} pair natural-language theorems and proofs with their Lean~4 formalizations, then train a Process-Supervised Verifier (PSV) on the compiler's fine-grained error messages, including error locations, type mismatches, and unresolved goals. The PSV provides step-level feedback to the autoformalization model during fine-tuning, creating an iterative loop: the autoformalizer generates a candidate, the compiler diagnoses specific failures, and the PSV guides the next training round toward higher compiler acceptance rates with less filtered data. This process-driven approach parallels the PRM/ORM distinction in informal reasoning (\S\ref{sec:verification}): outcome-level feedback (``does it type-check?'') is cheap but coarse, while process-level feedback (``which step fails and why?'') enables more targeted learning. Together with \textsc{APOLLO}'s compiler-guided repair, PDA illustrates a broader trend in which the proof assistant's feedback channel, long treated as a binary gate, is being elevated to a rich supervision signal.

\subsection{Large-Scale LLM-Based Provers}
The \textsc{DeepSeek-Prover} series illustrates the rapid maturation of this area. \textsc{DeepSeek-Prover}~\cite{xin2024deepseekprover} synthesized 8 million Lean~4 statement--proof pairs by (i) autoformalizing 870K natural-language competition problems using DeepSeekMath-7B; (ii) filtering low-quality statements with chain-of-thought scoring and negation-based hypothesis rejection; and (iii) iteratively expert-iterating the model on successful proofs. This established that expert iteration with high-quality synthetic data could substantially improve theorem-proving ability in Lean.

\textsc{DeepSeek-Prover-V1.5}~\cite{xin2025deepseekproverv15} added RLHF using proof-assistant feedback and integrated Monte Carlo Tree Search to guide tactic selection. \textsc{DeepSeek-Prover-V2}~\cite{ren2025deepseekproverv2}, released in April~2025, took a recursive decomposition approach: DeepSeek-V3 produces a natural-language proof sketch together with a Lean template containing \texttt{sorry} placeholders, and a 7B prover recursively solves each subgoal; the resulting verified proofs serve as training data for a 671B RL-fine-tuned CoT prover. On \textsc{MiniF2F-test}~\cite{zheng2022minif2f}, \textsc{DeepSeek-Prover-V2-671B} reaches 88.9\% pass rate at Pass@8192, solves 49 of 658 PutnamBench problems, and correctly proves 6 of 15 AIME problems in ProverBench. Complementary systems include \textsc{Lean-STaR}~\cite{lin2025leanstar}, which interleaves natural-language thoughts with tactic choices trained via STaR~\cite{zelikman2022star}, and community projects such as Goedel-Prover, Kimina-Prover, and Harmonic's \textsc{Aristotle} which have pushed the frontier further.

\subsection{AlphaProof and IMO 2024}
The landmark result of the formal track in 2024 was \textsc{AlphaProof}~\cite{hubert2025alphaproof}, a reinforcement-learning system trained in a Lean~4 environment using an AlphaZero-inspired algorithm. Its proof network is a 3B-parameter encoder--decoder transformer that predicts both a tactic policy and a value estimate for the current proof state; search is organized as an AND--OR proof tree so that decomposed subgoals must all be solved. Training begins with broad code/math pretraining and Mathlib state--tactic supervision, but the central phase is RL over roughly 80~million Lean statements produced from about one million natural-language problems by an autoformalization system. For the hardest targets, \textsc{AlphaProof} uses \emph{test-time reinforcement learning} (TTRL): a Gemini-based variant generator creates a large local curriculum of related Lean problems, and a specialist prover is trained around the target instance before final search.

During IMO~2024, with the five non-geometry problems manually translated into Lean by experts, \textsc{AlphaProof} solved P1, P2, and P6; P6 was the hardest problem of the competition and was fully solved by only five of 609 human contestants. \textsc{AlphaGeometry~2} solved the geometry problem P4, so the combined system scored 28/42 points, equivalent to a silver medalist. This result is methodologically important but should be interpreted carefully: the competition pipeline used expert formalization, answer-guessing assistance for ``find all'' problems, and 2--3 days of TTRL for the solved non-geometry problems. At IMO~2025, an advanced version of Gemini Deep Think~\cite{deepmind2025geminideepthink}, operating end-to-end in natural language within the 4.5-hour time limit, achieved the gold-medal threshold (35/42, five problems solved perfectly), closing the formal-vs-informal gap that had characterized the 2024 result.

\begin{table*}[t]
\centering
\scriptsize
\setlength{\tabcolsep}{3pt}
\renewcommand{\arraystretch}{1.10}
\resizebox{\textwidth}{!}{%
\begin{tabular}{p{0.20\textwidth}lllccc}
\tabletop
\thead{System} & \thead{Params} & \thead{Approach} & \thead{Pass@$N$} & \thead{MiniF2F-test} & \thead{PutnamBench} & \thead{Formal-IMO} \\ \tablemid
\tgroup{7}{Search-first tactic prediction}
GPT-$f$~\cite{polu2020generative}                          & 774M  & Expert iteration over Metamath              & @8         & 36.6\% & --      & -- \\
HyperTree Proof Search~\cite{lample2022hypertree}         & 600M  & Online hyper-tree proof search               & @64        & 41.0\% & --      & -- \\
\tgroup{7}{Retrieval-augmented proving}
LeanDojo ReProver~\cite{yang2023leandojo}                 & 300M  & Retrieval-augmented tactic prediction        & --         & 48.4\% & --      & -- \\
InternLM2-Math-Plus~\cite{ying2024internlm}               & 7B   & Open math LLM, verifiable reasoning          & @32        & 43.4\% & --      & -- \\
\tgroup{7}{Thought-interleaved and whole-proof generation}
Lean-STaR~\cite{lin2025leanstar}                          & 7B   & NL thoughts interleaved with tactics         & @64        & 46.3\% & --      & -- \\
Goedel-Prover V1~\cite{lin2025goedelprover}               & 7B   & Iterative SFT on autoformalized proofs       & @32        & 57.6\% & 7/658   & -- \\
\tgroup{7}{Expert iteration + RL provers}
DeepSeek-Prover V1~\cite{xin2024deepseekprover}           & 7B   & Synthetic Lean data + expert iteration       & @65536     & 50.0\% & --      & -- \\
DeepSeek-Prover V1.5~\cite{xin2025deepseekproverv15}      & 7B   & RLHF + MCTS over proof-assistant feedback    & RMaxTS     & 63.5\% & --      & -- \\
Kimina-Prover Preview~\cite{wang2025kimina}               & 72B  & Large-scale RL + formal reasoning traces     & @32        & 80.7\% & 1.6\%\textsuperscript{$\star$} & -- \\
DeepSeek-Prover V2 (7B)~\cite{ren2025deepseekproverv2}  & 7B   & Recursive subgoal decomposition              & @8192      & 82.0\% & 23/658  & -- \\
Goedel-Prover V2~\cite{lin2025goedelproverv2}      & 8B   & Scaffolded synthesis + self-correction        & @32        & 84.6\% & --      & -- \\
Goedel-Prover V2~\cite{lin2025goedelproverv2} & 32B  & Scaffolded synthesis + self-correction        & @32        & 90.4\%\textsuperscript{$\ddagger$} & 86/658\textsuperscript{$\ddagger$}  & -- \\
DeepSeek-Prover V2~\cite{ren2025deepseekproverv2}  & 671B & Large CoT prover from verified subproofs     & @8192      & 88.9\% & 49/658  & -- \\
\tgroup{7}{Compiler-guided repair}
\textsc{APOLLO}+Goedel-Prover-SFT~\cite{ospanov2026apollo} & -- & Lean compiler diagnostics for proof repair & few hundred & 65.6\% & -- & -- \\
\textsc{APOLLO} (best $\leq$8B)~\cite{ospanov2026apollo} & $\leq$8B & Recursive subgoal repair with compiler feedback & -- & 84.9\% & -- & -- \\
\tgroup{7}{Neuro-symbolic RL + formal search}
AlphaProof + TTRL\textsuperscript{$\dagger$}~\cite{hubert2025alphaproof} & 3B & AlphaZero-style Lean RL, AND--OR proof tree & TTRL & 99.6\%\textsuperscript{$\dagger$} & 56.1\%\textsuperscript{$\dagger$} & 58.3\%\textsuperscript{$\dagger$} \\
\tablebottom
\end{tabular}%
}
\caption{Performance of representative neural theorem provers on formal benchmarks, grouped by approach. \textbf{Pass@$N$} reports the search budget: ``@$k$'' = $k$ samples, RMaxTS = reward-MaxTS, TTRL = test-time RL, and ``--'' = unreported. PutnamBench entries are full-benchmark counts unless $^\star$ marks PutnamBench-test percentage; $^\ddagger$ denotes two rounds of compiler-guided self-correction. $^\dagger$AlphaProof uses expert Lean formalizations and large TTRL budgets, so it is not directly comparable with fully automatic systems. The table shows rapid open-prover catch-up alongside a still distinct expert-assisted TTRL regime.}
\label{tab:provers}
\end{table*}

\subsection{Ecosystem, Libraries, and Human Workflow}
Formal mathematical reasoning is as much an infrastructure problem as a modeling problem. Proof assistants are useful because they reduce correctness to kernel checking, but the cost of reaching the kernel is paid through library coverage, notation alignment, premise retrieval, and proof-state interaction. Lean~4 has become especially influential because \texttt{mathlib} supplies a rapidly expanding shared corpus and because the surrounding tool ecosystem increasingly resembles modern software development: editor integration, continuous integration, search tools, package management, and proof suggestions.

Tao frames the central practical barrier as the \emph{de Bruijn factor}: the ratio between the effort required to write a correct formal proof and the effort required to write a correct informal proof~\cite{tao2024machineassisted}. His estimate of this ratio is still well above one, but falling as proof assistants, libraries, tactics, SMT solvers, and LLM-based copilots become better integrated. This is an important lens for AI-for-mathematics: the decisive threshold is not whether a model can occasionally solve an isolated Lean benchmark, but whether the combined human--AI--proof-assistant workflow makes formalization cheaper than informal proof maintenance, error checking, and referee labor.

This matters for survey-level comparison because a theorem-proving benchmark does not only measure the model. It also measures whether the relevant definitions already exist in the library, whether the theorem is stated with the same conventions as \texttt{mathlib}, whether useful lemmas can be retrieved, and whether the model receives proof-state feedback. A problem that is easy for a human mathematician may be difficult for Lean if the background theory has not been formalized, while a problem that looks advanced may become easy once an existing library theorem exactly matches it. For this reason, future evaluations should report library version, allowed imports, premise-retrieval access, search budget, and whether any human formalization assistance was used.

An emerging direction addresses the library-coverage bottleneck directly: rather than waiting for human contributors to formalize missing intermediate results, LLMs can proactively \emph{generate} them. \textsc{MathlibLemma}~\cite{liu2026mathliblemma} employs a multi-agent architecture, Discovery, Judge, and Formalizer agents, to identify ``folklore lemmas'': mathematical facts widely known to practitioners but absent from \texttt{mathlib}. The system uses \texttt{mathlib} entries as seeds, proposes candidate lemmas, filters them for mathematical soundness, and generates type-checked Lean~4 code. The accompanying benchmark comprises 4{,}028 type-checked Lean~4 statements, and an LLM-assisted human audit finds $78\%$ mathematically sound. A subset of the generated lemmas has been merged into the official \texttt{mathlib} repository, demonstrating that LLMs can transition from passive consumers to active contributors of formal library material. This is a concrete step toward reducing the de Bruijn factor: if the ``connective tissue'' of routine lemmas can be automatically generated and verified, human mathematicians can focus on the creative steps.

\section{Mathematical Discovery and Open Problems}\label{sec:discovery}
Systems discussed in this section correspond to the ``Discovery Systems'' band in Figure~\ref{fig:timeline}.
A third and arguably most consequential axis concerns the use of LLMs to produce \emph{new} mathematical knowledge, improved bounds, explicit constructions, counterexamples to conjectures, and, in the limit, novel proofs of open problems.

\subsection{Program Search: \textsc{FunSearch} and \textsc{AlphaEvolve}}
\textsc{FunSearch}~\cite{romera2024funsearch} introduced the paradigm of evolutionary program search guided by an LLM, in which a pre-trained code LLM proposes variations of a Python function and an automated evaluator selects the highest-scoring programs. In \emph{Nature}~2024, the authors demonstrated its application to the cap-set problem in additive combinatorics (producing new lower bounds for cap-set sizes in several dimensions) and to online bin packing (discovering novel heuristics).

\textsc{AlphaEvolve}~\cite{novikov2025alphaevolve}, released in 2025, generalized \textsc{FunSearch} by evolving entire programs (hundreds of lines) rather than single functions, combining Gemini Flash (breadth) with Gemini Pro (depth), and using richer natural-language feedback. Its most publicized result was the discovery of an algorithm for multiplying two $4\times 4$ complex-valued matrices using 48 scalar multiplications, the first improvement over Strassen's 1969 algorithm for complex matrices. The larger arXiv study by Georgiev, G\'omez-Serrano, Tao, and Wagner~\cite{georgiev2025math} evaluates this paradigm on 67 problems spanning analysis, combinatorics, geometry, and number theory. \textsc{AlphaEvolve} rediscovered best-known solutions in most cases and \emph{improved} them in about $20\%$, including the finite-field Kakeya problem, the kissing-number problem in 11 dimensions (593-sphere construction), and the Nikodym problem. In some cases it also extrapolated finite computational evidence into formulas valid for all input sizes, making the system closer to a conjecture-and-construction engine than a mere optimizer. The study further combines \textsc{AlphaEvolve} with Gemini Deep Think for proof generation and \textsc{AlphaProof} for formal verification in Lean, illustrating a broader pipeline from search over constructions to machine-assisted proof.

\subsection{Erd\H{o}s Problems and the AI-Assisted Attack Surface}
Paul Erd\H{o}s (1913--1996) left behind over a thousand conjectures; the erdosproblems.com database~\cite{bloom2024erdos} currently catalogs~1{,}133 of them, of which roughly 680 remained open as of early~2026. Beginning in late 2025 and accelerating through early~2026, a succession of these problems has been solved with non-trivial AI contribution~\cite{alexeev2025erdos}.

An early claim from OpenAI in October~2025 that GPT-5 had autonomously solved ten Erd\H{o}s problems was quickly rebutted by T.~Bloom, who pointed out that these were in fact literature lookups: the system had located existing papers that the database curator was unaware of. The subsequent bona fide solves reflect a much more rigorous division of labor. In December~2025, Erd\H{o}s problem~\#1026 was solved: the \textsc{Aristotle} system~\cite{achim2025aristotle} (operated by B.~Alexeev) generated the proof sketch and autonomously formalized it in Lean~4 with minimal human steering. Between January and March~2026, problems~\#728, \#729, and \#397 were solved using a collaborative pipeline: GPT-5.2 Pro generated the informal proof strategies, the \textsc{Aristotle} system was used to formalize the resulting arguments into Lean, and human experts orchestrated the translation and lemma decomposition. Across all four cases, T.~Tao independently verified the proofs and documented them as legitimate examples of AI-assisted mathematics rather than literature search. As of April~2026, 22 of 279 proved Erd\H{o}s problems carry formal Lean verification. Tao himself~\cite{tao2026primetime} stresses that these solves involve ``lowest hanging fruit'', problems amenable to standard techniques once the right retrieval and connection is made, but notes the qualitative shift: ``compression matters'', and problems that might have taken hours or days of expert effort now fall in minutes.

A separate episode of genuine human--AI collaboration is documented in Tao's March~2026 paper \emph{Local Bernstein theory, and lower bounds for Lebesgue constants}, where one inequality was first suggested numerically by \textsc{AlphaEvolve}, then proved by ChatGPT via a duality argument drawing on approximation theory, which Tao verified and formalized in Lean, with the resulting 1{,}125-line Lean proof accepted into his repository. The episode illustrates the emerging modus operandi: AI as a cross-domain search-and-connection engine, coupled with formal verification to guarantee correctness.

\subsection{Discovery as a Workflow}
What is striking about the 2025--2026 discovery work is that it consistently instantiates a \emph{workflow}, not a single model, combining four specialized capabilities: (i)~neural search or proposal, (ii)~informal proof drafting, (iii)~autoformalization into Lean, and (iv)~formal verification. Tao's machine-assisted-proof framing explains why this architecture is plausible: LLMs are strong at generating candidate directions and code, whereas proof assistants, symbolic solvers, and rigorous computation are strong at rejecting invalid outputs~\cite{tao2024machineassisted}.

This workflow paradigm is most visible in evolutionary program search. \textsc{FunSearch}~\cite{romera2024funsearch} pioneered the approach by using an LLM to mutate a single target function (\textit{e.g.}, a priority function for bin packing), evaluating the new function's fitness, and feeding successful mutations back into a prompt pool. However, this single-function restriction limits the complexity of the algorithms it can discover. \textsc{AlphaEvolve}~\cite{novikov2025alphaevolve} extends this by implementing \emph{whole-program evolution}: the LLM mutates not just one function, but an entire multi-file Python repository representing an arbitrary computational workflow, allowing it to invent new data structures, helper functions, and optimization loops simultaneously. This architectural shift enables \textsc{AlphaEvolve} to attack a broader class of problems, as summarized in Table~\ref{tab:discovery_systems}.
\begin{table}[t]
\centering
\scriptsize
\renewcommand{\arraystretch}{1.2}
\setlength{\tabcolsep}{4pt}
\begin{tabularx}{\columnwidth}{@{}l >{\raggedright\arraybackslash}X >{\raggedright\arraybackslash}X >{\raggedright\arraybackslash}X@{}}
\tabletop
\thead{System} & \thead{Target problem class} & \thead{Primary verifier type} & \thead{Novelty guarantee} \\ \tablemid
\textsc{FunSearch} & Combinatorial optimization (\textit{e.g.}, cap sets, bin packing) & Python execution (fitness evaluation) & Exceeds best known numerical bounds \\
\textsc{AlphaEvolve} & Open-ended algorithmic discovery & Distributed sandbox execution & Exceeds known bounds; generalizes to full programs \\
Erd\H{o}s Workflow & Open mathematical conjectures & Lean 4 kernel (formal verification) & Resolves unproved conjecture listed in database \\
\tablebottom
\end{tabularx}
\caption{Comparison of recent mathematical discovery systems across their target problem classes, verification mechanisms, and the standards used to guarantee that a discovery is genuinely novel.}
\label{tab:discovery_systems}
\end{table}

The AlphaEvolve-scale study makes the same point operationally: evolutionary program search proposes constructions, Deep Think supplies mathematical argumentation, and AlphaProof or Lean supplies the final correctness filter~\cite{georgiev2025math}. The current bottleneck is no longer any single capability in isolation but rather the interfaces between them, a theme we return to in Section~\ref{sec:future}.

A fourth, distinct mode emerged with the \textbf{AI co-mathematician}~\cite{zheng2026comathematician} in May~2026: an asynchronous, stateful \emph{workbench} in which a mathematician orchestrates a hierarchy of agents (project coordinator, workstream coordinators, specialised sub-agents for literature retrieval, code execution, and proof drafting via Gemini Deep Think, and reviewer agents) under continuous human steering rather than autonomous end-to-end execution. The system reports 48\% on \textsc{FrontierMath} Tier 4, the hardest, research-grade tier~\cite{glazer2024frontiermath}, and is credited with non-trivial contributions to three open problems: Kourovka 21.10 on just-finite group presentations (Lackenby), two log-concavity conjectures on Stirling coefficients (B\'erczi), and a perturbation lemma in Hamiltonian systems (Rezchikov). The co-mathematician thus complements the program-search (\textsc{FunSearch}, \textsc{AlphaEvolve}) and autoformalisation-pipeline (Erd\H{o}s workflow, \textsc{Aristotle}) modes already discussed: rather than chasing fixed benchmarks, it targets the exploratory, iterative reality of mathematical research itself. A parallel ``autonomous mathematics research'' agent, \textsc{Aletheia}~\cite{feng2026aletheia}, was released in February~2026 with similar ambitions but a different orchestration strategy; the convergence of two industrial labs on this workbench paradigm within three months suggests it is becoming the default deployment mode for research-assistance mathematics.

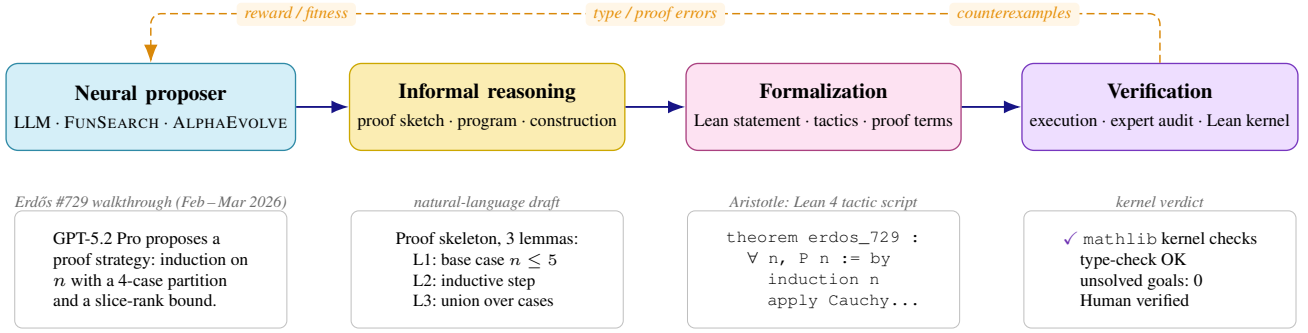
\begin{figure*}[t]
\centering
\begin{tikzpicture}[
    stage/.style={rectangle, rounded corners=1.6mm, line width=.5pt,
                  minimum width=3.6cm, minimum height=1.15cm,
                  align=center, font=\fontsize{8}{9.5}\selectfont,
                  inner sep=3pt},
    stageProp/.style={stage, draw=VivCyan!75!black, fill=VivCyan!22},
    stageInf/.style={stage, draw=VivYellow!85!black, fill=VivYellow!30},
    stageForm/.style={stage, draw=VivPink!70!black, fill=VivPink!18},
    stageVer/.style={stage, draw=VivPurple!75!black, fill=VivPurple!18},
    inst/.style={rectangle, rounded corners=1mm, line width=.35pt,
                 minimum width=3.6cm, minimum height=1.55cm,
                 align=left, font=\fontsize{6.6}{7.8}\selectfont,
                 fill=white, draw=black!25,
                 inner xsep=4pt, inner ysep=3pt},
    instHead/.style={font=\fontsize{6.0}{6.8}\selectfont\itshape,
                     text=black!58, anchor=south, inner sep=0pt},
    arr/.style={-{Latex[length=2.3mm]}, line width=.7pt, draw=Indigo!70!black},
    bind/.style={draw=black!28, line width=.3pt, dash pattern=on .8pt off .8pt},
    fbk/.style={-{Latex[length=2.4mm]}, line width=.55pt,
                draw=VivOrange!85!black, dash pattern=on 3pt off 1.6pt},
    fblabel/.style={font=\fontsize{6.4}{7.2}\selectfont\itshape,
                    text=VivOrange!90!black, fill=VivOrange!10,
                    rounded corners=.6mm,
                    inner xsep=3pt, inner ysep=1.5pt}
]
\def\xA{0.0}     
\def\xB{4.45}    
\def\xC{8.90}    
\def\xD{13.35}   
\def\yS{4.0}     
\def\yI{1.85}    

\node[stageProp] (s1) at (\xA,\yS)
   {\textbf{Neural proposer}\\[1pt]{\fontsize{6.6}{7.4}\selectfont LLM~\textperiodcentered~\textsc{FunSearch}~\textperiodcentered~\textsc{AlphaEvolve}}};
\node[stageInf]  (s2) at (\xB,\yS)
   {\textbf{Informal reasoning}\\[1pt]{\fontsize{6.6}{7.4}\selectfont proof sketch~\textperiodcentered~program~\textperiodcentered~construction}};
\node[stageForm] (s3) at (\xC,\yS)
   {\textbf{Formalization}\\[1pt]{\fontsize{6.6}{7.4}\selectfont Lean statement~\textperiodcentered~tactics~\textperiodcentered~proof terms}};
\node[stageVer]  (s4) at (\xD,\yS)
   {\textbf{Verification}\\[1pt]{\fontsize{6.6}{7.4}\selectfont execution~\textperiodcentered~expert audit~\textperiodcentered~Lean kernel}};

\draw[arr] (s1) -- (s2);
\draw[arr] (s2) -- (s3);
\draw[arr] (s3) -- (s4);

\node[inst] (i1) at (\xA,\yI)
   {GPT-5.2 Pro proposes a\\proof strategy: induction on\\$n$ with a 4-case partition\\and a slice-rank bound.};
\node[instHead] at ([yshift=0pt]i1.north) {Erd\H{o}s \#729 walkthrough (Feb\,--\,Mar 2026)};

\node[inst] (i2) at (\xB,\yI)
   {Proof skeleton, 3 lemmas:\\
   \quad{\fontsize{6.4}{7.2}\selectfont L1: base case $n\le 5$}\\
   \quad{\fontsize{6.4}{7.2}\selectfont L2: inductive step}\\
   \quad{\fontsize{6.4}{7.2}\selectfont L3: union over cases}};
\node[instHead] at ([yshift=0pt]i2.north) {natural-language draft};

\node[inst] (i3) at (\xC,\yI)
   {{\fontsize{6.4}{7.4}\selectfont\ttfamily theorem erdos\_729 :}\\
   {\fontsize{6.4}{7.4}\selectfont\ttfamily \ \ $\forall$ n, P n := by}\\
   {\fontsize{6.4}{7.4}\selectfont\ttfamily \ \ \ \ induction n}\\
   {\fontsize{6.4}{7.4}\selectfont\ttfamily \ \ \ \ apply Cauchy\dots}};
\node[instHead] at ([yshift=0pt]i3.north) {Aristotle: Lean 4 tactic script};

\node[inst] (i4) at (\xD,\yI)
   {\textcolor{VivPurple!70!black}{$\checkmark$} \texttt{mathlib} kernel checks\\
   \quad type-check OK\\
   \quad unsolved goals: 0\\
   \quad Human verified};
\node[instHead] at ([yshift=1pt]i4.north) {kernel verdict};

\draw[fbk, rounded corners=4pt]
   (s4.north) -- ++(0,0.65) -| (s1.north);

\def\yFB{5.225}  
\node[fblabel] at ({(\xA+\xB)/2 - .3}, \yFB) {reward / fitness};
\node[fblabel] at ({(\xB+\xC)/2}, \yFB) {type / proof errors};
\node[fblabel] at ({(\xC+\xD)/2 + .3}, \yFB) {counterexamples};

\end{tikzpicture}
\caption{The verified-discovery pipeline. \textbf{Top row}: four abstract stages, neural proposer (cyan), informal reasoning (yellow), formalization (pink), and verification (purple), each with its representative artifacts. \textbf{Bottom row}: a concrete instantiation by the GPT-5.2 Pro $+$ \textsc{Aristotle} team that produced a Lean-checked proof of Erd\H{o}s problem \#729 in early 2026, with one panel per stage showing the actual artifact it produced. The dashed orange feedback loop carries three signal types---counterexamples, type / proof errors, and reward / fitness---back to the proposer, closing the iterative search.}
\label{fig:verified_pipeline}
\end{figure*}

\section{Dataset Repository and Performance Analysis}\label{sec:datasets}
Understanding the arc of progress requires a panoramic view of the datasets on which progress is measured. We preserve the comprehensive catalog of MWP and geometry datasets from the earlier literature and extend it with the new benchmarks that have defined the 2023--2026 period.

\subsection{Training, Benchmark, and Augmentation Corpora}
Recent surveys emphasize that the dataset ecosystem should not be viewed only as a list of test benchmarks~\cite{liu2025mathematicallm,wang2025survey}. Mathematical datasets now play at least four distinct roles: they supply pretraining text, provide supervised reasoning traces, evaluate models, and create augmented feedback for verifiers or self-improvement. This distinction is important because benchmark saturation can coexist with genuine progress in training data, and conversely a larger training corpus does not guarantee robustness on fresh evaluations.

We return to the cross-cutting design principles implied by this functional view in Section~\ref{sec:synthesis}.

\begin{table*}[t]
\centering
\scriptsize
\setlength{\tabcolsep}{4pt}
\renewcommand{\arraystretch}{1.28}
\providecommand{\stripe}[1]{\textcolor{#1}{\rule[-.45ex]{1.6pt}{1.7ex}}\hspace{2pt}}
\begin{tabularx}{\textwidth}{@{}>{\raggedright\arraybackslash}p{0.155\textwidth}>{\raggedright\arraybackslash}X>{\raggedright\arraybackslash}p{0.20\textwidth}>{\raggedright\arraybackslash}p{0.22\textwidth}@{}}
\tabletop
\thead{Dataset role} & \thead{Representative resources (with scale)} & \thead{Typical artifact} & \thead{Primary function} \\ \tablemid
\tgroup{4}{Training-side resources}
\stripe{VivCyan!75!black}\textbf{Math pretraining}    & \textsc{AMPS} (23B tok), \textsc{ProofPile2} (55B tok), \textsc{OpenWebMath} (14.7B tok), \textsc{MathPile} (9.5B tok), \textsc{AutoMathText} (200B$+$ tok), \textsc{AlgebraicStack}                                                & Raw arXiv, textbooks, web math, code                       & Notation fluency, theorem familiarity; powers \textsc{Minerva}, \textsc{Llemma}, \textsc{DeepSeekMath} \\
\stripe{VivLime!60!black}\textbf{SFT / instruction}    & \textsc{MetaMathQA} (395K), \textsc{MathInstruct} (260K), \textsc{NuminaMath} (860K$+$), \textsc{OpenMathInstruct-2} (14M), \textsc{Lean Workbook} (57K, formal)                                                                    & CoT, PoT, tactic traces                                    & Solution format, tool use, competition patterns; formal subsets enable Lean SFT \\
\stripe{VivPurple!85!black}\textbf{Augmented / verifier} & \textsc{PRM800K} (800K step labels), \textsc{Math-Shepherd} (MC rollouts), \textsc{OmegaPRM} (1.5M$+$ labels), \textsc{MathQA-Python} (23.9K), \textsc{Lila} (134K), \textsc{NaturalProofs-Gen} (14.5K)                              & Step labels, programs, informal / formal alignments        & Train PRMs, program generators, autoformalizers, self-improvement loops \\
\tgroup{4}{Evaluation-side resources}
\stripe{VivOrange!75!black}\textbf{Evaluation}                  & \textsc{GSM8K} (8.8K), \textsc{MATH} (12.5K), \textsc{OlympiadBench} (8.5K), \textsc{MathVista} (6.1K), \textsc{MiniF2F} (488), \textsc{PutnamBench} (658), \textsc{FrontierMath} ($\sim$350), \textsc{MathArena} (149$+$, live)    & Answers, diagrams, formal statements, private checks       & Grade-school through research-level generalization; live benchmarks resist contamination \\
\stripe{VivLilac!80!black}\textbf{Multilingual}                  & \textsc{Math23K} (23K, ZH), \textsc{Ape210K} (210K, ZH), \textsc{MGSM} (2.5K, 10 langs), \textsc{HAWP} (2.3K, HI), \textsc{ArMATH} (6K, AR), \textsc{PatiGonit} (10K, BN), \textsc{BMWP} (8.7K, BN), \textsc{PolyMath} (9K, 18 langs), \textsc{M3Kang} (108K, 108 langs) & Native or parallel problems; equations, answers, diagrams & Language shift, morphology, script, low-resource robustness \\
\stripe{VivPink!75!black}\textbf{General expert suites}              & \textsc{MMLU} (15.9K, 57 subj.), \textsc{MMLU-Pro} (12K), \textsc{GPQA} (448, grad-level), \textsc{LiveBench} (monthly), \textsc{HLE} (2.5K, \emph{Nature} 2026), \textsc{SuperGPQA} (285 disciplines)                              & MCQ, exact match, live-scored tasks                        & Cross-domain quantitative transfer; calibration under contamination-resistant protocols \\
\tablebottom
\end{tabularx}
\caption{Functional taxonomy of mathematical datasets by lifecycle position. Training-side resources support pretraining, instruction tuning, and verifier construction; evaluation-side resources test held-out reasoning, multilingual robustness, and cross-domain transfer. Resource scales are shown in parentheses.}
\label{tab:data_roles}
\end{table*}

The practical lesson is that future papers should report not only benchmark scores but also the data path that produced them: whether the model was math-pretrained, whether CoT or program traces were used for SFT, whether synthetic data came from a stronger model, whether formal data were type-checked, and whether augmented data encode wrong steps as well as correct ones. These details determine whether a reported gain reflects better mathematical comprehension, better generation style, stronger verification, or simply closer overlap with the evaluation distribution.

Tables~\ref{tab:dataset_landscape} and~\ref{tab:model_landscape} condense the long catalog into two survey-level maps. They are intentionally organized by function rather than chronology: the most useful question for a reader is often not ``which benchmark is newest?'' but ``what failure mode or modeling choice does this resource expose?''

\begin{table*}
\centering
\tiny
\setlength{\tabcolsep}{2pt}
\renewcommand{\arraystretch}{1.06}
\providecommand{\spv}[2]{{\setlength{\fboxsep}{0.7pt}%
   \colorbox{#1}{\textcolor{white}{\fontsize{4.6}{5.4}\ttfamily\bfseries\,#2\,}}}}
\definecolor{ModalityCyan}{HTML}{58C6E8}
\definecolor{ModalityPink}{HTML}{F0A0B3}
\definecolor{LevelLight}{HTML}{6FA3F4}
\definecolor{LevelSoft}{HTML}{5B94F0}
\definecolor{LevelMid}{HTML}{4285F4}
\definecolor{LevelDeep}{HTML}{2F78E6}
\definecolor{LevelDark}{HTML}{1A66D8}
\providecommand{\mbadge}[2]{{\setlength{\fboxsep}{0.7pt}%
   \colorbox{#1}{\textcolor{white}{\fontsize{4.6}{5.4}\ttfamily\bfseries\,#2\,}}}}
\providecommand{\lbadge}[2]{{\setlength{\fboxsep}{0.7pt}%
   \colorbox{#1}{\textcolor{white}{\fontsize{4.6}{5.4}\ttfamily\bfseries\,#2\,}}}}
\providecommand{\lvlE}{\lbadge{LevelLight}{E}}
\providecommand{\lvlM}{\lbadge{LevelSoft}{M}}
\providecommand{\lvlH}{\lbadge{LevelMid}{H}}
\providecommand{\lvlU}{\lbadge{LevelMid}{U}}
\providecommand{\lvlC}{\lbadge{LevelDeep}{C}}
\providecommand{\lvlR}{\lbadge{LevelDark}{R}}
\providecommand{\lvlG}{\lbadge{LevelDark}{G}}
\providecommand{\modT}{\mbadge{ModalityCyan}{T}}
\providecommand{\modM}{\mbadge{ModalityPink}{M}}
\resizebox{\textwidth}{!}{%
\begin{tabular}{llrrrlllllp{2.8cm}}
\tabletop
\thead{Resource} & \thead{Year} & \thead{\#Train} & \thead{\#Test} & \thead{\#Total} & \thead{Language} & \thead{Task} & \thead{Solution} & \thead{Level} & \thead{Modality} & \thead{Diagnostic focus} \\ \tablemid
\tgroup{11}{Classical and large-scale MWP benchmarks}
\textsc{AI2}~\cite{hosseini2014learning}           & 2014 & --        & 395       & 395         & EN     & MWP   & Eq    & \lvlE    & \modT & Additive reasoning, verb categories \\
\textsc{SingleEQ}~\cite{koncel2015parsing}          & 2015 & --        & 508       & 508         & EN     & MWP   & Eq    & \lvlE    & \modT & Single-unknown algebraic parsing \\
\textsc{Alg514}~\cite{kushman2014learning}          & 2014 & --        & 514       & 514         & EN     & MWP   & Eq    & \lvlM    & \modT & Equation-template induction \\
\textsc{Dolphin18K}~\cite{huang2016well}            & 2016 & 14{,}768  & 3{,}692   & 18{,}460    & EN     & MWP   & Eq    & \lvlM    & \modT & Large-scale template diversity \\
\textsc{MAWPS}~\cite{koncel2016mawps}               & 2016 & --        & 3{,}320   & 3{,}320     & EN     & MWP   & Eq    & \lvlE    & \modT & Unified multi-dataset evaluation \\
\textsc{ASDiv}~\cite{miao2020diverse}               & 2020 & --        & 2{,}305   & 2{,}305     & EN     & MWP   & Eq    & \lvlE    & \modT & Lexical and syntactic diversity \\
\textsc{AQuA}~\cite{ling2017program}                & 2017 & 97{,}467  & 254       & 97{,}975    & EN     & MWP   & Rat   & \lvlU    & \modT & End-to-end rationale training \\
\tgroup{11}{Large-scale Chinese MWP}
\textsc{Math23K}~\cite{wang2017deep}         & 2017 & 22{,}162  & 1{,}000   & 23{,}162    & ZH     & MWP   & Eq    & \lvlE    & \modT & Central Chinese MWP benchmark \\
\textsc{Ape210K}~\cite{zhao2020ape210k}      & 2020 & 200{,}488 & 5{,}000   & 210{,}488   & ZH     & MWP   & Eq    & \lvlE--\lvlM & \modT & Largest template-rich Chinese corpus \\
\tgroup{11}{Robustness and perturbation benchmarks}
\textsc{SVAMP}~\cite{patel2021nlp}                  & 2021 & 700       & 300       & 1{,}000     & EN     & MWP   & Eq    & \lvlE    & \modT & Paraphrase, distractor, reordering \\
\textsc{ParaMAWPS}~\cite{raiyan2023paramawps}       & 2023 & 13{,}023  & 3{,}255   & 16{,}278    & EN     & MWP   & Eq    & \lvlE    & \modT & Linguistic variants and voting \\
\textsc{GSM-Symbolic}~\cite{mirzadeh2024gsm}        & 2024 & --        & 5{,}000   & 5{,}000     & EN     & MWP   & Ans   & \lvlE    & \modT & Name/number substitution variance \\
\tgroup{11}{Chain-of-thought and competition benchmarks}
\textsc{GSM8K}~\cite{cobbe2021training}             & 2021 & 7{,}473   & 1{,}319   & 8{,}792     & EN     & MWP   & CoT   & \lvlE    & \modT & Grade-school multi-step reasoning \\
\textsc{MATH}~\cite{hendrycks2021math}  & 2021 & 7{,}500   & 5{,}000   & 12{,}500    & EN     & MWP   & CoT   & \lvlH--\lvlC & \modT & 7 subjects, 5 difficulty tiers \\
\textsc{OlympiadBench}~\cite{he2024olympiadbench}   & 2024 & --        & 8{,}476   & 8{,}476     & EN/ZH  & MWP   & Ans   & \lvlC    & \modM & Bilingual multimodal olympiad \\
\textsc{Omni-MATH}~\cite{gao2024omni}               & 2024 & --        & 4{,}428   & 4{,}428     & EN     & MWP   & Ans   & \lvlC    & \modT & 33 domains, 10+ difficulty levels \\
\textsc{FrontierMath}~\cite{glazer2024frontiermath} & 2024 & --      & ${\sim}$350    & ${\sim}$350    & EN  & MWP   & Ans   & \lvlR    & \modT & Research-grade, private solutions \\
\textsc{MathArena}~\cite{balunovic2026matharena} & 2025 & --        & 149+/yr   & 149+/yr     & EN     & Mixed & Mixed & \lvlC    & \modT & Live competitions, contamination-free \\
\textsc{IMO-ProofBench}~\cite{luong2025imoproofbench} & 2025 & --     & 60        & 60          & EN     & Proof & Proof & \lvlC    & \modT & Proof-grading (basic / advanced splits) \\
\tgroup{11}{Multilingual and regional-language MWP}
\textsc{MGSM}~\cite{shi2022mgsm}            & 2022 & --        & 2{,}500   & 2{,}500     & 10     & MWP   & CoT   & \lvlE    & \modT & Multilingual CoT emergence \\
\textsc{HAWP}~\cite{sharma2022hawp}          & 2022 & --        & 2{,}336   & 2{,}336     & HI     & MWP   & Eq    & \lvlE    & \modT & Hindi equation equivalence \\
\textsc{ArMATH}~\cite{alghamdi2022armath}    & 2022 & 4{,}800   & 1{,}200   & 6{,}000     & AR     & MWP   & Eq    & \lvlE    & \modT & Arabic primary school, transfer \\
\textsc{CMATH}~\cite{wei2023cmath}           & 2023 & --        & 1{,}700   & 1{,}700     & ZH     & MWP   & CoT   & \lvlE    & \modT & Chinese grade-level robustness \\
\textsc{HRM8K}~\cite{ko2025ust}              & 2025 & 7{,}011   & 1{,}000   & 8{,}011     & KO/EN  & MWP   & CoT   & \lvlM    & \modT & Korean comprehension gap \\
\textsc{PatiGonit}~\cite{era2025patigonit}   & 2025 & 8{,}000   & 2{,}000   & 10{,}000    & BN     & MWP   & Eq    & \lvlE    & \modT & Bengali transformer baselines \\
\textsc{BMWP}~\cite{mondal2025bmwp}          & 2025 & 6{,}922   & 1{,}731   & 8{,}653     & BN     & MWP   & Eq    & \lvlE    & \modT & Bengali textbook, operations \\
\textsc{PolyMath}~\cite{wang2025polymath}     & 2025 & --       & ${\sim}$9K   & ${\sim}$9K    & 18  & MWP   & CoT   & \lvlE--\lvlC & \modT & Cross-lingual consistency \\
\textsc{M3Kang}~\cite{torrescamps2026m3kang} & 2026 & --        & 108K var. & 108K var.   & 108    & MWP   & MCQ   & \lvlE--\lvlH & \modM & Multilingual + multimodal \\
\tgroup{11}{Geometry and multimodal math}
\textsc{Geometry3K}~\cite{lu2021inter}              & 2021 & 2{,}101   & 601       & 3{,}002     & EN     & GPS   & FL    & \lvlH    & \modM & Formal-language diagram parsing \\
\textsc{GeoQA}~\cite{chen2021geoqa}                 & 2021 & 3{,}499   & 754       & 4{,}998     & ZH     & GPS   & Prog  & \lvlM    & \modM & Multimodal geometric programs \\
\textsc{PGPS9K}~\cite{zhang2023pgpsnet}             & 2023 & 8{,}022   & 1{,}000   & 9{,}022     & EN     & GPS   & Prog  & \lvlH    & \modM & Fine-grained diagram annotations \\
\textsc{MathVista}~\cite{lu2024mathvista} & 2024 & --      & 6{,}141   & 6{,}141     & EN     & Vis   & Ans   & \lvlE--\lvlC & \modM & 31-source visual math meta-bench \\
\textsc{MathVerse}~\cite{zhang2024mathverse}         & 2024 & --       & 2{,}612   & 15{,}672    & EN     & Vis   & CoT   & \lvlH    & \modM & Diagram ablation, 6 variants \\
\textsc{MATH-Vision}~\cite{wang2024mathvision}      & 2024 & --        & 3{,}040   & 3{,}040     & EN     & Vis   & Ans   & \lvlH--\lvlC & \modM & 16-discipline competition \\
\tgroup{11}{Process, instruction, and augmented data}
\textsc{PRM800K}~\cite{lightman2023verify}   & 2023 & 800K lbl  & --        & 800K lbl    & EN     & PRM   & Step  & \lvlH--\lvlC & \modT & Step-level correctness labels \\
\textsc{Math-Shepherd}~\cite{wang2024mathshepherd} & 2024 & rollouts & --   & rollouts    & EN     & PRM   & Step  & \lvlH--\lvlC & \modT & Annotation-free step estimation \\
\textsc{MetaMathQA}~\cite{yu2024metamath}    & 2024 & 395K      & --        & 395K        & EN     & SFT   & CoT   & \lvlH    & \modT & Backward question rewriting \\
\textsc{NuminaMath}                          & 2024 & 860K+     & --        & 860K+       & EN     & SFT   & Mixed & \lvlH--\lvlC & \modT & Multi-source competition SFT \\
\tgroup{11}{Formal theorem proving}
\textsc{MiniF2F}~\cite{zheng2022minif2f}            & 2022 & --        & 244       & 488         & Multi  & TP    & Proof & \lvlH--\lvlC & \modT & Cross-system formal olympiad \\
\textsc{ProofNet}~\cite{azerbayev2023proofnet}      & 2023 & --        & 371       & 371         & EN     & TP    & Proof & \lvlU    & \modT & Informal/formal statement pairs \\
\textsc{PutnamBench}~\cite{tsoukalas2024putnambench} & 2024 & --       & 658       & 658         & Multi  & TP    & Proof & \lvlC    & \modT & Hardest public formal benchmark \\
\textsc{Lean Workbook}~\cite{ying2024leanworkbook} & 2024 & 57{,}231 & --   & 57{,}231    & EN     & TP    & Proof & \lvlC    & \modT & Large-scale autoformalized \\
\textsc{DeepTheorem}~\cite{zhang2025deeptheorem} & 2025 & 121K  & --        & 121K        & EN     & TP    & CoT   & \lvlC    & \modT & Informal proving + RL-Zero \\
\tgroup{11}{General expert and live evaluation suites}
\textsc{MMLU}~\cite{hendrycks2021mmlu}  & 2021 & --        & 14{,}042  & 15{,}908    & EN     & MCQ   & MCQ   & \lvlU    & \modT & 57-subject academic breadth \\
\textsc{GPQA}~\cite{rein2023gpqa}       & 2023 & --        & 448       & 448         & EN     & MCQ   & MCQ   & \lvlG    & \modT & Graduate-level, Google-proof \\
\textsc{LiveBench}~\cite{white2025livebench} & 2025 & --   & live      & live        & EN     & Multi & Mixed & \lvlH--\lvlC & \modT & Contamination-free, auto-scored \\
\textsc{HLE}~\cite{cais2026hle}         & 2026 & --        & 2{,}500   & 2{,}500     & EN     & Multi & Mixed & \lvlR    & \modM & Expert-authored, Nature 2026 \\
\tablebottom
\end{tabular}%
}
\caption{Dataset and benchmark landscape with concrete scales and structured metadata. \textbf{Solution}: Eq = equation, Ans = numeric answer, CoT = chain-of-thought, Prog = program, FL = formal language, Proof = formal proof, Rat = rationale, MCQ = multiple choice, Step = step labels, Mixed = multiple. \textbf{Level badges}: \lvlE=elementary, \lvlM=middle, \lvlH=high school, \lvlU=undergraduate, \lvlC=competition, \lvlR=research-grade, \lvlG=graduate. \textbf{Modality badges}: \modT=text, \modM=multimodal.}
\label{tab:dataset_landscape}
\end{table*}

\begin{table*}[t]
\centering
\tiny
\setlength{\tabcolsep}{2.6pt}
\renewcommand{\arraystretch}{1.22}
\providecommand{\spv}[2]{{\setlength{\fboxsep}{0.7pt}%
   \colorbox{#1}{\textcolor{white}{\fontsize{4.6}{5.4}\ttfamily\bfseries\,#2\,}}}}
\providecommand{\spvblack}[2]{{\setlength{\fboxsep}{0.7pt}%
   \colorbox{#1}{\textcolor{black!82}{\fontsize{4.6}{5.4}\ttfamily\bfseries\,#2\,}}}}
\providecommand{\trkbar}[1]{%
   \textcolor{#1}{\rule[-.35ex]{1.6pt}{1.7ex}}\hspace{2pt}}
\definecolor{SupvTeal}{HTML}{38A9BF}
\definecolor{SupvAqua}{HTML}{55D1D0}
\definecolor{SupvLime}{HTML}{9BE51C}
\definecolor{SupvYellow}{HTML}{C88F00}
\definecolor{SupvOrange}{HTML}{FFA600}
\definecolor{SupvCoral}{HTML}{F47270}
\definecolor{SupvMagenta}{HTML}{E848A5}
\definecolor{SupvLilac}{HTML}{C872F2}
\definecolor{SupvPurple}{HTML}{8F31F2}
\providecommand{\supA}{\spv{SupvTeal}{A}}
\providecommand{\supE}{\spv{SupvAqua}{E}}
\providecommand{\supC}{\spv{SupvLime}{C}}
\providecommand{\supP}{\spv{SupvYellow}{P}}
\providecommand{\supR}{\spv{SupvOrange}{R}}
\providecommand{\supK}{\spv{SupvCoral}{K}}
\providecommand{\supS}{\spv{SupvMagenta}{S}}
\providecommand{\supV}{\spv{SupvLilac}{V}}
\providecommand{\supH}{\spv{SupvPurple}{H}}
\begin{tabularx}{\textwidth}{@{}p{0.115\textwidth}p{0.195\textwidth}p{0.16\textwidth}p{0.055\textwidth}p{0.13\textwidth}>{\raggedright\arraybackslash}X@{}}
\tabletop
\thead{Track} & \thead{Representative systems} & \thead{Core idea} & \thead{Supv.} & \thead{Main benchmarks} & \thead{Current lesson} \\ \tablemid
\trkbar{VivCyan!75!black}\textbf{Rule/stat MWPs}        & \textsc{WordPro}, \textsc{Aris}, Kushman et al.\                & Problem frames \& log-linear templates           & \supS\,\supE              & \textsc{AI2}, \textsc{Alg514}                                                 & Brittle coverage; high interpretability \\
\trkbar{VivLime!60!black}\textbf{Neural MWPs}           & Seq2Seq, \textsc{GTS}, \textsc{Graph2Tree}, \textsc{MWP-BERT}    & Tree / expression decoding from text             & \supE                     & \textsc{MAWPS}, \textsc{Math23K}, \textsc{ASDiv}                              & Strong in-domain; fragile under perturbation \\
\trkbar{VivYellow!75!black}\textbf{Prompted LLMs}              & CoT, self-cons., ToT, least-to-most                              & Sample natural-language reasoning traces         & \supA                     & \textsc{GSM8K}, \textsc{MATH}                                                 & Fluent traces $\neq$ valid ones; external checks required \\
\trkbar{VivOrange!70!black}\textbf{Tool-int.\ LLMs}            & \textsc{Pal}, \textsc{PoT}, \textsc{ToRA}, calc.\ agents          & NL reasoning interleaved with executable code    & \supC\,\supA              & \textsc{GSM8K}, \textsc{MATH}, tabular QA                                     & Best for arithmetic precision once the problem is parsed \\
\trkbar{VivOrange!85!black}\textbf{Math-spec.\ LMs}            & \textsc{Minerva}, \textsc{Llemma}, \textsc{DSMath}, \textsc{Qwen2.5-Math}, \textsc{MAmmoTH} & Math-heavy pretraining $+$ CoT/PoT SFT             & \supE\,\supC\,\supP       & \textsc{GSM8K}, \textsc{MATH}, AIME, MMLU-STEM                                & Data curation often rivals parameter count \\
\trkbar{VivSalmon!75!black}\textbf{Multilingual LMs}           & \textsc{MathOctopus}, mCoT, UST                                  & Parallel SFT; English-as-anchor reasoning        & \supA                     & \textsc{MGSM}, \textsc{MSVAMP}, \textsc{HRM8K}, \textsc{PolyMath}             & Source-language comprehension is the bottleneck \\
\trkbar{VivPink!75!black}\textbf{Reasoning models}           & OpenAI o1/o3, \textsc{DeepSeek-R1}, Kimi k1.5                    & Long internal search trained with RLVR           & \supR\,\supP              & AIME, \textsc{MATH}, \textsc{FrontierMath}, \textsc{HLE}, \textsc{GPQA}        & Inference budget becomes a first-class variable \\
\trkbar{VivPink!85!black}\textbf{Multi-agent}                & Debate, \textsc{ReConcile}, \textsc{DyLAN}, \textsc{MoA}, \textsc{GoA}, \textsc{MALT} & Propose, critique, route, verify, refine     & \supA\,\supP              & \textsc{GSM8K}, \textsc{MATH}, MMLU-Pro, \textsc{GPQA}                        & Diversity pays only with routing and verifiers \\
\trkbar{VivLilac!75!black}\textbf{Plane-geom solvers}   & \textsc{InterGPS}, \textsc{PGPSNet}, \textsc{GeoDRL}, \textsc{LANS}, \textsc{Pi-GPS}, \textsc{MARS-GPS} & Diagram $\to$ formal clauses; theorem search    & \supS\,\supV\,\supC       & \textsc{Geometry3K}, \textsc{PGPS9K}                                          & Diagram grounding $+$ multi-rollout voting drive gains \\
\trkbar{VivLilac!85!black}\textbf{Visual-CoT MLLMs}     & Visual CoT, Chain-of-Spot, \textsc{Sketchpad}, ICoT, \textsc{MINT-CoT} & Anchor each reasoning step to image regions      & \supV\,\supP              & \textsc{MathVista-Math}, \textsc{GeoQA}, \textsc{MMStar-Math}                 & Visual evidence enters the rationale, not just the input \\
\trkbar{VivLilac!95!black}\textbf{Olympiad geom.}       & \textsc{AlphaGeometry}, \textsc{AlphaGeometry2}                  & Neural proposer $+$ symbolic DDAR engine         & \supS                     & \textsc{IMO-AG-30}, IMO geometry                                              & Domain-specific symbolic engines outpace general MLLMs \\
\trkbar{VivPurple!75!black}\textbf{Formal provers}   & \textsc{LeanDojo}, \textsc{DSProver}, \textsc{AlphaProof}, \textsc{Lean-STaR} & Retrieve premises, decompose goals, gen.\ tactics & \supK\,\supR              & \textsc{MiniF2F}, \textsc{ProofNet}, \textsc{PutnamBench}                     & Correctness is free; autoformalization is the bottleneck \\
\trkbar{VivPurple!95!black}\textbf{Discovery systems}    & \textsc{FunSearch}, \textsc{AlphaEvolve}, Erd\H{o}s workflows & Programs/constructions filtered by objective    & \supC\,\supK\,\supH       & cap-set, bin-pack, live open problems                                         & Needs novelty checks and human/formal validation \\ \tablebottom
\end{tabularx}
\caption{Representative model families across the mathematical-reasoning literature, compressed into a single visual scan. Colored stripes in the \textbf{Track} column group rows by paradigm family, following Figure~\ref{fig:timeline}. The \textbf{Supv.}\ column uses badges in place of prose: \supA\,=\,final-answer grader; \supE\,=\,equation/expression labels; \supC\,=\,code execution; \supP\,=\,process reward model; \supR\,=\,RLVR / verifiable reward; \supK\,=\,proof-assistant kernel; \supS\,=\,symbolic solver; \supV\,=\,visual-region grounding; \supH\,=\,human audit.}
\label{tab:model_landscape}
\end{table*}

\subsection{Math Word Problem Datasets}
Early MWP corpora are small and topic-restricted. \textsc{Ai2}~\cite{hosseini2014learning} contains 395 addition-and-subtraction problems crawled from \texttt{math-aids.com} and \texttt{ixl.com}, divided into three subsets: mixed problems (\textsc{Mar}, 196 problems), irrelevant-information problems (\textsc{Ir}, 121), and multi-step problems (\textsc{Ms}, 78). \textsc{Il}~\cite{roy2015reasoning} contains 562 single-operator problems from \texttt{k5learning.com} (\textsc{Il-Addsub}, 395) and \texttt{dadsworksheets.com} (\textsc{Il-Muldiv}, 167). \textsc{SingleEQ}~\cite{koncel2015parsing} consists of 508 grade-school problems with one unknown. \textsc{AllArith}~\cite{roy2017unit} unifies \textsc{Ai2}, \textsc{Il}, \textsc{SingleEQ}, and \textsc{MA1} into 831 problems with near-duplicate removal, along with the subset \textsc{Perturb} in which quantity perturbations are introduced.

Algebra datasets were constructed to evaluate parameterized templates. \textsc{Alg514}~\cite{kushman2014learning} contains 514 algebra problems crawled from \texttt{algebra.com}. \textsc{Dolphin1878}~\cite{shi2015automatically} is a number-word-problem corpus with 1{,}878 problems and 1{,}183 equation templates. \textsc{Draw1K}~\cite{upadhyay2015draw} contains 1{,}000 algebra problems with a 774/226 train/test split. \textsc{Dolphin18K}~\cite{huang2016well} is a large-scale dataset of 18{,}460 problems with 5{,}871 templates. \textsc{AQuA}~\cite{ling2017program} provides 100{,}949 algebraic problems with rationales, the first corpus large enough for end-to-end neural training. \textsc{MathQA}~\cite{amini2019mathqa} augments a subset of AQuA with structured formal representations.

Multi-operator benchmarks include \textsc{Hmwp}~\cite{qin2021neural}, with 5{,}491 problems covering linear, nonlinear, and simultaneous equations; \textsc{Cm17K}~\cite{qin2021neural} with 17{,}659 problems across arithmetic, linear-system, non-linear-system, and equation-set subsets; \textsc{Mawps}~\cite{koncel2016mawps} unifying six earlier datasets into 3{,}320 problems; \textsc{Asdiv}~\cite{miao2020diverse} with 2{,}305 problems across 25 problem types for measuring lexical and syntactic diversity; and \textsc{Svamp}~\cite{patel2021nlp}, a 1{,}000-problem stress test in which minor variations of \textsc{Mawps} problems (question reordering, adding an irrelevant quantity, swapping object names) dramatically drop the accuracy of previously state-of-the-art systems, revealing their reliance on spurious surface cues.

\textsc{ParaMAWPS}~\cite{raiyan2023paramawps,raiyan2023variational} extends this robustness line in a complementary direction. Instead of only applying controlled template perturbations, it augments selected \textsc{Mawps} problems with paraphrased, adversarial, and inverse variants, then evaluates whether solvers preserve the underlying equation under linguistic change. This makes it especially useful for separating true mathematical invariance from memorized surface-to-equation mappings. The accompanying voting framework also anticipates the later LLM-era pattern of solving multiple transformed versions of the same problem and aggregating the answer.

Large-scale Chinese MWP datasets have played a central role. \textsc{Math23K}~\cite{wang2017deep} crawled 60{,}000 problems from Chinese K-12 websites and retained 23{,}161 for which templates could be automatically extracted; it became the central benchmark for Chinese MWP research. \textsc{Ape210K}~\cite{zhao2020ape210k} contains 210{,}488 problems with 56{,}532 templates, spanning both elementary and middle-school levels. \textsc{Dolphins}~\cite{shi2015automatically} (a small English subset, 7K problems) and derivatives complete the picture.

The watershed \textsc{Gsm8K} benchmark~\cite{cobbe2021training} comprises 8{,}500 high-quality, linguistically diverse grade-school problems requiring 2 to 8 reasoning steps. Its unique role stems from the explicit chain-of-thought annotations, which enabled the supervised verifier training that would later seed the entire reasoning-model paradigm.

\subsection{Multilingual and Non-English Math Benchmarks}
The English-centric view of mathematical reasoning is increasingly inadequate. Classical MWP solvers often failed because of surface linguistic variation; in multilingual settings, the same issue becomes more severe because number words, units, morphology, word order, honorifics, and script conventions all interact with the mathematical parse. Chinese benchmarks were the earliest large non-English success story: \textsc{Math23K} and \textsc{Ape210K} supplied enough scale for neural equation generation, while \textsc{CMATH}~\cite{wei2023cmath} later reframed Chinese MWPs as an LLM evaluation problem using 1.7K elementary-school problems from actual workbooks and exams, organized across six grade levels and augmented with distractor variants.

Low-resource and regional-language datasets reveal a different set of bottlenecks. \textsc{HAWP}~\cite{sharma2022hawp} introduced 2{,}336 Hindi arithmetic word problems and emphasized equation-equivalence evaluation rather than raw string matching. \textsc{ArMATH}~\cite{alghamdi2022armath} contributed 6{,}000 Modern Standard Arabic primary-school MWPs and showed that transfer from a high-resource Chinese solver improved Arabic performance. Turkish MWP corpora~\cite{gedik2023turkishmwp} were constructed by translating and manually correcting \textsc{MAWPS}, \textsc{ASDiv-A}, \textsc{SVAMP}, and \textsc{MathQA}, yielding 4{,}163 elementary problems and a filtered 19{,}555-problem Turkish MathQA subset. Early Korean work translated \textsc{CommonCore} and Illinois arithmetic datasets into \textsc{CC\_Ko} and \textsc{IL\_Ko} for \textsc{KoTAB}~\cite{ki2020kotab}, while \textsc{HRM8K}~\cite{ko2025ust} scales this line to 8{,}011 English--Korean parallel math problems and finds that the gap is mostly in understanding Korean inputs rather than in the underlying reasoning once the problem is correctly represented.

Bangla/Bengali resources have also emerged rapidly. \textsc{PatiGonit}~\cite{era2025patigonit} contains 10{,}000 Bengali math problems and evaluates transformer models including Basic Transformer, mT5, BanglaT5, and mBART50 for equation generation, with mT5 reported as the strongest model. \textsc{BMWP}~\cite{mondal2025bmwp} provides 8{,}653 Bengali arithmetic problems drawn largely from Bengali-medium textbooks, annotated with equations, solutions, and operation classes; it highlights pronoun resolution, compound sentences, irrelevant information, and object-keyword identification as Bengali-specific obstacles. These resources matter for Bangla-language educational AI because direct translation from English does not preserve local curriculum, units, names, or the syntactic phenomena that determine the equation.

Parallel multilingual benchmarks make controlled cross-language comparison possible. \textsc{MGSM}~\cite{shi2022mgsm} manually translates 250 \textsc{GSM8K} problems into ten typologically diverse languages, including Bengali and Swahili, and shows that multilingual CoT ability emerges with scale. \textsc{MathOctopus}~\cite{chen2023mathoctopus} extends this into training data with \textsc{MGSM8KInstruct} (about 73.6K examples across ten languages) and \textsc{MSVAMP} (10K out-of-domain test examples). mCoT-MATH~\cite{lai2024mcot} covers eleven languages for multilingual CoT instruction tuning and focuses on reasoning consistency across languages. More recent resources raise the ceiling: \textsc{PolyMath}~\cite{wang2025polymath} covers 18 languages and four difficulty levels, revealing large language-to-language variation and input-output language inconsistency even for advanced reasoning models; \textsc{MathMist}~\cite{sobhani2026mathmist} builds approximately 30K aligned question--answer pairs across thirteen languages from Bangla-English gold artifacts; and \textsc{M3Kang}~\cite{torrescamps2026m3kang} brings the same question to multimodal reasoning with 1{,}747 Kangaroo competition problems translated into 108 languages, many with diagrams, plus human student baselines.

\begin{table}[t]
\centering
\tiny
\setlength{\tabcolsep}{2.4pt}
\renewcommand{\arraystretch}{1.22}
\begin{tabularx}{\columnwidth}{@{}p{0.27\columnwidth}p{0.07\columnwidth}p{0.14\columnwidth}>{\raggedright\arraybackslash}X@{}}
\tabletop
\thead{Resource} & \thead{Lang} & \thead{Scale} & \thead{Survey-level lesson} \\ \tablemid
\tgroup{4}{Sinitic web crawls}
\textsc{Math23K}, \textsc{Ape210K}      & ZH       & 23K\,/\,210K           & Scale enabled neural non-EN MWPs; template bias persists \\
\textsc{CMATH}                           & ZH       & 1.7K                   & Grade-level + distractor analysis for Chinese LLMs \\
\tgroup{4}{Indic native corpora}
\textsc{HAWP}                                   & HI       & 2.3K                   & Equation-equivalence beats raw string match \\
\textsc{PatiGonit}, \textsc{BMWP}   & BN       & 10K\,/\,8.7K           & Local curriculum cannot be replaced by EN-only tests \\
\tgroup{4}{Other low-resource native \& translated}
\textsc{ArMATH}                                 & AR       & 6K                     & Cross-lingual ZH\,$\to$\,AR transfer helps \\
Turkish MWP corpora                             & TR       & 4.2K\,$+$\,19.6K       & Agglutinative morphology breaks seq2seq transfer \\
\textsc{KoTAB}, \textsc{HRM8K}                  & KO       & 1.2K\,/\,8K            & Korean comprehension dominates the reasoning gap \\
\tgroup{4}{Parallel multilingual}
\textsc{MGSM}, \textsc{MathOctopus}, mCoT       & 10--11   & 2.5K--73K              & Multilingual CoT emerges with scale, but stays uneven \\
\textsc{MathMist}, \textsc{PolyMath}            & 13\,/\,18 & 30K\,/\,9K            & Reasoning models still show language inconsistency \\
\tgroup{4}{Multilingual multimodal}
\textsc{M3Kang}                     & 108      & 1.7K\,$\times$\,108    & Multilingual math must be tested jointly with diagrams \\
\tablebottom
\end{tabularx}
\caption{Non-English and multilingual mathematical-reasoning datasets at a glance. Resources are grouped by construction strategy in the violet sub-headers. \textbf{Lang}\,=\,ISO-style code, or the number of languages for parallel and multilingual-multimodal resources.}
\label{tab:multilingual_math}
\end{table}

\subsection{Competition-level and Olympiad Benchmarks}
The 2021 release of \textsc{MATH}~\cite{hendrycks2021math}, 12{,}500 problems drawn from AMC, AIME, and other US mathematics competitions, with step-by-step solutions and five difficulty tiers across seven subject areas, fundamentally shifted the community's expectation of what ``hard'' meant. Accuracies on \textsc{MATH} climbed from 6.9\% (GPT-3, 2020) to $>95\%$ (reasoning models, 2025).

\textsc{OlympiadBench}~\cite{he2024olympiadbench} extends this frontier with 8{,}476 Olympiad-level problems in mathematics and physics, including bilingual (English/Chinese) and multimodal variants. \textsc{Omni-MATH}~\cite{gao2024omni} collects 4{,}428 competition problems covering 33 domains and over 10 difficulty levels, constructed specifically to evade leakage concerns. \textsc{AIME 2024} and \textsc{AIME 2025} (30 problems each) are the natural continuation: short, difficult, and not-yet-memorized at the time of release, they became the standard measure by which o1, o3, DeepSeek-R1, and Kimi k1.5 were compared.

\textsc{FrontierMath}~\cite{glazer2024frontiermath}, released in late~2024, represents a deliberate effort to outrun the cycle of benchmark saturation. It consists of several hundred problems (exact count withheld; ``several hundred'' verified, approximately 350 in tier 1--3 and a smaller tier~4) authored by expert mathematicians across all major branches of modern mathematics, with a research-grade difficulty tier (tier 4) whose problems require novel ideas rather than application of standard techniques. Solutions are held privately; only numerical final answers are submitted for grading. At release, leading models scored $<2\%$; by April~2025, GPT-5's predecessor o3 reached $\sim 25.2\%$; by late 2025, o4 and Gemini reach higher on tiers 1--2 but remain at single digits on tier 4.

\paragraph*{Live competition evaluation}
A complementary response to the contamination problem is
\emph{dynamic benchmarking} via live competitions.
\textsc{MathArena}~\cite{balunovic2026matharena} evaluates
models on newly released math competitions, including AIME,
HMMT, BRUMO, SMT, and USAMO, within days of problem
release, effectively eliminating memorization.  Across 149
problems from five 2025 competitions, o3~(high), o4-mini~(high), and Gemini~2.5~Pro each exceed $86\%$ average
accuracy on final-answer competitions, outperforming the
top~1\% of human participants.  However, on the
proof-based USAMO~2025 (6 problems, 42 points maximum),
even the best model (Gemini~2.5~Pro, 10.1/42) scores far
below the human median of 15/42, exposing a persistent gap
between answer production and rigorous proof writing.
 
The \textsc{MathArena} framework also provides direct
evidence of contamination in older benchmarks.  By comparing
model scores on AIME~2024 \textit{vs.}\ AIME~2025 against
human-performance quantiles, the authors find that most
models score 10--20 percentage points higher on the 2024
version than their 2025 performance would predict, with one
model (QwQ-Preview-32B) inflated by nearly 60\%.  This
confirms the concern, raised in
Section~\ref{sec:failures}, that reported AIME~2024 scores
should be interpreted with caution.

\subsection{General-Purpose Expert and Live Benchmarks}
Not all mathematically informative benchmarks are math-only. Broad academic suites have become important because frontier systems are now evaluated as general problem solvers: a model that can solve AIME-style algebra but fails at quantitative chemistry, formal logic, or data-analysis questions has not acquired robust mathematical reasoning in the wider sense. \textsc{MMLU}~\cite{hendrycks2021mmlu} initiated this style of evaluation with 57 academic and professional subjects, including elementary mathematics, computer science, law, and history. \textsc{MMLU-Pro}~\cite{wang2024mmlupro} raises the difficulty by adding more reasoning-focused questions, expanding the answer set from four to ten options, and filtering noisy items; the paper reports a 16--33\% accuracy drop relative to \textsc{MMLU} and finds that CoT helps more on the harder benchmark. \textsc{BBH}~\cite{suzgun2023bbh}, a 23-task subset of \textsc{BIG-Bench}~\cite{srivastava2023bigbench}, is less explicitly mathematical but includes algorithmic, logical, and multi-step reasoning tasks for which CoT was shown to substantially change model performance.

Several newer benchmarks make the connection to mathematics more direct. \textsc{AGIEval}~\cite{zhong2023agieval} uses human standardized exams, including SAT Math, math competitions, and college-entrance examinations, thereby measuring mathematical reasoning as part of a broader human-test-taking profile. \textsc{GPQA}~\cite{rein2023gpqa} is a 448-question graduate-level benchmark in biology, physics, and chemistry; although not a mathematics benchmark, it stresses the same skills needed for scientific mathematical reasoning: quantitative inference, symbolic manipulation, and resistance to quick web lookup. \textsc{LiveBench}~\cite{white2025livebench} explicitly targets contamination by releasing fresh, automatically scored tasks from recent sources, including math competitions and arXiv papers, across math, coding, reasoning, language, instruction following, and data analysis. \textsc{SuperGPQA}~\cite{map2025supergpqa} extends the broad-expert paradigm to 285 graduate disciplines and notes that existing evaluations overrepresent mainstream areas such as mathematics, physics, and computer science relative to the full breadth of specialized human knowledge.

The most visible recent example is \textsc{HLE} (\emph{Humanity's Last Exam})~\cite{cais2026hle}, published online in \emph{Nature} on 28~January~2026. \textsc{HLE} contains 2{,}500 expert-authored, closed-ended questions across dozens of subjects, including mathematics, the natural sciences, humanities, and social sciences. Questions may be text-only or multimodal, are either multiple-choice or short-answer for automated grading, and are designed to have unambiguous, verifiable answers that cannot be quickly recovered by internet retrieval. For mathematical reasoning, \textsc{HLE}'s role is complementary to \textsc{FrontierMath}: \textsc{FrontierMath} asks whether models can solve deep mathematics, whereas \textsc{HLE} asks whether mathematical skill transfers into a broad frontier-of-knowledge exam where calibration, domain recognition, and tool-independent reasoning all matter.

A related question is how \emph{scientific-reasoning} benchmarks, suites that test mathematics in service of a scientific argument rather than as an end in itself, bear on the picture surveyed here. We treat them as adjacent rather than central: \textsc{SciBench}~\cite{wang2024scibench} aggregates 869 open-ended college-level problems in physics, chemistry, and calculus that demand multi-step quantitative derivation; \textsc{SciEval}~\cite{sun2024scieval} adds 18K objective and subjective items spanning research-paper comprehension, experimental reasoning, and formula application; and \textsc{ScienceQA}~\cite{lu2022scienceqa} provides 21K multimodal grade-school science questions with stepwise lecture and explanation annotations. \textsc{PhysicsEval}~\cite{siddique2025physicseval} sharpens the physics slice by pairing 19{,}609 textbook-sourced physics problems with solutions scraped from physics forums and educational websites, then evaluating inference-time strategies, including multi-agent verification, on both mathematical and descriptive physics questions. These suites are most informative for our purposes when the underlying reasoning is mathematical, a kinematics problem reduces to algebra; a stoichiometry question, to rational arithmetic; an experimental-design question, to combinatorics. For that reason, we report scientific-reasoning numbers only when they expose a failure mode also visible on mathematical benchmarks (\textit{e.g.}, calibration under contamination-resistant evaluation, or the brittleness of multi-step derivations under surface paraphrase), and otherwise refer the reader to dedicated scientific-reasoning surveys for a complete treatment. The boundary is admittedly porous, and the convergence of mathematical, scientific, and engineering reasoning into a single ``quantitative reasoning'' construct is an open empirical question that we revisit in Section~\ref{sec:future}.

\subsection{Geometry and Visual-Math Datasets}
Classical geometry datasets include \textsc{Geos}~\cite{seo2015solving}, 186 SAT plane-geometry problems; \textsc{Geos++}~\cite{sachan2017textbooks}, 1{,}406 problems from Grade 6--10 geometry textbooks; \textsc{GeoShader}~\cite{alvin2017synthesis}, 102 shaded-area problems; \textsc{Geos-OS}~\cite{sachan-xing-2017-learning}, 2{,}235 problems; \textsc{Geometry3K}~\cite{lu2021inter}, 3{,}002 problems with formal-language ground truth; \textsc{GeoQA}~\cite{chen2021geoqa}, 5{,}010 multiple-choice problems with annotated operation programs; \textsc{GeometryQA}~\cite{tsai2021geometryqa}, 1{,}398 re-annotated problems from a geometry subset of \textsc{Math23K}; and \textsc{PGPS9K}~\cite{zhang2023pgpsnet}, 9{,}022 plane-geometry problems with fine-grained diagram annotations and interpretable solution programs.

The vision-language era introduced broader multimodal benchmarks: \textsc{MathVista}~\cite{lu2024mathvista} with 6{,}141 examples across 31 source datasets; \textsc{MathVerse}~\cite{zhang2024mathverse} with 2{,}612 problems rendered in six variants (15K total) for step-wise CoT evaluation; \textsc{MATH-Vision}~\cite{wang2024mathvision} with 3{,}040 competition problems across 16 disciplines; \textsc{MV-MATH} with multi-visual-context problems; and \textsc{We-Math} targeting fine-grained mathematical-knowledge evaluation. A common finding across these benchmarks is that current multimodal LLMs rely heavily on textual cues: \textsc{MathVerse} reports cases where several MLLMs improve when the diagram is removed~\cite{zhang2024mathverse}. \textsc{MINT-CoT}~\cite{chen2025mintcot} extends this diagnostic direction from benchmark construction to supervision, adding a 54K-example visual-interleaved CoT dataset in which individual reasoning steps are aligned with selected image tokens. This makes step-level grounding a trainable object rather than an \textit{a posteriori} explanation of a completed answer.

\subsection{Tabular Mathematical Reasoning}\label{sec:tabular_math}
A distinctive strand of mathematical reasoning that is neither purely textual nor diagrammatic operates over \emph{semi-structured tables}: numerical inference must traverse rows, columns, and hierarchical headers before any arithmetic can begin. \textsc{TabMWP}~\cite{lu2023dynamic} contains 38{,}431 grade-school problems each paired with a table; solutions are expressed either as free-text answers or as multi-step programs, and the dataset becomes a standard test of whether models can ground numerical references in tabular evidence. In the financial domain, \textsc{FinQA}~\cite{chen2021finqa} provides 8{,}281 expert-annotated question--answer pairs over earnings reports in which the gold rationale is a small executable arithmetic program, while \textsc{TAT-QA}~\cite{zhu2021tat} (16{,}552 questions) and \textsc{MultiHiertt}~\cite{zhao2022multihiertt} (10{,}440 questions over multi-hierarchy tables) raise the difficulty by requiring joint reasoning over text and several nested tables. The more recent \textsc{TabularBench}~\cite{lu2024chameleon}-style suites and \textsc{MultiTabQA}~\cite{pal2023multitabqa} push the frontier toward multi-table aggregation and cross-table joins.

Tabular math is methodologically important for three reasons. \emph{First}, it isolates the \emph{retrieval-plus-arithmetic} sub-skill: the model can only get the answer right if it locates the correct cells \emph{and} composes the right operations, separating comprehension failures from arithmetic failures more cleanly than narrative MWPs allow. \emph{Second}, it is a natural target for tool-augmented reasoning, PoT/PAL approaches that compile the question to a Python or SQL program achieve gains here that are larger than on text-only MWPs, because table grounding maps cleanly onto a structured indexing operation. \emph{Third}, financial and scientific tabular benchmarks (\textsc{FinQA}, \textsc{TAT-QA}, \textsc{MultiHiertt}) are now standard slices in agentic-reasoning evaluations: their gold programs serve simultaneously as supervision for code-generating agents and as verifiable rewards for RLVR-style training. We mention tabular math as a coequal task family alongside textual MWPs, geometry, and formal proving, but defer a fuller treatment to the dedicated table-reasoning surveys; its principal interest here is as evidence that the comprehension--generation--verification triad applies cleanly even when the comprehension stage is dominated by structured-evidence retrieval rather than diagram parsing.

\subsection{Formal-Proof and Autoformalization Datasets}
\textsc{MiniF2F}~\cite{zheng2022minif2f} provides 488 high-school and early-undergraduate problems formalized in Lean, Metamath, Isabelle, and HOL Light. \textsc{ProofNet}~\cite{azerbayev2023proofnet} pairs 371 undergraduate math theorems with informal/formal statements and proofs in Lean. \textsc{PutnamBench}~\cite{tsoukalas2024putnambench} formalizes 658 Putnam Competition problems across Lean~4, Isabelle, and Coq, establishing a new high-water mark for difficulty in formal evaluation. \textsc{Lean Workbook}~\cite{ying2024leanworkbook} contains 57{,}231 math competition problems formalized from natural-language sources, of which 21{,}197 have been verified and carry proofs. \textsc{ProverBench}~\cite{ren2025deepseekproverv2} adds 325 problems including 15 from AIME~24/25. The \texttt{mathlib}~\cite{mathlib2020} library itself, now exceeding 1.6M lines of Lean~4 formalization, serves both as a training corpus and as the premise pool against which retrievers are evaluated.

\subsection{Probing and Functional Benchmarks}
A complementary class of benchmarks attempts not to raise the ceiling but to probe the floor. \textsc{Svamp}~\cite{patel2021nlp}, mentioned above, was the first to systematically demonstrate that surface perturbations could collapse performance. \textsc{GSM-Symbolic}~\cite{mirzadeh2024gsm} extends this idea to the reasoning-model era: by rendering \textsc{Gsm8K} problems as templates and systematically varying surface features, it shows that even state-of-the-art LLMs exhibit significant accuracy variance under name and number substitutions, and that performance degrades more sharply than humans' when additional clauses are appended. \textsc{Putnam-AXIOM} provides an auto-generated perturbation suite for the Putnam problems. The functional benchmarks of~\cite{srivastava2024functional} generate unbounded problem variants to defeat memorization.

\subsection{Performance Across Eras}
\begin{figure*}[t]
\centering
\includegraphics[width=0.98\textwidth]{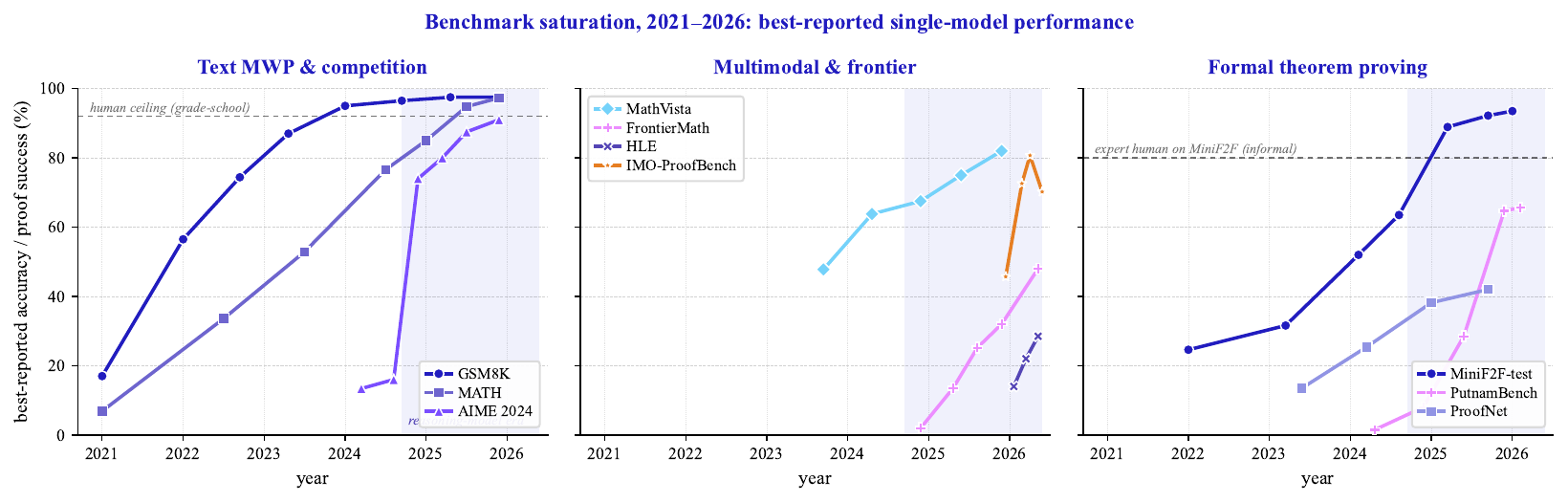}
\caption{Benchmark saturation, 2021--2026: best-reported single-model performance on nine canonical mathematical-reasoning benchmarks, grouped by task family. Pass@1 (or proof-success rate, for formal benchmarks) is plotted against publication year. The lavender band marks the reasoning-model era inaugurated by OpenAI~o1 in late~2024 (Section~\ref{sec:llm}); dashed horizontal lines indicate informal human-performance reference points. Numbers are collected from the system reports and benchmark papers cited throughout the survey; where multiple inference protocols were reported (greedy, majority vote, tool-augmented, expert-formalized), we plot the higher number and discuss caveats in the body text.}
\label{fig:benchmark_trends}
\end{figure*}

Figure~\ref{fig:benchmark_trends} traces the trajectory of best-reported single-model performance on nine canonical benchmarks from 2021 through early~2026, partitioned into the three families that organize the rest of this section: textual MWP and competition tasks, multimodal and frontier evaluations, and formal theorem proving. Three patterns are immediately visible. \emph{First}, \textsc{GSM8K} and \textsc{MATH} have effectively saturated, both clear $95\%$ by 2025, confirming that grade-school and high-school mathematics no longer discriminate between frontier systems. \emph{Second}, the curves for \textsc{AIME~2024}, \textsc{FrontierMath}, and \textsc{PutnamBench} all exhibit a discontinuity at the onset of the reasoning-model era (shaded band), with single-month jumps that exceed the entire 2021--2024 trajectory. \emph{Third}, formal proving, long thought to be the slowest-moving axis, has in fact moved fastest in relative terms: \textsc{MiniF2F-test} rose from roughly $25\%$ in 2022 to $93\%$ by early~2026, and \textsc{PutnamBench} from near-zero to over $60\%$ in a little over a year. Table~\ref{tab:benchmarks} provides the matching numerical summary; Tables~\ref{tab:geo} and~\ref{tab:plane_geo_perf} focus on Olympiad geometry and plane-geometry solvers, respectively.

\begin{table}[t]
\centering
\scriptsize
\setlength{\tabcolsep}{2.5pt}
\renewcommand{\arraystretch}{1.32}
\providecommand{\stripe}[1]{\textcolor{#1}{\rule[-.45ex]{1.6pt}{1.7ex}}\hspace{2pt}}
\providecommand{\lvA}{\cellcolor{VivLime!10}}   
\providecommand{\lvB}{\cellcolor{VivLime!24}}   
\providecommand{\lvC}{\cellcolor{VivLime!38}}   
\providecommand{\lvD}{\cellcolor{VivLime!52}}   
\providecommand{\lvE}{\cellcolor{VivLime!64}}   
\providecommand{\na}{\cellcolor{black!5}\textcolor{black!42}{--}}
\providecommand{\mdl}[1]{\textit{\fontsize{5.2}{5.8}\selectfont(#1)}}
\begin{tabularx}{\columnwidth}{@{}p{0.21\columnwidth}>{\centering\arraybackslash}X>{\centering\arraybackslash}X>{\centering\arraybackslash}X>{\centering\arraybackslash}X@{}}
\tabletop
\thead{Benchmark} & \thead{Pre-DL} & \thead{Seq2Tree} & \thead{LLM era} & \thead{Reasoning era} \\ \tablemid
\stripe{VivCyan!75!black}\textsc{Mawps}             & \lvC$\sim$60\%   & \lvD 92.0\%~\cite{jie2022deductive}     & \lvD $>$95\%                & \lvE $>$99\% \\
\stripe{VivCyan!75!black}\textsc{Math23K}           & \lvC$\sim$60\%   & \lvC 85.4\%~\cite{shen2021generate}     & \lvD $>$90\%                & \lvD $>$95\% \\
\stripe{VivLime!60!black}\textsc{Gsm8K}              & \na               & \lvB 55\%~\cite{cobbe2021training}        & \lvD 94.6\%\,\mdl{GPT-4o}  & \lvE 97.3\%\,\mdl{Kimi K2} \\
\stripe{VivLime!60!black}\textsc{MATH-500}           & \na               & \lvA 6.9\%\,\mdl{GPT-3}                    & \lvB 52\%\,\mdl{GPT-4}      & \lvE 99.2\%\,\mdl{LongCat} \\
\stripe{VivPink!75!black}\textsc{AIME~2024}         & \na               & \na                                         & \lvA 12\%\,\mdl{GPT-4o}     & \lvD 91.6\%\,\mdl{o3} \\
\stripe{VivPink!75!black}\textsc{AIME~2025}         & \na               & \na                                         & \na                          & \lvE 94.6\%\,\mdl{SU-01\,/\,GPT-5.5 H} \\
\stripe{VivOrange!75!black}\textsc{GPQA-D}          & \na               & \na                                         & \lvB 53.6\%\,\mdl{GPT-4o}  & \lvD 94.6\%\,\mdl{Claude Mythos} \\
\stripe{VivPurple!85!black}\textsc{FrontierMath}\textsuperscript{$\ddagger$}    & \na               & \na                                         & \lvA $<$2\%\,\mdl{GPT-4o}   & \lvB 48\%\,\mdl{co-math.~T4} \\
\tablebottom
\end{tabularx}
\caption{Illustrative performance trajectory on canonical MWP, competition, and expert-reasoning benchmarks. Cell shading is a five-step green heatmap of accuracy ($<$25\,\%, 25--60\,\%, 60--90\,\%, 90--98\,\%, $\geq$98\,\%); gray ``--'' cells indicate the benchmark did not exist or was not evaluated in that era. Stripes follow the paradigm palette of Figure~\ref{fig:timeline}. The 2025--2026 entries draw on the LLM Stats live leaderboard~\cite{llmstats2026math}, accessed May~10, 2026, and should be treated as provisional self-reported frontier indicators. $^\ddagger$The 48\,\% \textsc{FrontierMath} figure is the AI co-mathematician's Tier-4 score~\cite{zheng2026comathematician}; the o3 number to its left is the Tier 1--3 average, so the two are not directly comparable.}
\label{tab:benchmarks}
\end{table}

 
\begin{table}[t]
\centering
\scriptsize
\renewcommand{\arraystretch}{1.30}
\setlength{\tabcolsep}{3pt}
\begin{tabularx}{\columnwidth}{@{}>{\raggedright\arraybackslash}p{0.42\columnwidth}>{\centering\arraybackslash}X>{\centering\arraybackslash}X>{\centering\arraybackslash}X@{}}
\tabletop
\thead{System} & \thead{\textsc{IMO-AG-30}} & \thead{\textsc{IMO-AG-50}} & \thead{\textsc{IMOSL-AG-30}} \\ \tablemid
\tgroup{4}{Purely symbolic baselines}
Wu's method\textdagger~\cite{wu1978mechanical,sinha2024wu}                                  & 15/30\textdagger & --    & -- \\
DD+AR (DDAR alone)~\cite{trinh2024alphageometry}                                              & 14/30            & 14/50 & -- \\
Wu $+$ DD+AR ensemble~\cite{sinha2024wu}                                                       & 21/30            & --    & -- \\
\tgroup{4}{Neuro-symbolic geometry solvers}
\textsc{AlphaGeometry} (AG1)~\cite{trinh2024alphageometry}                                    & 25/30            & 27/50 & -- \\
\textsc{AlphaGeometry} $+$ Wu~\cite{sinha2024wu}                                              & 27/30            & --    & -- \\
\textsc{TongGeometry}~\cite{zhang2026tonggeometry}                                            & \textbf{30/30}   & --    & -- \\
\textsc{AlphaGeometry~2} \tiny\textsf{(AG1 setup)}~\cite{chervonyi2025alphageometry2}    & 28/30            & 38/50 & -- \\
\timportant \textsc{AlphaGeometry~2} \tiny\textsf{(full)}~\cite{chervonyi2025alphageometry2} & \textbf{30/30}   & \textbf{42/50} & \textbf{20/30} \\
\tgroup{4}{General reasoning LLMs \scriptsize\textnormal{(formal IMO 2024 eval)}}
OpenAI o1~\cite{openai2024o1}                                                                 & --               & 0/50  & -- \\
Gemini Deep Think~\cite{deepmind2025geminideepthink}                                          & --               & 0/50  & -- \\
\tgroup{4}{Human reference}
Avg.\ human gold medalist                                                                                                 & 25.9/30          & 40.9/50 & -- \\
\tablebottom
\end{tabularx}
\caption{Performance on IMO geometry benchmarks. \textsc{IMO-AG-30}: 30 IMO 2000--2022 problems formalizable in AG1~\cite{trinh2024alphageometry}; \textsc{IMO-AG-50}: broader 2000--2024 set~\cite{chervonyi2025alphageometry2}; \textsc{IMOSL-AG-30}: 30 hard IMO shortlist problems never selected for the contest. \textdagger{}Wu's method was originally reported as 10/30; Sinha et al.~\cite{sinha2024wu} found a JGEX reimplementation solves 15/30. o1 and Gemini Deep Think score 0/50 on the \emph{formal} eval (no symbolic engine); their natural-language IMO performance is much higher. The magenta-shaded row marks the overall state of the art.}
\label{tab:geo}
\end{table}

 
\begin{table*}[t]
\centering
\scriptsize
\setlength{\tabcolsep}{3.5pt}
\renewcommand{\arraystretch}{1.10}
\resizebox{\textwidth}{!}{%
\begin{tabular}{p{0.22\textwidth}cccccc}
\tabletop
\thead{System} & \thead{Type} & \thead{Geo3K} & \thead{PGPS9K} & \thead{GeoQA} & \thead{MathVista-Math} & \thead{MMStar-Math} \\ \tablemid
\tgroup{7}{\textit{Symbolic and neural-symbolic solvers}}
\textsc{InterGPS}~\cite{lu2021inter}                    & Sym.     & 57.5 & --    & --    & --    & -- \\
\textsc{GeoDRL}~\cite{peng2023geodrl}                   & Sym.+RL  & 68.4 & 66.7  & --    & --    & -- \\
\textsc{LANS}~\cite{li2024lans}                         & Sym.+NN  & 82.3 & 74.0  & --    & --    & -- \\
\textsc{Pi-GPS}~\cite{zhao2025pigps}                    & Sym.+MLLM & 77.8 & 69.8 & --    & --    & -- \\
\tgroup{7}{\textit{General-purpose MLLMs (zero-/few-shot)}}
GPT-4o~\cite{achiam2023gpt4}                            & MLLM     & 58.6 & 51.0  & --    & 66.67 & -- \\
Claude 3.5 Sonnet                                       & MLLM     & 56.4 & 45.9  & --    & 67.41 & -- \\
Gemini 1.5 Pro                                          & MLLM     & --   & --    & --    & 63.33 & 52.4 \\
Qwen2-VL-7B-Instruct~\cite{wang2024qwen2vl}             & MLLM     & --   & --    & 37.80 & 41.11 & 46.4 \\
Qwen2.5-VL-7B-Instruct~\cite{bai2025qwen25vl}           & MLLM     & --   & --    & 43.50 & 66.66 & 66.8 \\
InternVL2.5-8B                                          & MLLM     & --   & --    & 34.60 & 49.63 & 54.8 \\
\tgroup{7}{\textit{Reasoning and math-specialized MLLMs}}
R-CoT                                            & MLLM+CoT & --   & --    & 45.78 & 55.93 & 44.0 \\
Math-LLaVA                                       & MLLM+SFT & --   & --    & 47.32 & 54.81 & 55.6 \\
MAVIS-7B                                         & MLLM+SFT & --   & --    & 52.66 & 65.19 & 54.8 \\
R1-V-7B~\cite{chen2025r1v}                        & MLLM+RL  & --   & --    & 59.00 & 53.33 & 54.0 \\
MM-Eureka-Qw-7B~\cite{meng2025mmeureka}           & MLLM+RL  & --   & --    & 40.40 & 70.37 & 63.2 \\
\textsc{MINT-CoT-7B} (SFT)~\cite{chen2025mintcot} & MLLM+SFT & --   & --    & 60.96 & 70.37 & 66.4 \\
\textsc{GeoGRPO-Qw-7B}~\cite{10.1007/978-3-032-09368-4_21} & MLLM+RL & 53.2 & 50.6 & 80.6  & 73.4  & -- \\
\textsc{MINT-CoT-7B}~\cite{chen2025mintcot}  & MLLM+RL  & --   & --    & 64.72 & 73.70 & 69.6 \\
\tgroup{7}{\textit{Multi-rollout / test-time scaling GPS}}
\textsc{MARS-GPS}~\cite{siddique2026marsgps} & MLLM+TTS & 88.8 & 77.48 & --   & --    & -- \\
\tablebottom
\end{tabular}%
}
\caption{Comparison of plane-geometry and visual-math solvers across five benchmarks. ``Type'' abbreviates the system category: Sym.\ = symbolic, NN = neural network, MLLM = multimodal LLM, RL = reinforcement learning, SFT = supervised fine-tuning, and TTS = test-time scaling. Dashes indicate unreported results. Values are aggregated across sources and protocols, so the table should be read as a cross-source comparison rather than a controlled leaderboard; all figures are accuracy (\%).}
\label{tab:plane_geo_perf}
\end{table*}

\section{Cross-Cutting Methodological Synthesis}\label{sec:synthesis}
Across the literature reviewed above, progress is best understood less as a sequence of isolated model architectures and more as a repeated pattern: systems become powerful when they combine expressive generation with a strong external constraint. The constraint may be an equation grammar, a unit-dependency graph, a Python interpreter, a geometry theorem database, a process reward model, or a proof assistant. Each new constraint narrows the search space and supplies a training signal that a pure next-token objective cannot provide.

\subsection{From Representation Engineering to Verifier Engineering}
The earliest systems invested most of their effort in representation engineering: problem frames, templates, quantified sets, semantic parses, equation trees, and unit graphs. Contemporary systems invest more in verifier engineering: exact-answer graders, PRMs, symbolic solvers, execution harnesses, and Lean kernels. The difference is not absolute. Modern systems still need representations, and old systems still used checks. The shift is one of emphasis: instead of hand-designing a representation that will always be correct, researchers increasingly allow large models to propose many candidate representations and rely on external verifiers to keep only the useful ones.

\subsection{The Supervision Ladder}
Mathematical reasoning has advanced by climbing a supervision ladder. At the bottom are final answers; above them are equations and programs; above those are step-level rationales and proof sketches; at the top are mechanically checked proofs. Each rung is more expensive to obtain but more informative when available. This ladder also explains why mathematics is unusually attractive for AI research: unlike many open-ended language tasks, it offers natural sources of objective feedback. The long-term opportunity is to make the top rungs cheaper by using models to generate candidate formalizations and proof attempts, and then using the proof assistant to filter them.

\subsection{Design Principles for Future Benchmarks}
The benchmark literature suggests five design principles. First, benchmarks should separate reasoning difficulty from notation difficulty, because failure to parse a diagram or answer format is different from failure to solve the mathematical core. Second, they should report robustness under paraphrase, number substitution, distractor insertion, and diagram ablation. Third, they should distinguish public training benchmarks from private or live evaluation sets, since leakage becomes more likely as models train on broad web corpora. Fourth, they should record the full evaluation protocol: sample count, tool access, time budget, verifier access, human intervention, and library version. Fifth, they should pair math-specific leaderboards with broad expert suites such as \textsc{HLE}, \textsc{GPQA}, and \textsc{LiveBench}; this reveals whether apparent mathematical competence transfers to scientific and cross-domain tasks rather than remaining a competition-math specialty. Without these details, comparisons across models can reward inference-budget differences more than genuine reasoning differences.

\begin{table*}[t]
\centering
\scriptsize
\renewcommand{\arraystretch}{1.32}
\setlength{\tabcolsep}{4pt}
\providecommand{\stripe}[1]{\textcolor{#1}{\rule[-.45ex]{1.6pt}{1.7ex}}\hspace{2pt}}
\begin{tabularx}{\textwidth}{@{}p{0.175\textwidth}>{\columncolor{Cyan!8}\raggedright\arraybackslash}p{0.255\textwidth}>{\columncolor{Magenta!8}\raggedright\arraybackslash}p{0.245\textwidth}>{\raggedright\arraybackslash}X@{}}
\tabletop
\thead{Era} & \thead{Dominant constraint \scriptsize\textsf{(what makes it tractable)}} & \thead{Typical bottleneck \scriptsize\textsf{(what limits it)}} & \thead{Representative artifacts} \\ \tablemid
\stripe{VivCyan!75!black}\textbf{Rule / statistical MWP}      & Hand-coded schemata and templates                                  & Coverage and brittleness                                 & Problem frames, equation templates \\
\stripe{VivLime!60!black}\textbf{Neural MWP}                   & Supervised expression generation                                   & Generalization outside dataset form                       & Prefix equations, expression trees \\
\stripe{VivYellow!75!black}\textbf{Prompted LLMs}                      & Natural-language reasoning traces                                  & Arithmetic errors and hallucinated steps                  & CoT, self-consistency samples \\
\stripe{VivOrange!70!black}\textbf{Tool-integrated LLMs}                         & Executable interpreters and calculators                            & Program faithfulness to the problem text                  & Python snippets, tool traces \\
\stripe{VivSalmon!75!black}\textbf{Multi-agent systems}                  & Debate, routing, role-specialized critique                         & Correlated errors and communication cost                  & Agent transcripts, confidence votes, graph messages \\
\stripe{VivSalmon!90!black}\textbf{Distilled reasoning}                            & Internalized debate $+$ RLVR on verifiable rewards                  & Token cost; oversight of latent reasoning                 & Long CoT traces, distilled trajectories \\
\stripe{VivPink!80!black}\textbf{Formal provers}          & Proof-assistant kernels and libraries                              & Autoformalization and search cost                         & Lean statements, tactic scripts, proof terms \\
\stripe{VivPurple!85!black}\textbf{Discovery systems}            & Domain-specific evaluators $+$ expert / formal verification         & Novelty, literature audit, and proof certification        & Programs, constructions, verified theorems \\
\tablebottom
\end{tabularx}
\caption{The field as a supervision ladder. Each row pairs an era with the external constraint that made it tractable, the bottleneck that motivated the next transition, and the artifacts it produced. Column tints distinguish constraints from bottlenecks, while era stripes follow the paradigm palette of Figure~\ref{fig:timeline}.}
\label{tab:synthesis}
\end{table*}

\subsection{Cross-Era Patterns}
Reading Table~\ref{tab:synthesis} as a narrative rather than a catalog reveals a recurrent pattern: each generation's key innovation becomes the next generation's baseline assumption, and each generation's bottleneck is precisely what the successor is designed to eliminate.
Templates were revolutionary when they replaced hand-coded schemata, but became a liability when problem language exceeded the template vocabulary.
Supervised expression generation was revolutionary when it replaced templates, but became a liability when evaluation moved beyond in-distribution test sets.
Chain-of-thought prompting was revolutionary when it elicited multi-step reasoning from frozen LLMs, but became a liability when fluent traces proved unreliable without external checking.
At every stage, the resolution was to bring in a stronger external constraint, and the ultimate constraint, a mechanically checked proof, is precisely what the formal track provides.

This pattern has two important implications for practitioners. First, \emph{no single architectural innovation is likely to be sufficient in isolation}; the history of mathematical reasoning is a history of combining generation with progressively stronger verification, and systems that skip the verification step consistently overfit to benchmark form. Second, the pattern predicts where diminishing returns will emerge next: once formal verification of competition-level proofs becomes routine, the bottleneck will shift to autoformalization of research-level mathematics, library scaling, and the sociological challenge of integrating AI-assisted proofs into the human mathematical community.

\subsection{The Convergence Hypothesis}
A central question for the field is whether the four research axes reviewed in this survey, informal reasoning, multimodal reasoning, formal proving, and mathematical discovery, are converging into a single unified pipeline or whether they will remain distinct specialties requiring different models and architectures.

Evidence for convergence is accumulating. The Gemini~Deep~Think result at IMO~2025, scoring 35/42 in natural language, with officially certified gold-medal performance, demonstrates that a single end-to-end system can combine informal intuition, visual diagram interpretation (for geometry problems), and multi-step deductive reasoning without requiring a separate formalization step. Similarly, the Erd\H{o}s-problem workflow of early 2026, in which GPT-5.2 Pro generated conjectures that Harmonic's Aristotle system then formalized and verified in Lean, shows informal and formal tracks operating as complementary stages within a single discovery pipeline rather than as isolated research programs.

However, strong evidence also counsels against premature claims of convergence. The MathArena live evaluation shows that even the best reasoning models score roughly 10/42 on USAMO-style proof problems that require sustained multi-page arguments, compared to a human competition median of 15/42. FrontierMath problems that require novel insight rather than pattern-matching still defeat most models. And the gap between answer-level accuracy (high on AIME, AMC) and proof-level completeness (low on USAMO, PutnamBench) reveals that answer generation and proof generation remain fundamentally different competencies: a model may ``know'' the answer without being able to justify it formally, or may construct a fluent justification that a proof assistant rejects.

The practical resolution may be architectural pluralism within a shared infrastructure: a generator that proposes conjectures and solution sketches in natural language, a verifier that checks them against a formal kernel, a search controller that allocates compute between exploration and exploitation, and domain-specific modules (geometry solvers, code interpreters, retrieval systems) that plug in as needed. The emerging ``verified-discovery workflow'' discussed in Section~\ref{sec:future} is the closest existing approximation to this vision.

\section{Failure Modes, Critiques, and Open Questions}\label{sec:failures}
Enthusiasm for recent progress has been balanced by a growing body of critique, which we survey honestly before turning to future directions.

\subsection{Robustness and Spurious Correlations}
The legacy of classical MWP work in Section~\ref{sec:mwp} already foreshadowed this failure mode: quantity attachment, unit interpretation, irrelevant clauses, and template overfitting remained diagnostic long after explicit templates disappeared. \cite{patel2021nlp}~showed that a seq2seq solver reaching $87\%$ on \textsc{Mawps} dropped to $37\%$ on its own minor perturbations. \cite{mirzadeh2024gsm}~extended this critique to the LLM era: on \textsc{GSM-Symbolic}, models exhibit variance of several percentage points across renderings, and introducing an irrelevant clause (``GSM-NoOp'') causes accuracy drops of up to~65\% even for the strongest models, which frequently change their answer under name and number substitutions. These results suggest that much of the improvement visible in benchmark accuracies is accompanied by only partial gains in genuine robustness.

The deeper question raised by \textsc{GSM-Symbolic} is
\emph{what kind of failure} the observed variance represents.
Three hypotheses compete.  The \emph{training-data}
hypothesis holds that models memorize problem--answer
associations from web-scale corpora and degrade when surface
features shift; contamination-controlled benchmarks such as
\textsc{FrontierMath}~\cite{glazer2024frontiermath} support
this reading.  The \emph{architectural} hypothesis, advanced
by Apple researchers~\cite{mirzadeh2024gsm}, holds that
autoregressive next-token prediction is fundamentally
misaligned with the compositional, non-monotonic structure of
mathematical reasoning: the model must commit to early tokens
before ``seeing'' later constraints.  The \emph{emergent but
incomplete} hypothesis, favored by the reasoning-model
community, holds that long CoT and RLVR do produce genuine
compositional reasoning, but that this capability is
\emph{shallow}, reliable on well-represented problem types
and brittle under rare structural variations.  Distinguishing
these hypotheses experimentally would require two
methodological commitments that are not yet standard practice:
\emph{distribution-controlled} evaluations that hold problem
structure fixed while varying surface
features~\cite{mirzadeh2024gsm,patel2021nlp}, and
\emph{causal interventions} on intermediate reasoning steps
that test whether each step is functionally necessary for the
final answer.

\subsection{Metric Mismatch and Path Optimality}
Recent surveys also stress that final-answer accuracy is a lossy proxy for mathematical reasoning quality~\cite{wang2025survey,liu2025mathematicallm}. Pass@1 rewards the first sampled answer; Pass@k rewards diversity and search; majority voting rewards agreement; and proof compilation rewards formal validity. These metrics can disagree. A model may improve Pass@1 while leaving Pass@k nearly unchanged, suggesting better selection rather than broader reasoning capacity; conversely, long-CoT or RL systems may improve Pass@k by producing more diverse traces while still generating inefficient or locally invalid derivations.

This creates a path-optimality problem. Outcome-based training often treats a verbose, circuitous, or partially mistaken derivation as acceptable if the final answer is correct. For educational applications, theorem proving, and human-AI collaboration, this is insufficient: the reasoning path should be concise, checkable, and faithful to the mathematical dependencies actually used. We therefore recommend reporting answer accuracy together with at least one process-sensitive measure, such as verifier pass rate, step-level PRM score, average reasoning tokens per solved problem, or the fraction of solutions accepted by an external proof checker.

The emerging literature on step-level evaluation offers
partial solutions.  PRM scores provide a continuous proxy for
step quality but are themselves trained on data that may
conflate fluency with validity.
\textsc{DeepTheorem}~\cite{zhang2025deeptheorem} operationalizes
a four-dimensional rubric (logical validity $40\%$,
completeness $30\%$, correctness $20\%$, clarity $10\%$) and
shows that process scores and outcome scores can diverge
substantially:\ a model may produce a correct final judgment
(proved/disproved) via an incomplete or locally invalid
argument.  More ambitiously, \emph{proof economy}, the ratio
of useful reasoning steps to total generated tokens, is
beginning to be tracked, motivated by the observation that
long-CoT models often spend thousands of tokens on
exploration that contributes nothing to the final answer.
For educational and collaborative applications, these
process-sensitive metrics matter more than Pass@1:\ a
student or mathematician working with an AI assistant needs
not just the right answer but a derivation they can trust,
extend, and learn from.

\subsection{Benchmark Contamination and the Race to Saturation}
A second class of concerns relates to training-set contamination. The sheer scale of LLM pretraining makes it plausible that test-problem formulations (and, in some cases, solutions) appear in the training corpus. Cautious researchers now publish benchmarks with delayed release schedules (\textsc{FrontierMath}), live versions (\textsc{LiveMathematicianBench}, \textsc{RealMath}), functional generators \cite{srivastava2024functional}, expert-written retrieval-resistant suites (\textsc{HLE}, \textsc{GPQA}), or monthly refreshed objective tasks (\textsc{LiveBench}). Empirical evidence for contamination is strongest on classical benchmarks; results on newly-constructed olympiad and research-level benchmarks are more difficult to explain by contamination alone.

The community's methodological response to contamination has
developed along five lines, each with trade-offs.
\emph{Delayed release} (\textit{e.g.}, \textsc{FrontierMath} holding
solutions privately) prevents direct leakage but limits
reproducibility and community auditing.  \emph{Live
benchmarks} (\textsc{LiveBench}, \textsc{LiveMathematicianBench})
refresh tasks periodically, defeating memorization but
requiring continuous curation effort.  \emph{Functional
generators}~\cite{srivastava2024functional} produce unlimited
problem variants from parameterized templates, ensuring that
no specific instance can be memorized, but they risk testing
only the narrow structural family captured by the template.
\emph{Expert-authored suites} (\textsc{HLE}, \textsc{GPQA})
rely on the obscurity and difficulty of the questions to make
web-lookup infeasible, but they are expensive to produce and
may themselves leak over time.  \emph{Embedding-based
decontamination} (as used by
\textsc{DeepTheorem}~\cite{zhang2025deeptheorem}, which
removed $\sim$199K contaminated samples via cosine-similarity
recall and LLM-based justification) addresses training-data
hygiene rather than benchmark design.  No single strategy is
sufficient; the strongest evaluations combine at least two
(\textit{e.g.}, live tasks with functional variants, or
expert-authored questions with delayed solutions).

\textsc{MathArena}'s contamination analysis provides the
strongest empirical evidence to date that AIME~2024
scores are inflated by data leakage.  The study also
reveals a subtler form of contamination: 8~of~30
AIME~2025 problems and 1~of~30 HMMT~2025 problems
appeared online in similar form before the competition,
even though the competitions themselves were new.  This
``prior-problem leakage'' is distinct from
training-data contamination and is harder to detect,
since the leaked problems may come from online forums
or earlier competitions rather than from the benchmark
itself.  The implication is that even live evaluation is
not fully contamination-proof; the strongest protocol
combines live timing with cross-competition correlation
analysis to identify anomalously easy problems.

\subsubsection{Recommendations for Survey-Level Reporting}
The discussion above implies several standards that future surveys and leaderboards should adopt when compiling results across systems:
\begin{enumerate}
    \item \textbf{Decontamination audits:} These should be reported for every training-data pipeline. At minimum, this involves $n$-gram overlap ($n\!\geq\!13$) between the training corpus and each benchmark's test split, supplemented by embedding-based cosine-similarity recall (as used by \textsc{DeepTheorem}~\cite{zhang2025deeptheorem}, which removed ${\sim}$199K contaminated samples) and, where feasible, LLM-based justification for flagged near-matches.
    \item \textbf{Inference budget reporting:} Every accuracy figure should be accompanied by its \emph{inference budget}: the number of sampled solutions ($k$), the selection mechanism (greedy, majority vote, ORM, PRM, execution, or Lean checking), and an approximate token cost per problem. Without these annotations, a ``$90\%$ on \textsc{MATH}'' claim is ambiguous by at least $20$ percentage points depending on whether it reflects pass@1 or best-of-256 with a PRM.
    \item \textbf{Cross-paper comparison flags:} For compiled benchmark tables that aggregate results from multiple papers, authors should explicitly flag which figures come from the original paper and which are reproduced under controlled conditions, since minor differences in prompting, sampling temperature, and evaluation harness can shift scores by several points.
\end{enumerate}

\subsection{Reward Hacking in RLVR Training}
 
The RLVR paradigm that underwrites reasoning
models~\cite{shao2024deepseekmath,guo2025deepseekr1} assumes
that the verifiable reward, typically exact-match comparison
of a final numerical answer, faithfully represents the
training objective.  In practice, this assumption fails in
several documented ways.  First, rule-based verifiers are
surprisingly inaccurate: a recent analysis of a standard
empirical reports suggest that a substantial fraction (up to
$38\%$) of responses flagged as incorrect by a rule-based grader
were in fact correct, because the grader could not handle equivalent but
differently formatted answers (\textit{e.g.}, $\frac{12}{36}$
\textit{vs.}\ $\frac{1}{3}$).  This false-negative rate deprives the
model of informative gradients and slows convergence.
Conversely, false positives, in which a lucky final answer
receives reward despite an invalid derivation, reinforce
hackable surface patterns.
 
Second, models trained with outcome-only RLVR can learn to
exploit format cues: revealing the answer early in the
reasoning trace, generating repetitive or templated padding
to reach the expected length, or producing correct-looking
\latex that embeds the answer without genuine derivation.
Third, when model-based verifiers (\textit{e.g.}, learned PRMs) replace
rule-based graders, a subtler failure emerges: the policy
model learns to produce traces that \emph{score highly on the
verifier} without being mathematically valid, a
classical Goodhart's-law dynamic.  Empirical work shows
that more accurate verifiers do not always produce better RL
outcomes; in some cases, higher-accuracy verifiers are
\emph{more} susceptible to hacking during training because
the policy has a richer signal to exploit.
 
These findings have practical implications for the survey's
central claim that verification drives progress.
Verification improves reasoning only when the verifier is
(i)~accurate, (ii)~resistant to adversarial exploitation by
the generator, and (iii)~rich enough to provide gradient
signal on partial progress.  Kernel-checked formal proofs
satisfy all three conditions, which is one reason the formal
track has advanced so rapidly.  Rule-based answer graders
satisfy (iii) cheaply but fail on (i) and (ii), which
explains why outcome-only RLVR produces models that are
strong on benchmarks but fragile under perturbation.

\subsection{Multimodal-Specific Failure Modes}
 
The \textsc{MathVerse} finding that some MLLMs improve when
diagrams are removed~\cite{zhang2024mathverse} is the
best-known multimodal failure, but it is not the only one.
At least three additional failure classes deserve attention.
First, \emph{diagram hallucination}: models sometimes
``see'' geometric elements that are not present in the image,
or misidentify which angle or segment a label refers to; this
is especially common when diagrams are low-resolution,
contain overlapping labels, or use non-standard notation.
Second, \emph{modality shortcutting}: when the textual
problem statement contains enough information to solve the
problem without the diagram, models learn to ignore visual
input entirely; this is not a ``bug'' but a rational
response to training distributions in which text is more
reliably informative than images.  \textsc{MINT-CoT}'s
interleave-token mechanism~\cite{chen2025mintcot} is
explicitly designed to counter this shortcut by forcing the
model to select image regions before each reasoning step.
Third, \emph{cross-modal grounding failure}: the model
correctly parses both text and image but fails to align
them, for instance, identifying $\angle ABC$ in the text
but measuring $\angle ABD$ in the diagram.  This failure is
diagnostic of a deeper architectural limitation: current
multimodal encoders fuse text and image at a global level
rather than maintaining fine-grained correspondence between
symbolic names and spatial locations.

\subsection{Hallucination in Mathematical Derivations and Proofs}
 
A failure mode that cuts across informal and formal tracks is
\emph{mathematical hallucination}: the generation of
derivation steps that are syntactically well-formed and
superficially plausible but logically invalid.  Unlike
factual hallucination in open-domain QA, mathematical
hallucination is dangerous precisely because it is
\emph{difficult to detect without line-by-line verification}.
Common patterns include: (i)~citing a theorem that does not
exist or does not apply to the given hypotheses;
(ii)~introducing an unjustified inequality or bound that
happens to yield the correct final answer;
(iii)~silently dropping a case in a case analysis; and
(iv)~circular reasoning in which the conclusion is assumed
in a disguised form.  The \textsc{DeepTheorem} process
evaluation framework~\cite{zhang2025deeptheorem}, which
separately scores logical validity, completeness,
correctness, and clarity, provides one operational response
to this problem.  However, such evaluations currently rely on
LLM judges (GPT-4o in that work), which are themselves
susceptible to the same hallucination patterns they are asked
to detect.  To quantify this, the \textsc{DeepTheorem} audit revealed that among logically flawed informal proofs generated by state-of-the-art models, roughly $40\%$ suffer from unjustified inferential leaps, $35\%$ from calculation or algebraic errors, and $25\%$ from circular or logically invalid structural assumptions. Furthermore, the nature of these hallucinations has evolved: whereas older prompted models typically failed by abandoning the logical thread entirely (generating non-sequiturs), modern long-CoT reasoning models are far more likely to produce \emph{coherent but circular} arguments or subtly drop inconvenient cases deep within a 5,000-token trace. 

This evolution directly connects back to the supervision ladder discussed in Section~\ref{sec:synthesis}. Outcome supervision (the lowest rung) is blind to these long-trace structural errors, as it only checks the final answer. Step-level process reward models (PRMs, the middle rung) can catch algebraic slips and local non-sequiturs, but often fail to detect circular reasoning that spans multiple paragraphs. Consequently, the only hallucination-proof verification mechanism available today is a proof-assistant kernel (the highest rung), which is precisely why the formal track exists, but at the cost of formalization effort that remains prohibitive for most mathematical communication.

\subsection{Language Transfer and Localization}
Multilingual evaluations show that mathematical reasoning is not language-neutral in practice. A model may possess the algebraic skill needed to solve a problem but fail to parse the problem statement in Korean, Bangla, Arabic, Hindi, Turkish, or Swahili. Translation-based benchmarks such as \textsc{MGSM} and \textsc{PolyMath} are valuable because they hold mathematical content approximately constant across languages; native or localized datasets such as \textsc{HAWP}, \textsc{ArMATH}, \textsc{BMWP}, and \textsc{PatiGonit} are valuable because they expose curriculum, names, units, discourse patterns, and morphology that translations often erase. For this reason, multilingual results should report the problem language, reasoning language, answer language, and whether the model was allowed to translate internally. Otherwise, an English-anchored system can appear mathematically stronger than it is for users who actually need reliable non-English educational support.

\subsection{The ``Genuine Reasoning'' Question}
A third, and more philosophical, question concerns whether reasoning models are \emph{actually} reasoning or are very effective pattern-matchers over reasoning trajectories. Apple researchers~\cite{mirzadeh2024gsm} have argued for the latter. Tao~\cite{tao2024copilot,tao2026primetime} has taken a more nuanced position: he observes that reasoning models are now extremely strong at the ``discovery modulo expertise'' regime, connecting a problem to an existing technique, recalling the relevant literature, and producing a candidate proof, but remain weak at introducing genuinely novel ideas. This is consistent with both the AlphaEvolve results (improvements of~$\sim 20\%$ on the 67-problem benchmark, mostly via clever application of known techniques) and the Erd\H{o}s-problem solves (which, in Tao's phrasing, involve ``lowest hanging fruit'').

\subsection{Formal \textit{vs.} Informal and the Cost of Verification}
The formal track exists precisely to counter the
fluency-correctness gap. However, the practical question is
whether the formalization bottleneck is shrinking fast enough
to make the formal track useful at scale within the 12--24
month horizon considered by this survey.  The evidence is
mixed.  On the positive side, \textsc{Lean Workbook}'s
first-attempt type-check rate of $36.5\%$ can be raised
substantially with iterative refinement and type-checking
feedback~\cite{poiroux2024typechecking}; Tao's experience
suggests that the human--AI collaborative de~Bruijn factor
is already below the solo-human factor for some problem
classes~\cite{tao2024machineassisted}; and the
\textsc{DeepSeek-Prover-V2} pipeline demonstrates that
decomposition and verified subproofs can scale formal proving
to problems previously considered out of reach.  On the
negative side, the set of mathematical domains covered by
\texttt{mathlib} remains a small fraction of research
mathematics; autoformalization of novel definitions, as
opposed to textbook theorems, remains unreliable; and the
compute cost of formal proof search is orders of magnitude
higher than informal generation, which limits the
technique to high-value targets.  A realistic assessment is
that formal verification will be routinely available for
competition-level problems and well-formalized domains (real
analysis, group theory, basic combinatorics) within two
years, but that research-level formalization across the full
breadth of mathematics will require at least a further
generation of library expansion and autoformalization
improvement.

\subsection{Multi-Agent Coordination and Correlated Errors}
Multi-agent mathematical reasoning introduces its own failure modes. A debate protocol can converge on the wrong answer if all agents share the same misconception; a judge model can favor eloquent but invalid explanations; and adding irrelevant specialist agents can actively degrade performance, as \textsc{Graph-of-Agents} demonstrates when comparing full-pool aggregation against relevance-aware node sampling~\cite{yun2026goa}. Communication cost is also non-trivial: naive all-to-all discussion grows with the number of agents and the length of their traces. The AI co-mathematician deployment of Zheng et al.~\cite{zheng2026comathematician} adds two further failure modes specific to research-assistant pipelines: \emph{reviewer-pleasing bias}, in which reviewer agents converge on a flawed argument because they share the inductive biases of the proposer agents that produced it, and \emph{non-termination death spirals}, in which iterative review loops fail to converge and degrade into hallucinated reasoning that a human must recognise and break out of. A related observation from the same deployment: high-quality LaTeX typesetting can create a false impression of rigor when the underlying argument is broken, a presentation-layer phenomenon to which formal verifiers are immune but human readers are not. For mathematics, the most promising designs therefore combine agent diversity with external checks, Python execution, PRM scores, symbolic solvers, or Lean kernels, rather than relying on dialogue alone.

\subsection{Energy, Carbon, and Access}
Reasoning-model inference is dramatically more expensive than
single-forward-pass inference, and the cost is growing with
each generation.  The Arc Prize Foundation's revised
estimates place the cost of solving a single ARC-AGI problem
with o3-high at approximately \$30{,}000 per task,
reflecting the combinatorial search over $1{,}024$ candidate
solutions of roughly $137$ pages each.  At the API level,
o3 is priced at \$10 per million input tokens and \$40 per
million output tokens; a single hard mathematical problem
that requires $50{,}000$--$100{,}000$ reasoning tokens thus
costs \$2--\$4, and budget-uncapped search can push this
orders of magnitude higher.  By contrast, o4-mini and
distilled models such as DeepSeek-R1-Distill-7B operate at
roughly $1/10$--$1/100$ of this cost, demonstrating that the
trade-off between reasoning depth and compute budget is
already a design variable.
 
The distributional consequences are significant.  If the
strongest mathematical reasoning requires frontier-lab
APIs and multi-thousand-dollar inference budgets, then
AI-assisted mathematics risks becoming the exclusive
purview of well-funded institutions, widening the gap
between research universities in wealthy countries and the
rest of the world.  The open-weights movement (DeepSeek-R1,
Qwen-Math, Gemma-Math) partially addresses this by enabling
local deployment, but even local deployment of a 671B model
requires hardware that is unavailable to most research groups
in the Global South.  Distilled models at the 1.5B--7B scale
offer the most realistic path to broad access, and their
performance on benchmarks like AIME (\textit{e.g.},
\textsc{DeepScaleR}'s $43.1\%$ at 1.5B) suggests that strong
mathematical reasoning need not be confined to
frontier-scale parameters.  Nevertheless, the community
should be explicit about the compute assumptions underlying
reported results: a system that achieves $91.6\%$ on AIME
with $1{,}000$-sample reranking at \$10 per sample occupies
a fundamentally different point on the Pareto frontier from
one that achieves $79.8\%$ with greedy single-sample
decoding.

\subsection{The October 2025 Erd\H{o}s Incident}
A recent cautionary tale is the October~2025 OpenAI claim, rapidly retracted, that GPT-5 had autonomously solved ten open Erd\H{o}s problems. T.~Bloom, the curator of erdosproblems.com, pointed out that all ten were in fact literature lookups: the system had located existing papers that the database had marked as open due to incomplete cataloging. The incident is doubly instructive. First, it underscores the importance of careful benchmark auditing by domain experts before accepting model claims. Second, and more positively, it prompted the development of clear protocols for what constitutes an ``AI solve'' of an open problem, including the requirement of autonomous discovery, independent verification, and (increasingly) formal verification in Lean. The four subsequent bona fide Erd\H{o}s-problem solves (\#1026, \#728, \#729, \#397) in late~2025 and early~2026, each accompanied by Lean formalization, suggest that the community has absorbed this lesson quickly.

\section{Future Directions}\label{sec:future}
Synthesizing across the four axes, we identify ten directions in which the field appears most likely to advance over the next 12--24~months.

\subsection{Verified-Discovery Workflows}
The most consequential architectural shift of the past eighteen months has been the emergence of a canonical four-stage pipeline: (i)~an LLM proposes a candidate answer or proof sketch; (ii)~an informal LLM proof fills in the steps in natural language; (iii)~an autoformalizer translates the proof into Lean~4; (iv)~Lean mechanically verifies the result. Each stage has improved rapidly, but the \emph{interfaces} between stages remain brittle: autoformalization still fails on many mathematically natural statements, and type-checking errors rarely propagate useful information back into proof revision. A natural research agenda is to train these stages jointly with end-to-end reward from the final verification step, in the same spirit that AlphaZero learned policy and value through a unified training loop.

\subsection{Research Assistance Beyond Competitions}
The IMO-gold result of mid-2025 and the Erd\H{o}s solves of early~2026 suggest that competition-style evaluation is approaching saturation. The next frontier is research-level mathematics: improvements to existing theorems, counterexamples to standing conjectures, and explicit constants or bounds in results that were previously only asymptotic. Benchmarks such as \textsc{RealMath}, \textsc{LiveMathematicianBench}, and \textsc{ResearchBench} have begun to operationalize this shift by measuring not single-problem accuracy but end-to-end productivity in tasks drawn from the actual research workflow. Specifically, \textsc{RealMath} draws problems directly from newly published mathematics papers to evaluate frontier capabilities on unmemorized concepts, while \textsc{LiveMathematicianBench} tests proof-generation models against live, frequently refreshed theorems.

\subsection{Reasoning Efficiency and Open Models}
Frontier reasoning models today require tens of thousands of output tokens per non-trivial problem. Substantial academic and industrial effort is therefore directed at \emph{reasoning efficiency}: distilled models that retain most of the capability at a fraction of the compute; methods for adaptive reasoning budgets that deliberate only as long as necessary; and open-weight models that democratize research access. The DeepSeek-R1-Distill series, Gemma-Math, and Qwen-Math-RL lines demonstrate that strong reasoning need not be confined to frontier-lab APIs; this is likely to become a central axis of competition.

\subsection{Multi-Agent Orchestration}
The next generation of multi-agent mathematical systems should move beyond generic debate toward role- and domain-aware orchestration: one agent searches for invariants, another writes executable checks, another attempts a formalization, another critiques hidden assumptions, and a router allocates budget dynamically. \textsc{Graph-of-Agents} suggests that selecting a small, relevant subgraph can outperform using every available agent~\cite{yun2026goa}; \textsc{MAgICoRe} and \textsc{MALT} suggest that reviewer and verifier roles can be converted into training signal~\cite{chen2025magicore,motwani2025malt}. The open problem is credit assignment: when a multi-agent solve succeeds, which agent, message, or verification step deserves the learning signal?

\subsection{Multilingual and Localized Reasoning}
The next wave of mathematical-reasoning benchmarks should be multilingual by design rather than translated only after English saturation. The key research questions are whether models can reason in the user's language, whether they silently translate through English, whether code-switched rationales improve or damage reliability, and whether local textbook distributions are represented. Resources such as \textsc{HRM8K}, \textsc{PatiGonit}, \textsc{BMWP}, \textsc{MathMist}, \textsc{PolyMath}, and \textsc{M3Kang} make it possible to study these questions systematically, including for Bangla and other languages that have historically been absent from math-reasoning leaderboards.

\subsection{Neuro-Symbolic Integration Beyond Geometry}
\textsc{AlphaGeometry} demonstrated the power of coupling a neural proposer with a specialized symbolic engine for geometry. The natural generalization, coupling LLMs with decision procedures, SMT solvers, and computer-algebra systems across domains, remains comparatively underexplored. Early work on LLM + Z3, LLM + SageMath, and LLM + Lean-hammer pipelines indicates substantial gains over pure-LLM baselines. We expect this line of work to expand sharply, particularly for inequality proving, real analysis, and algebraic geometry.

\subsection{The Verifiable-Reward Frontier}
The reinforcement-learning-from-verifiable-rewards (RLVR) paradigm, introduced in~\cite{shao2024deepseekmath,openai2024o1,guo2025deepseekr1}, currently relies on exact-match graders for final answers (\textsc{Gsm8K}, \textsc{Math}, \textsc{Aime}) and for formal-proof completion (Lean). Extending this paradigm to \emph{proof quality}, \emph{generalizability of a solution strategy}, and \emph{economy of argument} remains an important open problem. The PRM literature~\cite{lightman2023verify,wang2024mathshepherd,luo2024improve} offers partial answers, but a principled theory of step-level reward modeling is still lacking.

\paragraph*{Curriculum and difficulty scheduling}
The effectiveness of RLVR depends not only on the reward
signal but also on the \emph{distribution of training
problems}.  \textsc{WizardMath}~\cite{luo2023wizardmath}
applies iterative difficulty amplification;
\textsc{MetaMath}~\cite{yu2024metamath} uses backward
question rewriting;
\textsc{DeepTheorem}~\cite{zhang2025deeptheorem} filters for
difficulty $\geq 5$ and constructs entailing/contradictory
theorem variants to ensure non-trivial binary rewards.
More systematic curriculum design, scheduling problem
difficulty to match the model's current capability,
progressively introducing formal constraints, and balancing
domain coverage, remains underexplored.  The RL literature
outside NLP offers relevant frameworks (self-paced learning,
automatic curriculum generation, regret-based task
selection), but their adaptation to mathematical reasoning
training at the scale of millions of problems is an open
engineering and research challenge.  Getting this right could
substantially reduce the compute required for reasoning-model
training, which is itself a significant access and
sustainability concern (Section~\ref{sec:failures}).

\subsection{Reasoning Robustness and Uncertainty}
The GSM-Symbolic results highlight the continued lack of robust reasoning under surface perturbation, particularly under the introduction of irrelevant information. Work on calibrated uncertainty for reasoning models, abstention under ambiguity, and adversarial robustness via data augmentation represents an essential complement to raw-accuracy scaling. The emergence of benchmarks like MathCheck and Putnam-AXIOM~\cite{srivastava2024functional} gives the community the measurement instruments required to track this axis.

\subsection{Community Infrastructure for Formalization}
Finally, as Tao~\cite{tao2024machineassisted,tao2026primetime} has repeatedly emphasized, the bottleneck for AI-assisted formalization is no longer model capability alone but \emph{community infrastructure}: the formal libraries required to state and prove research-level results, the lightweight tooling that makes Lean approachable for early-career mathematicians, and the cultural acceptance of formalization as a legitimate mode of mathematical work. Reducing the de Bruijn factor below one would change formalization from a costly after-the-fact certificate into a productivity multiplier. Initiatives such as the Lean FRO, the Lean Focused Research Organization, and the \texttt{mathlib} community's expansion to undergraduate curricula represent the kind of investment that will determine whether AI-for-mathematics becomes a narrow technical specialty or a broadly transformative tool. Microgrants for postdocs and graduate students in formalization, structured visiting programs at formalization hubs, and teaching material integrating Lean with standard curricula are concrete levers for the community to pull.

\subsection{Evaluation Protocols for Assisted Mathematics}
The next generation of evaluations should measure not only whether a model can answer a fixed problem, but whether it improves the productivity of a mathematically competent user. Such protocols should record the division of labor between human and model, the number of failed attempts, whether the model searched the literature, whether the final proof was formalized, and how much expert editing was required. This is especially important for open-problem claims, where the distinction between rediscovery, literature retrieval, heuristic suggestion, and autonomous proof generation determines the scientific meaning of the result.

\section{Limitations}\label{sec:survey_limitations}
This survey covers a rapidly moving field in which frontier results are often first reported through technical reports, benchmark websites, public demonstrations, or project repositories rather than peer-reviewed publications. We therefore annotate source status where possible, but some post-2024 results remain difficult to compare because complete model details, inference budgets, tool access, and verifier settings are not always public. The performance tables aggregate numbers reported under heterogeneous protocols, greedy decoding, majority voting, pass@k, tool-augmented inference, learned reranking, expert formalization, and proof-assistant feedback, so they should be read as a structured map of the literature rather than as a strictly controlled leaderboard. Similarly, open-problem and discovery claims require special care: a reported AI contribution may range from literature retrieval or candidate generation to fully formalized proof production, and only the latter provides a strong correctness certificate. Finally, although the survey emphasizes Lean~4 because of its current centrality in AI-for-mathematics, the supervision-ladder analysis, the comprehension--generation--verification triad, and the discovery-vs-verification division of labor all apply \textit{mutatis mutandis} to Isabelle, Coq/Rocq, and emerging assistants; we expect that the specific systems will be displaced more rapidly than the structural picture, and that future iterations of this survey will rename the actors without significantly altering the script.

\section{Conclusion}\label{sec:conclusion}
The six decades of research surveyed here take us from \textsc{WordPro}'s four hand-crafted schemata for single-step arithmetic in Interlisp-D~\cite{fletcher1985understanding} to an advanced version of Gemini Deep Think producing five perfect IMO~2025 solutions in natural language within the competition's own 4.5-hour time limit~\cite{deepmind2025geminideepthink}. We have organized this progression along four axes, informal text, multimodal, formal, and discovery, each with its own characteristic methods, datasets, and triumphs.

If there is a single thread running through this history, it is that \emph{every expansion of AI mathematical reasoning capability has been made possible by a corresponding expansion of supervision}. Rule-based systems were supervised by grammarians. Statistical MWP solvers were supervised by equation templates. Seq2Tree neural networks were supervised by prefix-notation expressions. LLM-based reasoners were supervised first by chain-of-thought exemplars, then by process reward models, and most recently by the deterministic grader of Lean~4. Each successive supervisor is richer than the last: Lean does not merely tell the model that an answer is correct, it certifies \emph{why}.

In this light, the near-term future seems clear. The supervisor of tomorrow is not a single model but a \emph{pipeline}---often a small team of specialized agents, in which LLMs propose, critics and tools filter, autoformalizers translate, and Lean verifies. The mathematician's role is neither to be replaced by this pipeline nor to operate beside it as a passive equal, but to direct it: choosing problems, deciding which angles deserve pursuit, and, crucially, integrating the formal output back into the informal discourse that remains the medium of mathematical understanding. The \emph{primetime} that Tao~\cite{tao2026primetime} forecast in early~2026 is not one in which AI replaces mathematicians, but one in which the distance between a good research idea and a verified theorem becomes substantially shorter, and in which open problems that once required decades of sustained effort may, in some cases, yield to weeks of human-AI collaboration. The practical, philosophical, and infrastructural questions raised by this shift deserve the community's best attention over the coming years.

\section*{Acknowledgements}
We convey our heartfelt gratitude, in advance, to the anonymous reviewers for their constructive criticisms and insightful feedback, which will surely be conducive to the improvement of the research work outlined in this paper. Syed Rifat Raiyan, in particular, wishes to thank his parents, Syed Sirajul Islam and Kazi Shahana Begum, for everything.

\bibliographystyle{ieeetr}
\bibliography{ref3_ieee_preprint}

\end{document}